\documentclass[runningheads]{llncs}

\usepackage{eccv}

\usepackage{eccvabbrv}

\usepackage{graphicx}
\usepackage{booktabs}
\usepackage{soul}

\usepackage[accsupp]{axessibility}  

\usepackage[pagebackref,breaklinks,colorlinks]{hyperref}

\usepackage{multirow}
\usepackage{physics}
\usepackage{array,  makecell}
\usepackage[dvipsnames]{xcolor}
\usepackage{soul}  
\usepackage{tikz}
\usepackage{overpic}
\usepackage{svg}
\usepackage{subcaption}
\usetikzlibrary{tikzmark,arrows.meta,calc}

\newcommand{\ExtraLargeImageSize}{0.19\textwidth}
\newcommand{\LargeImageSize}{0.16\textwidth}

\newcommand{\ImageSize}{0.12\textwidth}
\newcommand{\SmallImageSize}{0.105\textwidth}
\newcommand{\TinyImageSize}{0.16\columnwidth}
\newcommand{\rd}{\mbox{RoofDiffusion}}
\newcommand{\rdN}{\mbox{No-FP RoofDiffusion}}
\newcommand{\rdF}{\mbox{RoofDiffusion}}

\newcommand{\img}[1]{\includegraphics[width=\ImageSize, height=\ImageSize, keepaspectratio]{#1}}

\newcommand{\ExtralargeImg}[1]{\includegraphics[width=\ExtraLargeImageSize, height=\ExtraLargeImageSize]{#1}}
\newcommand{\largeImg}[1]{\includegraphics[width=\LargeImageSize, height=\LargeImageSize]{#1}}

\newcommand{\smallImg}[1]{\includegraphics[width=\SmallImageSize, height=\SmallImageSize]{#1}}
\newcommand{\tinyImg}[1]{\includegraphics[width=\TinyImageSize, height=\TinyImageSize]{#1}}

\newcommand{\largeSubImg}[3]{
    \begin{subfigure}{\LargeImageSize}
        \centering
        \includegraphics[width=\textwidth, height=\textwidth]{#1}
        \vspace{-1.2\baselineskip}
        \caption{#2}
        \label{#3}
    \end{subfigure}
}

\newcommand{\smallSubImg}[3]{
    \begin{subfigure}{\SmallImageSize}
        \centering
        \includegraphics[width=\textwidth, height=\textwidth]{#1}
        \vspace{-1.2\baselineskip}
        \caption{#2}
        \label{#3}
    \end{subfigure}
}

\newcommand{\KK}{k}
\newcommand{\MM}{M}
\newcommand{\dsmmeter}{\boldsymbol{z}}
\newcommand{\iminx}{\dsmmeter_{\text{min}}}
\newcommand{\imaxx}{\dsmmeter_{\text{max}}}
\newcommand{\topbot}{\underline{\overline{\dsmmeter}}}

\newcommand{\estDSMmeter}{\hat{\dsmmeter}}
\newcommand{\gtDSMmeter}{\dsmmeter_{\text{gt}}}

\newcommand{\image}{\boldsymbol{x}}
\newcommand{\condImage}{\image}
\newcommand{\gtImage}{\image_{\text{gt}}}
\newcommand{\terminalImage}{\image_{T}}
\newcommand{\predictedImage}{\hat{\image}_0}
\newcommand{\treeImage}{\image_{\text{tree}}}
\newcommand{\nextEstImage}{\hat{\image}_{t-1}}

\newcommand{\model}{\boldsymbol{\epsilon}_{\theta}}
\newcommand{\normalDistribution}{\mathcal{N}(\boldsymbol{0}, \boldsymbol{I})}

\newcommand{\footprint}{\boldsymbol{m}}
\newcommand{\footprintinv}{\footprint'}

\newcommand{\pdata}{\mbox{PoznanRD}}

\newcommand{\imageH}{H}
\newcommand{\imageW}{W}
\newcommand{\gaussMix}{p_{\text{GMM}}}
\newcommand{\incompleteMask}{\boldsymbol{m}_{\text{incomplete}}}
\newcommand{\varMin}{\sigma_{\text{min}}}
\newcommand{\varMax}{\sigma_{\text{max}}}
\newcommand{\gaussMean}{\boldsymbol{\mu}_g}

\newcommand{\treeCount}{N_{\text{tree}}}
\newcommand{\treeCountMin}{N_{\text{tree}}^{\text{min}}}
\newcommand{\treeCountMax}{N_{\text{tree}}^{\text{max}}}

\newcommand{\xyScale}{s_{\text{xy}}}
\newcommand{\xyScaleMin}{s_{\text{xy}}^{\text{min}}}
\newcommand{\xyScaleMax}{s_{\text{xy}}^{\text{max}}}

\newcommand{\zScale}{s_{\text{z}}}
\newcommand{\zScaleMin}{s_{\text{z}}^{\text{min}}}
\newcommand{\zScaleMax}{s_{\text{z}}^{\text{max}}}

\usepackage{algorithm} 
\usepackage{algpseudocode} 

\begin{document}

\title{RoofDiffusion: 
Constructing Roofs from Severely Corrupted Point Data via Diffusion}

\titlerunning{RoofDiffusion}

\author{Kyle Shih-Huang Lo\inst{1} \and
Jörg Peters \inst{1} \and
Eric Spellman\inst{2}}

\authorrunning{K.~Lo et al.}

\institute{University of Florida, Gainesville FL 32611, USA
\\
\and
Meta Platforms, Inc.\\
\email{kyleshihhuanglo@ufl.edu, jorg@cise.ufl.edu, espellman@meta.com}}

\maketitle

\begin{figure}
\vspace{-2.0\baselineskip}
    \tiny
    \newcolumntype{G}{@{\hskip 2pt}}
    \newcolumntype{T}{m{0.01\textwidth}}
    \newcolumntype{I}{>{\centering\arraybackslash}m{\ImageSize}}
    \newcolumntype{B}{>{\centering\arraybackslash}p{\ImageSize}}
    
    \centering
    \begin{tabular}{T G I G I G I G I G I G I G I}

        \tikzmarknode{f}{\rotatebox{90}{Sparse}} & \img{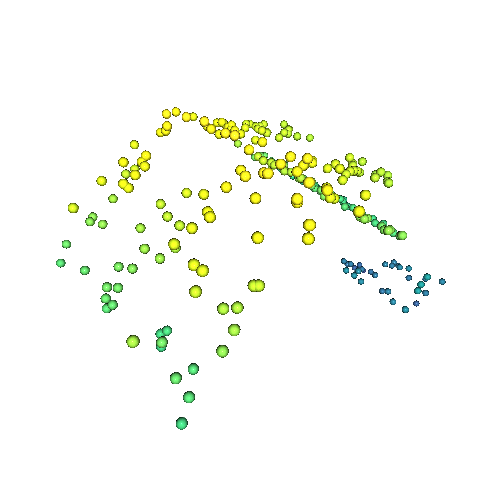} & \img{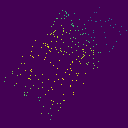} & \img{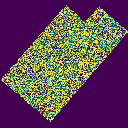} & \img{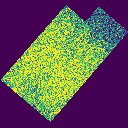} & \img{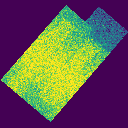} & \img{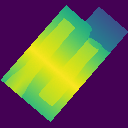} & \img{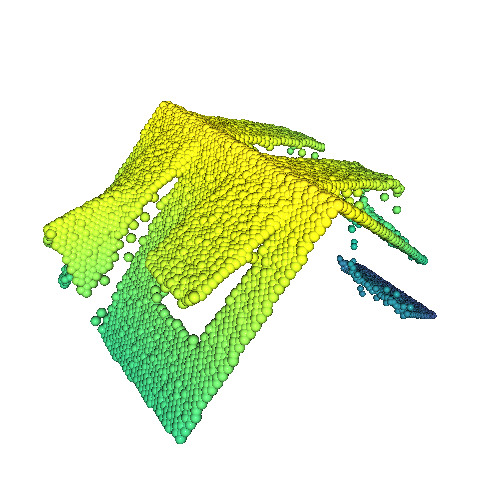} \\ [-0.3ex]
    
        \tikzmarknode{f}{\rotatebox{90}{Incomplete}} & \img{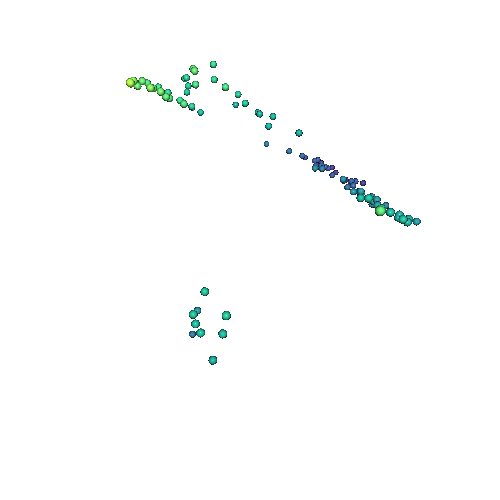} & \img{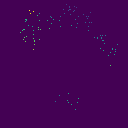} & \img{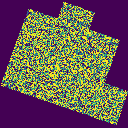} & \img{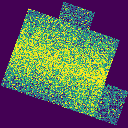} & \img{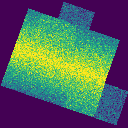} & \img{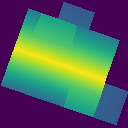} & \img{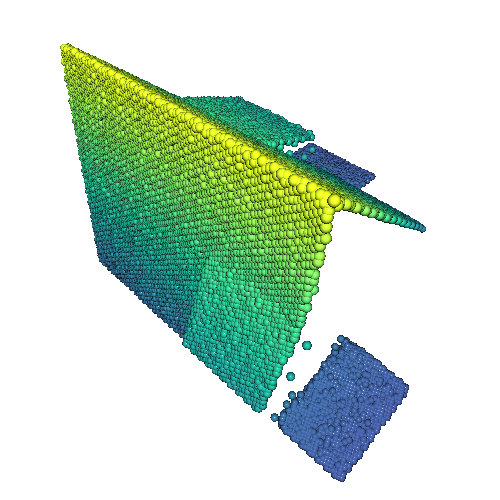} \\ [-0.3ex]
        
        \tikzmarknode{f}{\rotatebox{90}{Tree Noise}} & \img{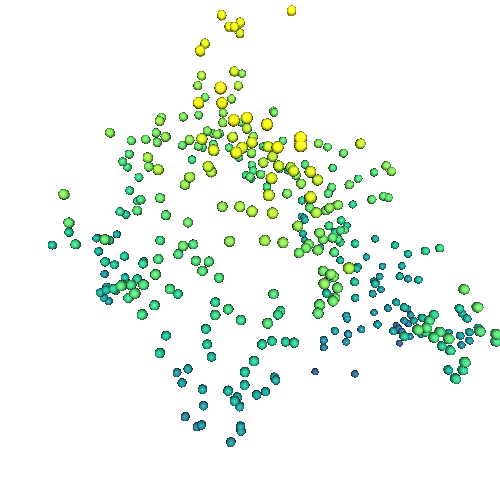} & \img{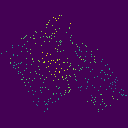} & \img{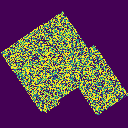} & \img{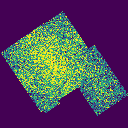} & \img{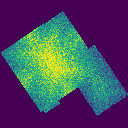} & \img{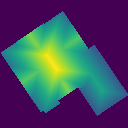} & \img{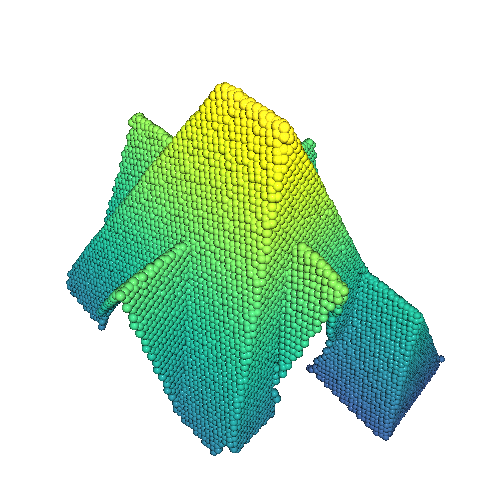} \\ [-0.3ex]
    \end{tabular}
    \begin{tabular}{T G B G B G B G B G B G B G B}
        & \makecell{Corrupted \\ Point Cloud}  & \makecell{Corrupted \\ Height Map} & \makecell{Denoise \\ 50\%} & \makecell{Denoise \\ 90\%} & \makecell{Denoise \\ 96\%} & \makecell{Restored \\ Height Map} & \makecell{Restored \\ Point Cloud} \\
    \end{tabular}

    \captionof{figure}{\rd{} restores height maps of
    challenging roof geometry, even under conditions of extreme sparsity, regional incompleteness, and noise.
    The bookend columns of point clouds are 3D views, the other columns are top views of height maps. 
    }
    \label{fig:overview}
    \vspace{-4.0\baselineskip}
\end{figure}

\begin{abstract}

Accurate completion and denoising of roof height maps are crucial to reconstructing high-quality 3D buildings. 
Repairing sparse points can enhance low-cost sensor use and reduce UAV flight overlap.
\rd\ is a new end-to-end self-supervised diffusion technique for robustly completing, in particular difficult, roof height maps.
\rdF{} leverages widely-available curated
footprints and can so handle up to 99\% point sparsity and 80\% roof area occlusion (regional incompleteness). A variant, \rdN{}, simultaneously predicts building footprints and heights. 
Both quantitatively outperform state-of-the-art unguided depth completion and representative inpainting methods for Digital Elevation Models (DEM), on both a roof-specific benchmark and the BuildingNet dataset.
Qualitative assessments show the effectiveness of \rd{} for datasets with real-world scans including AHN3, Dales3D, and USGS 3DEP LiDAR.
Tested with the leading City3D algorithm, preprocessing height maps with \rd{} noticeably improves 3D building reconstruction.
\rd{} is complemented by a new dataset of 13k complex roof geometries, focusing on long-tail issues in remote sensing;
a novel simulation of tree occlusion;
and a wide variety of large-area roof cut-outs for data augmentation and benchmarking.
Code and dataset\footnote{Created and released by the University of Florida}: \href{https://github.com/kylelo/RoofDiffusion}{github.com/kylelo/RoofDiffusion}

\end{abstract}

\begin{figure}[ht!]
  \centering
  \begin{minipage}[b]{0.49\textwidth}
    \centering
    \begin{subfigure}{0.31\linewidth}
        \centering
        \includegraphics[width=\linewidth]{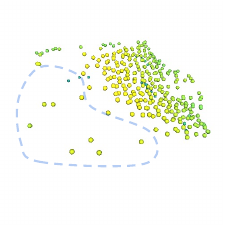}
        \caption{Sparse}
        \label{fig:realworld-sparse}
    \end{subfigure}
    \hfill
    \begin{subfigure}{0.33\linewidth}
        \centering
        \includegraphics[width=\linewidth]{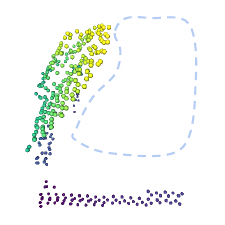}
        \caption{Incomplete}
        \label{fig:realworld-incomplete-a}
    \end{subfigure}
    \hfill
    \begin{subfigure}{0.31\linewidth}
        \centering
        \includegraphics[width=\linewidth]{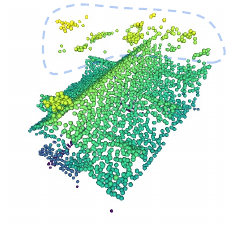}
        \caption{Tree Noise}
        \label{fig:realworld-tree}
    \end{subfigure}
    
    \caption{Types of corrupted roof height maps with real-world scan.}
    \label{fig:realworldscan}
  \end{minipage}
  \hfill
  \begin{minipage}[b]{0.48\textwidth}
    \centering
    \includegraphics[width=\textwidth]{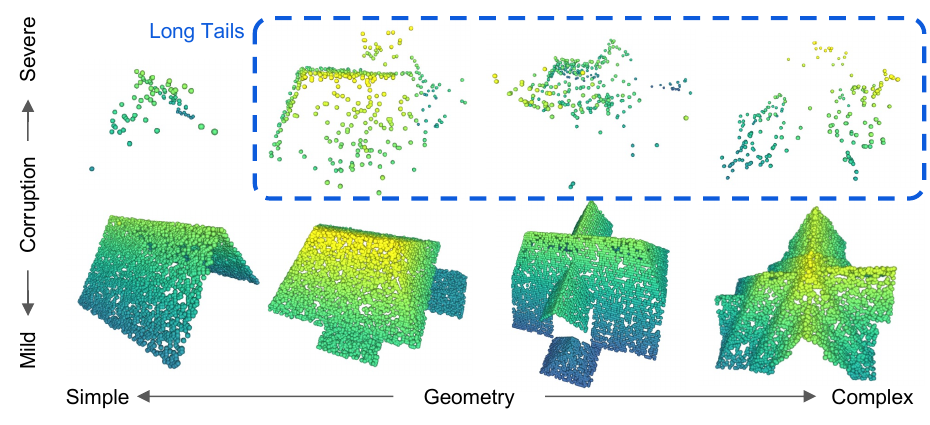} 
    \vspace{-1.8\baselineskip} 
    \caption{
    \rd{} focuses on the
    ``long~tail'' of hard-to-handle complex, corrupted geometry
    (inside the dashed blue).}
    \label{fig:longTail}
    \label{fig:second}
  \end{minipage}
  \vspace{-1.0\baselineskip} 
\end{figure}

\section{Introduction}
\label{sec:intro}

Digital Surface Models (DSMs), a.k.a.\ \emph{height maps}, are monochromatic images where the pixel value captures the elevation of features, both natural and artificial. Height maps can be created by rasterizing airborne Light Detection and Ranging (LiDAR) point clouds.
Height maps serve as an indispensable data source for reconstructing 3D models of urban buildings.
Numerous studies \cite{nan2017polyfit, holzmann2018semantically, huang2022city3d, yu2022finding} have explored methods for generating compact 3D building models from such data.
 3D building models are crucial in a wide range of applications, including navigation \cite{cappelle2012virtual}, urban planning \cite{liu2020three}, and simulations \cite{kolbe2005citygml, ntinas2018evaluation}. 
With $500$\MM\ buildings labeled in OpenStreetMap \cite{OpenStreetMap},
even a 1\% failure rate implies that $5$\MM\ buildings
need to be repaired. 
A close scrutiny of USGS 3DEP LiDAR sampled over Wayne County, MI \cite{wayneco2017} and 
Cambridge, MA \cite{cambridge2021}
reveals, respectively,
as much as $34$\% and $50$\% 
of height maps corrupted by incompleteness or noise.
While height maps are broadly accessible, the following challenges often impede compact and precise building reconstructions.

\begin{itemize}
    \item \textbf{Sparsity.} 
    Factors such as low sensor resolution and poor surface reflectance 
   cause low point density \cite{yan2023airborne}.
    (\cref{fig:realworld-sparse})
    \item \textbf{Incompleteness.} 
    Portions of the roof data can be missing, due to environmental interference, occlusions by taller surrounding objects, or roof substructure when the camera angle is not orthogonal to the ground \cite{shih2009building}.
    (\cref{fig:realworld-incomplete-a})
    \item \textbf{Noise.} Intrusions on building footprints, such as trees, can lead to incorrect reconstruction, resulting in artifacts like non-existing dormers. (\cref{fig:realworld-tree}).
   Environmental factors like light and dust can also introduce noise.
\end{itemize}

Since the reconstruction quality of the 3D models is heavily influenced by the precision and completeness of the underlying height maps, roof repair is crucial, but is under-explored.

\rd{} addresses all three bulleted challenges
by building on and adjusting reliable techniques.
Unlike typical image inpainting tasks such as \cite{lugmayr2022repaint, li2022mat, dong2022incremental, liu2022reduce, anciukevivcius2023renderdiffusion},
available height map pixels can be noisy, and 
completing sparse data does not fit the standard super-resolution paradigms \cite{gu2015convolutional, lai2017deep, zhang2020deep, liu2023spectral},
because missing pixels are usually unevenly distributed. 
Acknowledging these complexities, we 
nevertheless, succeed in conceptualizing roof repair as an image restoration task. 

Thanks to the efforts of contributors worldwide, OpenStreetMap \cite{OpenStreetMap} now increasingly provides vectorized high-precision footprints \cite{huang2022city3d,he2023high}. Consequently, recent building reconstruction algorithms \cite{huang2022city3d,he2023high,bagheri2019fusion} have incorporated footprints into the reconstruction process.
However, these approaches still struggle with sparsity,
incompleteness and noise.
Footprint-guided \rd{} addresses these 
long-tail challenges (\cref{fig:longTail}).

Specifically, we learn a strong prior that accurately approximates the actual distribution of the roof height map. 
Taking inspiration from the restoration approach in
\cite{saharia2022palette} of Joint Photographic Experts Group (JPEG) images,
we frame the roof height map restoration challenge as a denoising diffusion process.
While JPEG restoration focuses on reverting compressed images to their original state, our strategy seeks to fill in roof heights and remove noise.
The following are the major contributions of this paper:

\begin{itemize}
    \item We introduce \rd, a novel method based on conditional Diffusion Probabilistic Model (DPM) \cite{saharia2022palette}.
    Both footprint-guided \rdF{} and a variant, \rdN{}, trained without footprint, allow robust roof height map completion under extreme conditions, surpassing the state-of-the-art depth completion methods \cite{wang2022cu,eldesokey2020uncertainty} and representative DEM inpainting algorithms \cite{keys1981cubic,shepard1968two,biasutti2019diffusion}.
    The footprint-guided version handles up to 99$\%$ missing data points and 80$\%$ regional incompleteness while remaining resilient to tree occlusion noise 
    (for some applications, construction from very few data might be flagged 
    for hallucination potential).
    \rdN{} predicts both footprint and height.
    \item We propose a novel ``tree planting'' method for simulating tree occlusion noise, and we introduce multi-Gaussian masking for synthesizing incompleteness in roof height maps. These techniques enable data augmentation for self-supervised learning and benchmark creation for quantitative comparisons.
    \item We unveil the \pdata\ (Poznan Roof Dataset) with 13k Level of Detail (LoD) 2.2 \cite{biljecki2016improved} noise-free roof meshes and height maps. When treated with our noise and incompleteness algorithm,
    our complex roofs dataset can effectively generate ample training data to address ``long tail'' challenges, see \cref{fig:longTail}.
\end{itemize}

\section{Related Works}

Reviewing Digital Elevation Model (DEM) inpainting from remote sensing is most relevant, followed by unguided depth completion methods for restoring dense, clean depth maps from sparse LiDAR points.
Also, we review denoising diffusion models in computer vision tasks and roof datasets.

\textbf{DEM Inpainting} restores terrain height data missing caused by occlusions, such as mountains obstructing their own opposite sides and areas covered by water. 
Researchers commonly use Inverse Distance Weighting (IDW) \cite{shepard1968two}, Kriging \cite{reuter2007evaluation,mcbratney1986choosing}, and Spline fitting \cite{keys1981cubic} for inpainting voids in DEMs. 
These methods work well for small areas but can fail in larger, complex regions due to the lack of terrain geometry knowledge. To address larger inpainting regions, Delta surface-based approaches \cite{grohman2006filling,luedeling2007filling} 
fill voids with auxiliary DEM patches. 
Learning-based methods have gained prominence in DEM inpainting due to their superior feature learning capabilities. 
In particular, several methods \cite{dong2018filling,gavriil2019void,qiu2019void,zhang2020void,zhu2020spatial,zhou2022voids,yan2021high,dong2020shadow,li2022integrating} utilize Generative Adversarial Nets (GAN) \cite{goodfellow2014generative,mirza2014conditional} for inpainting by conditioning models on voids to predict filled areas.

Compared to DEM inpainting, roof DSM inpainting deals with higher sparsity, more noise from elements such as trees, and larger areas of incompleteness.

\textbf{Unguided Depth Completion} aims to restore the dense depth map solely from sparse depth data, mainly for urban driving data \cite{uhrig2017sparsity}.
The traditional Convolutional Neural Networks (CNNs) \cite{lecun2015deep} often suffer from a mosaic effect of images with sparse pixel values. 
This issue is addressed by applying convolutions solely to valid pixels using a binary mask \cite{uhrig2017sparsity,huang2019hms}.
However, a binary mask uniformly weights each pixel, conflicting with the reality that pixels have varying importance. 
To counter this, a continuous confidence mask was proposed in \cite{eldesokey2018propagating}. 
Furthermore, a learnable mask is introduced in \cite{eldesokey2020uncertainty} to address the issue that directly inferring a confidence mask from data may be problematic due to noise.
To incorporate semantic learning,
the works \cite{lu2020depth, yu2021grayscale} train the network to additionally predict auxiliary images, including RGB and normal maps.
CU-Net \cite{wang2022cu} employs a two-stage U-Net for coarse-to-fine depth completion. 

Most depth completion methods target evenly distributed, sparse depth maps without large missing areas.
\rd{} 
can handle 
height maps  missing  more than $80\%$ of the area and still generate
inpainting
harmonious with the existing roof structure.

\textbf{Diffusion Models} have been proposed in pioneering works \cite{ho2020denoising, song2020score, dhariwal2021diffusion} showing remarkable image generation capability. Subsequent research has built upon this foundation by introducing text-conditioned models, enabling text-guided image generation \cite{ruiz2023dreambooth, zhang2023adding, rombach2022high, ramesh2022hierarchical, nichol2021glide, saharia2022photorealistic}.
The technology has further been employed to significantly improve image enhancement tasks like super-resolution \cite{gao2023implicit, wu2023hsr, saharia2022image, ho2022cascaded}, inpainting \cite{saharia2022palette, lugmayr2022repaint, xie2023smartbrush}, 
and translation \cite{meng2021sdedit,li2023bbdm, parmar2023zero, saharia2022palette}.
There are also a few diffusion model-based methods for RGB-guided depth completion \cite{lei2023rgbd2, ran2023few}. 

To the best of our knowledge, our work pioneers the use of the diffusion model for DSM completion.
We formulate our problem as an image restoration task, leveraging a conditional DPM. 
Inspired by \cite{jaritz2018sparse},
our approach eliminates the need for a confidence map \cite{eldesokey2020uncertainty, uhrig2017sparsity, eldesokey2018propagating}, directly conditioning the model on the sparse image.

\textbf{Roof Datasets} commonly provide pairs of point clouds and ground truth mesh for 3D reconstruction research. 
UrbanScene3D \cite{lin2022capturing} and STPLS3D \cite{chen2022stpls3d} offer real-world LiDAR point clouds along with reconstructed dense triangular meshes. City3D \cite{huang2022city3d} and Building3D \cite{wang2023building3d} provide substantial datasets featuring more compact meshes for Computer-Aided Design (CAD)-like building-reconstruction research.
However, in these datasets where ground truth is constructed via 3D reconstruction algorithms, mesh accuracy can be suboptimal
due to algorithmic limitations and real-world scanning noise.
Government datasets like \cite{poznan3d,HelsinkiRegionInfoshare} provide compact noise-free meshes but exhibit misalignment with point clouds, often due to modeling simplifications.
To ensure alignment between compact meshes and point clouds, BuildingNet \cite{selvaraju2021buildingnet} generates point clouds directly from compact meshes crafted by artists. 
However, BuildingNet \cite{selvaraju2021buildingnet} provides only 2\KK\ buildings and contains non-building elements such as people and vehicles. 
Similarly, we can sample point clouds from datasets with compact meshes, \cite{wichmann2018roofn3d,qian2021roof,ren2021intuitive}, but these have limited roof types.

Considering the limitations in quantity and variety of compact meshes, we propose the \pdata, featuring 13\KK\ noise-free complex roofs. Besides, existing datasets do not focus on corrupted data, so we introduce a corruption synthesizing approach to generate
challenging data.

\begin{figure*}[t]
  \centering
    \tiny
    \begin{overpic}[width=\linewidth]{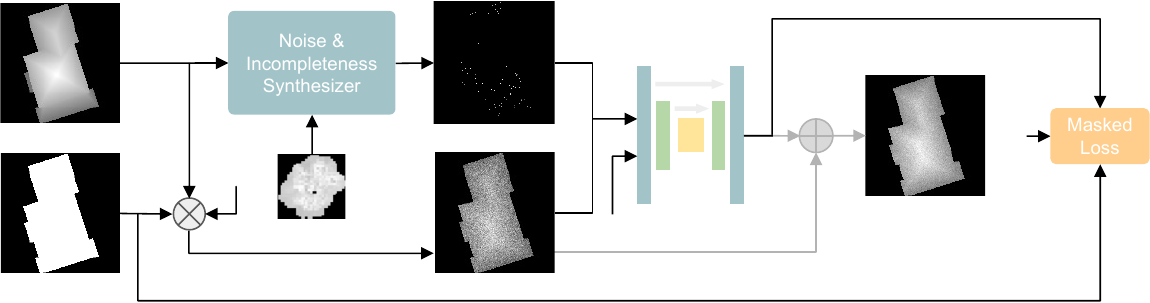}
        \put(0,14) {$\gtImage$: GT Heights}
        \put(0,1) {$\footprint$: Footprint}
        \put(17,11) {$\boldsymbol{\epsilon}$: Noise}
        \put(20,5.5) {$\treeImage$: Tree Heights}
        \put(32.8,14) {$\condImage$: Corrupted Heights}
        \put(56.2,7) {$\model$: U-Net}
        \put(52.8,6.2) {$\bar{\alpha}_t$}
        \put(35,1.1) {$\image_t$: Current State}
        \put(74.5,7.6) {$\nextEstImage$: Estimated}
        \put(74.5,6.0) {Previous State}
        \put(88,14) {$\boldsymbol{\epsilon}$}
     \end{overpic}
   \caption{\emph{Training the \rd{} reverse transition kernel.}
   GT denotes ground truth, $\scriptsize{\bigotimes}$ is transformation \eqref{equation:xt}, $\scriptsize{\bigoplus}$\protect\footnotemark removes predicted noise from $\boldsymbol{x}_t$ by \eqref{equation:reverse}. The masked loss is defined by \cref{equation:loss}.
   }
   \label{fig:framework}

\end{figure*}

\footnotetext{$\nextEstImage$ calculation can be ignored during training (the arrows colored as gray).}

\section{Problem Statement}
Given a noisy, sparse, and incomplete roof height map $\dsmmeter$ along with a building footprint $\footprint$, the goal of \rdF{} is to estimate a complete and noise-free height map $\estDSMmeter$ approximating to the ground truth $\gtDSMmeter$. 
Measured from ground level, $\dsmmeter$, $\estDSMmeter$, and $\gtDSMmeter$ are single-channel images whose pixel values indicate building heights in meters. In $\dsmmeter$, zero-valued pixels signify either missing measurements or ground level. For $\estDSMmeter$ and $\gtDSMmeter$, zero values represent ground level. 
Generally, the number of non-zero pixels in $\estDSMmeter$ and $\gtDSMmeter$ is significantly greater than in $\dsmmeter$.
The mask
 $\footprint$ delineates the footprint,
 i.e.\ equals 1 for pixels belonging to 
 the roof, and 0 otherwise.

Accurate reconstruction of real-world height maps is challenging due to a wide range of building heights, noise, and missing data. 
While the typical normalization
for images or depth scenes of autonomous driving can employ fixed min-max values,
the normalization required for roof heights needs to adapt to the wide range of roof heights for each building.
\section{\rdF{}}
We introduce a diffusion model-based roof completion method to reconstruct $\estDSMmeter$ from $\dsmmeter$ so that $\estDSMmeter$ is close to $\gtDSMmeter$.
\Cref{subsection:normalization} presents the conversion of input height maps $\dsmmeter$ to roof-focused height images $\condImage$ in the range $[-1, 1]$ suitable for diffusion models.
\Cref{subsection:completion} applies a diffusion model, conditioned on corrupted normalized height maps $\condImage$ within their building footprint $\footprint$, to predict complete and noise-free height maps $\predictedImage$.

\subsection{Roof-Focused Height Map Normalization}
\label{subsection:normalization}
Diverse building heights and the fixed value range of diffusion models require a careful normalization approach.
To focus solely on the roof structure, we identify the lowest roof pixel
and subtract this value from the entire height map.
Let 
$\delta_i$ 
be the difference between the maximum and minimum height of the roof structure for $i$-th building .
Then $\topbot$ is the maximum of the $\delta_i$ after removing the largest 1\%
of $\delta_i$.
The normalized input height image is
\begin{equation}
    \condImage := \frac{2}{\topbot}\left(\dsmmeter - \frac{1}{2}\left( \imaxx + \iminx \right) \right)
    \in
    [-1,1],
    \label{eq:normalize}
\end{equation}
where
$\iminx$ denotes the smallest nonzero value in $\dsmmeter$
and $\imaxx$ the largest. 
 Our analysis of 13\KK\ buildings shows the cut-off value for $\topbot$ is 10 meters.

\subsection{Height Completion based on a Diffusion Model}
\label{subsection:completion}
Given  $\condImage$, our goal is to generate a predicted image, $\predictedImage$, that closely approximates the normalized ground truth image, $\gtImage$. We conceptualize this height completion task as a diffusion process. Drawing on the work of \cite{ho2020denoising}, our approach employs both a \textit{forward} and a \textit{reverse} Markov chain. The forward chain perturbs the ground truth data through noise injection, while the reverse chain utilizes a learnable model, conditioned on the input, to restore the data, see \cref{fig:framework}.

\textbf{The Forward Process} transforms a complex distribution
$q(\gtImage)$, the ground truth height maps, 
into the  Gaussian distribution $\normalDistribution$.
Starting with  $\image_0 = \gtImage \sim q(\gtImage)$, a transition kernel $q(\image_t \mid \image_{t-1})$ generates a series of random variables
$\image_t$,  $t \in \{1, \ldots, T \}$.
The joint distribution of $q(\image_1, ..., \image_T \mid \image_0)$, namely 
$\prod_{t=1}^{T} q(\image_t \mid \image_{t-1})$, 
is marginalized based on \cite{sohl2015deep}
to derive closed-form expression 
\begin{align}
    \image_t &:= f_{\text{forward}} \left( t, \image_0, \boldsymbol{\epsilon}, \footprint \right) \\
    &~= \footprint \odot \left(\sqrt{\bar{\alpha}_t} \image_0 + \sqrt{1 - \bar{\alpha}_t} \boldsymbol{\epsilon} \right) - \footprintinv,
\label{equation:xt}
\end{align}
where $\odot$ denotes element-wise multiplication, $\alpha_i$ is a hyperparameter, $\bar{\alpha}_t = \prod_{i=0}^t \alpha_i$, and $\boldsymbol{\epsilon} \sim \normalDistribution$.
Subtracting $\footprintinv$, the complement mask of $\footprint$, sets pixel values outside the footprint to -1 for
two advantages.
First, the model can infer the building footprint directly from the noise-injected areas in $\image_t$, obviating the need for an additional channel to represent the footprint. Second, non-building heights are prevented from influencing the prediction.

\textbf{The Reverse Process} sequentially removes noise from the data to yield a complete, noise-free version. Specifically, we can draw a noisy image $\terminalImage$ from $\normalDistribution$ at time $t=T$. 
Then use a reverse transition kernel to recursively eliminate noise until $t=0$, resulting in the restored height image, $\predictedImage$.
This transition kernel is parameterized by a learnable model $\model$. 
While $\predictedImage$ conforms to $q(\gtImage)$, it may not always correspond to
the normalized corrupted height image, $\condImage$. 
In our case, the objective is to ensure that the repaired roofs 
seamlessly integrate with pre-existing structures. 
To  ensure a close association between $\condImage$ and $\predictedImage$,
we condition $\model$ on $\image_t$,
$\condImage$, and $\bar{\alpha}_t$, in the spirit of  %
\cite{saharia2022palette}.
Each step of the reverse process can be formulated~as
\begin{equation}
    \nextEstImage
    \leftarrow \frac{1}{\sqrt{\alpha_t}} \left( 
    \image_t
    - \frac{1 - \alpha_t}{\sqrt{1 - \bar{\alpha}_t}} \model \right) + \sqrt{1 - \alpha_t} \boldsymbol{\epsilon}_t,
    \label{equation:reverse}
\end{equation}
where $\alpha_t$ and $\bar{\alpha}_t$ are hyperparameters for variance scheduling, and $\boldsymbol{\epsilon}_t \sim \normalDistribution$. 

\textbf{Loss Function.} For accurate noise prediction at various time steps $t$ 
 we train $\model$ with  $L_1$ loss restricted to the \mbox{footprint} $\footprint$.
 For $t \sim \mathcal{U}\left(1, T\right)$ and  $\boldsymbol{\epsilon} \sim \normalDistribution$,
\begin{gather}
    L 
    := \mathbb{E}_{\left(\image_0, \condImage, \footprint \right),t,\boldsymbol{\epsilon}}
    \norm{ \footprint \odot \left( \boldsymbol{\epsilon} - \Tilde{\boldsymbol{\epsilon}}_{\theta} \right) }_1
    \label{equation:loss}
    \\
  \text{ where } \Tilde{\boldsymbol{\epsilon}}_{\theta}
    := \model \left( f_{\text{forward}} \left( t, \image_0, \boldsymbol{\epsilon}, \footprint \right) , \condImage, \bar{\alpha}_t \right).
\end{gather}

\noindent
To restore the original scaling in meters for the minimum and maximum extent of a roof, we reverse \eqref{eq:normalize}, i.e.\ multiply $\predictedImage$ by $\frac{1}{2}\topbot$ and add
$\frac{1}{2}(\iminx + \imaxx )$.

\section{Datasets \& Benchmarks}
Simulating the real-world requires
noisy, sparse, and incomplete height maps.
Ground truth requires a high-resolution, noise-free, complete height map;
and to tackle long-tail 
geometric complexity illustrated
in \cref{fig:longTail}, the buildings
should be more complex than existing datasets with a LoD 2.2 \cite{biljecki2016improved}
(LoD 2.2 captures intricate roof details like dormers and gables, that LoD 2.0 overlooks).

To aid machine learning research, we curated a new dataset that not only serves our research needs, but future research needs in the field. 
We started with 16\KK\ compact and high-detail LoD 2.2 roof meshes from the city of Poznan, Poland \cite{poznan3d}.
To match our focus on complex roof structures (\cref{fig:longTail}), we re-balanced the dataset by reducing the number of flat roof from 5\KK\ to 2\KK\ resulting in a new dataset of 13\KK\ buildings.

\begin{figure}[t]
    \begin{minipage}[b]{0.6\textwidth}
        \tiny
        \newcolumntype{G}{@{\hskip 3pt}}
        \newcolumntype{T}{m{0.015\textwidth}}
        \newcolumntype{I}{>{\centering\arraybackslash}m{\LargeImageSize}}
        \centering
        \begin{tabular}{IIIIII}
            \tikzmarknode{a}{\makecell{GT}}  & \tikzmarknode{b}{\makecell{GT Height}} & \tikzmarknode{c}{\makecell{Footprint}} & \tikzmarknode{d}{\makecell{Tree Noise}} & \tikzmarknode{e}{\makecell{Sparsity}} & \tikzmarknode{f}{\makecell{Incomplete}} \\ 
        \end{tabular}
        \begin{tabular}{IIIIII}
            \largeImg{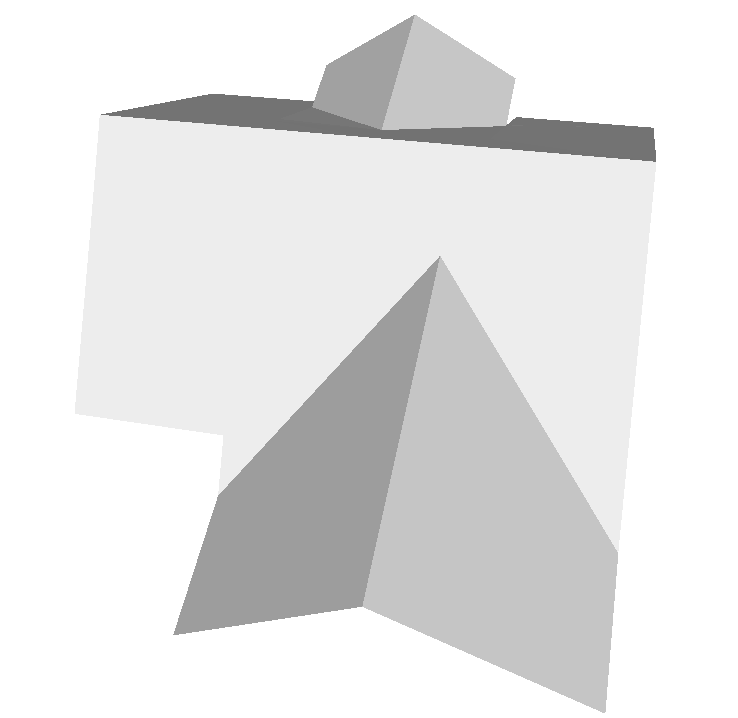} &
            \largeImg{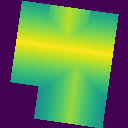} & \largeImg{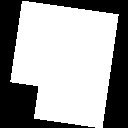} & \largeImg{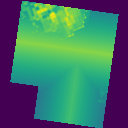} & \largeImg{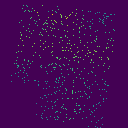} & \largeImg{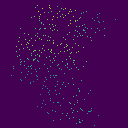} \\
            \largeImg{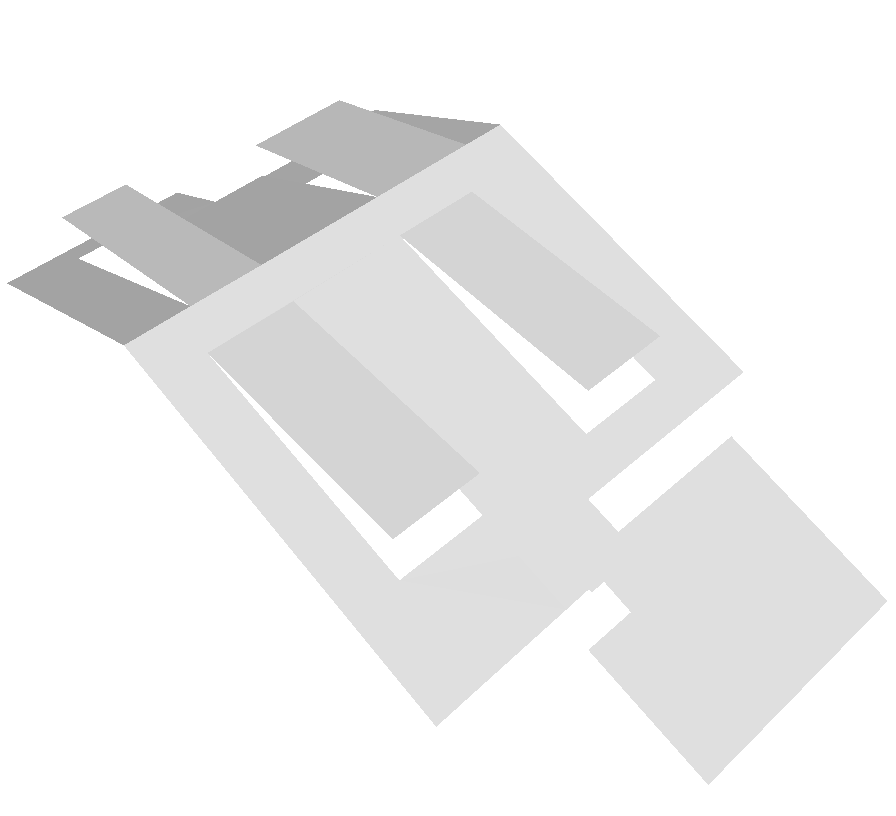} &
            \largeImg{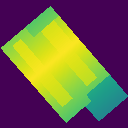} & \largeImg{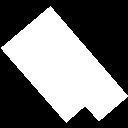} & \largeImg{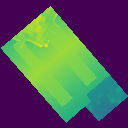} & \largeImg{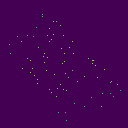} & \largeImg{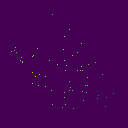} \\
            
        \end{tabular}
        
        \begin{tikzpicture}[overlay,remember picture]
            \coordinate (X) at (a.east);
            \draw[->, line width=0.1mm] let \p0=(X), \p2=(b.west) in (\x0 + 4pt,\y0) -- (\x2 - 2pt,\y2);
            \draw[->, line width=0.1mm] let \p1=(b.east), \p2=(c.west) in (\x1+ 1pt,\y1) -- (\x2- 1pt,\y2);
            \draw[->, line width=0.1mm] let \p1=(c.east), \p2=(d.west) in (\x1+ 1pt,\y1) -- (\x2- 1pt,\y2);
            \draw[->, line width=0.1mm] let \p1=(d.east), \p2=(e.west) in (\x1+ 1pt,\y1) -- (\x2- 1pt,\y2);
            \draw[->, line width=0.1mm] let \p1=(e.east), \p2=(f.west) in (\x1+ 1pt,\y1) -- (\x2- 1pt,\y2);
        \end{tikzpicture}

        \vspace{-1.5\baselineskip}
        
        \caption{Examples of synthesizing corrupted height maps from the \pdata.}
        \label{fig:synsthesizingdata}

    \end{minipage}
    \hfill
    \begin{minipage}[b]{0.35\textwidth}
        \includegraphics[width=\textwidth]{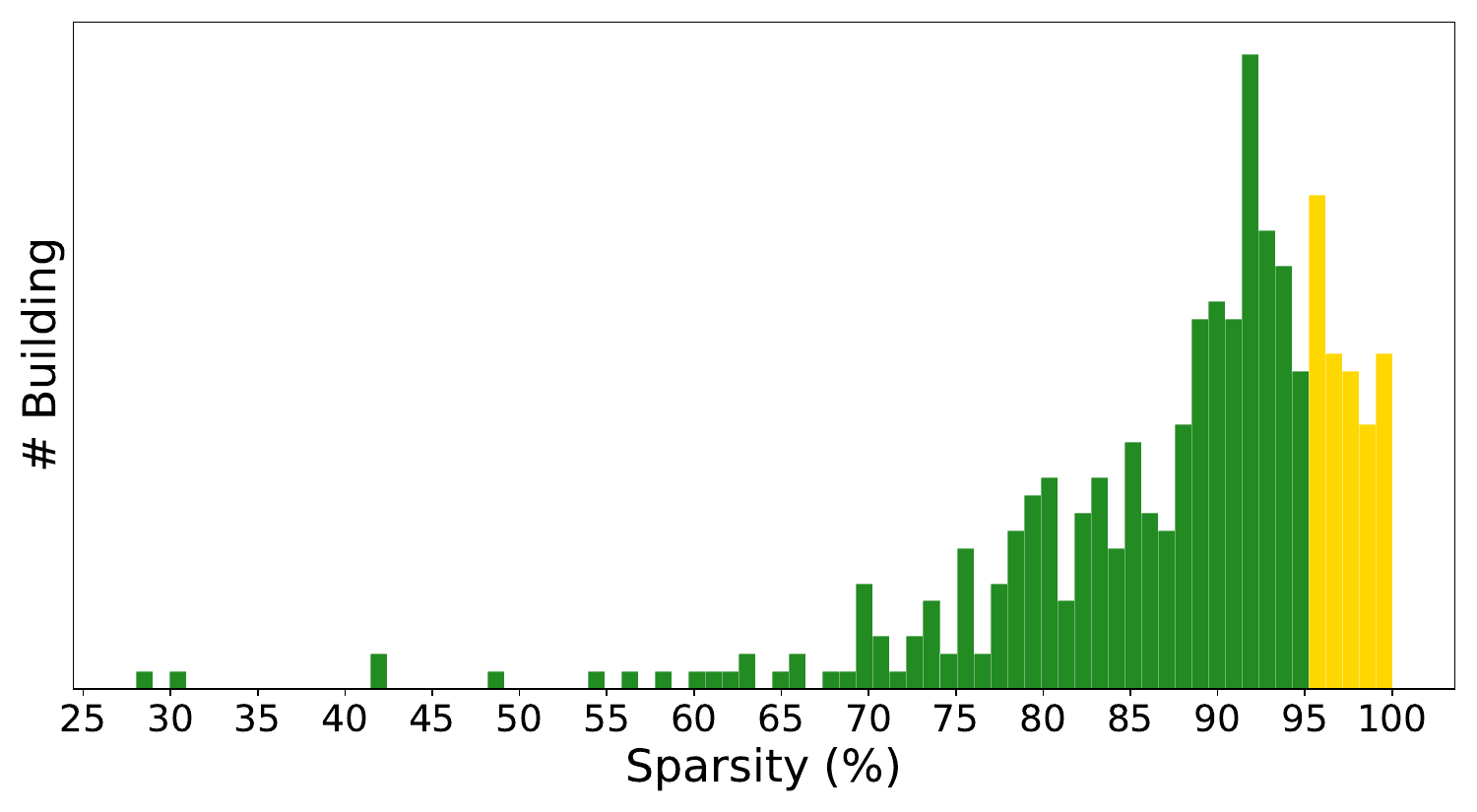}
        \vspace{-1.5\baselineskip}
        \caption{Sparsity of the real-world USGS 3DEP LiDAR dataset \cite{cambridge2021,wayneco2017}.}
        \label{fig:usgsdistribution}
    \end{minipage}
    \vspace{-1.0\baselineskip}
\end{figure}

To address the long-tail corruption depicted in \cref{fig:longTail}, collecting diverse corrupted height maps is essential. 
However, acquiring real-world corrupted data is challenging due to the difficulty in obtaining ground truths and identifying corrupted height maps.
Therefore, we developed a procedure for synthesizing intentionally corrupted height maps 
suitable for training or benchmarking, as illustrated in \cref{fig:synsthesizingdata}.
We first rasterize the roof mesh triangles into ground truth height maps. This process provides complete height maps without missing points and also reconciles real-world captured height maps and artist-generated meshes. Next, footprints are inferred from non-zero pixels in height maps. 
Then, we simulate real-world conditions by adding synthetic tree noise\footnote{Although we focus on tree noise for clarity, we also simulate global Gaussian noise and outlier noise in practice.}, creating sparsity through random point removal, and injecting incompleteness.

In this paper, ``s'' denotes Sparsity (\%), the ratio of randomly removed to total pixels in the footprint. Similarly, ``i'' signifies the Incompleteness (\%), the proportion of pixels removed due to incompleteness in the footprint.

\vspace{-0.5\baselineskip}

\subsection{Tree Points}
\vspace{-0.2\baselineskip}

\label{subsection:treepoint} 
In analyzing height maps within a footprint, tree crowns may intrude, causing LiDAR to capture the tree canopy instead of the roof, see \cref{fig:realworld-tree}.
Thus, we simulate tree intrusion by planting virtual trees around the roof.
We collected a real-world database with 1\KK\ tree height maps from \cite{poznan3d}. 
Small trees are randomly placed outside the footprint, with random rotations and height adjustments. 
A ``max'' operation is used to combine the roof $\image$ and tree height map $\treeImage$.  Algorithm 3 in supplements has the details.

\vspace{-0.5\baselineskip}

\subsection{Incompleteness}
\vspace{-0.2\baselineskip}
\label{subsection:incompleteness}

To synthesize incompleteness, we employ multiple Gauss masks with various means and variances to indicate missing points. This approach emulates occlusions caused by more intricate objects with soft (\cref{fig:realworld-sparse}) or hard (\cref{fig:realworld-incomplete-a}) boundaries, in the spirit of a Gaussian Mixture Model \cite{reynolds2009gaussian}. The rationale behind using Gauss masks is as follows: 
When multiple Gaussian masks are randomly positioned on the same side of a building, they can completely block one side of the roof point clouds, akin to the ray-tracing-free simulation of occlusion by \cite{yu2021pointr}, where the authors remove points distant from the camera.
Conversely, when scattered and smaller Gaussian masks are utilized within the roof structure, they effectively simulate the occlusion caused by small roof features.
Lastly, a larger variance of Gaussian simulates softer boundaries. The supplement details efficient mask generation for training in Algorithm 1, and mask creation with a specific portion of missing points in Algorithm 2.

\subsection{Benchmarks} 
\label{subsection:benchmark} 
The above height map synthesizing technique allows us to create benchmarks tailored to various research needs. 
Specifically, we can adjust different levels of sparsity, incompleteness, and the number of trees introduced. 
This makes it ideal for both training and evaluating machine learning models geared towards 3D reconstruction 
and other related fields. 
We have partitioned the dataset into a training split of 10k samples and a test split of 3k samples. Additionally, for tree-height maps, we provide 766 samples for training and 255 for testing.

\section{Experiments}

\Cref{subsection:hmquantitative} demonstrates the effectiveness of \rd{} and \rdN{}, trained without footprint, by quantitative evaluation on two datasets with ground truth: our \pdata{} and the BuildingNet \cite{selvaraju2021buildingnet}.
\Cref{subsection:footprint-recovery} illustrates the ability of \rdN{} to recover footprints from corrupted height maps.
\Cref{subsection:reconstruction} assesses the 3D mesh reconstruction quality for the City3D \cite{huang2022city3d}, comparing results with, and without using \rd{} as a pre-processor.
\Cref{subsection:hmqualitative} showcases qualitative outcomes through tests on real-world scans in AHN3 \cite{huang2022city3d}, Dales3D \cite{varney2020dales}, and USGS 3DEP \mbox{LiDAR} sampled over Cambridge, MA \cite{cambridge2021}, and Wayne County, MI \cite{wayneco2017}. 
\Cref{subsection:limitations} discusses the limitations. 

The supplement provides ablation of tree augmentation, sampling step analysis, impact of LiDAR scan pattern, additional qualitative results, and implementation details.
Additionally, the supplement shows an extension of \linebreak \rd{} to be comparable to existing unguided depth completion methods on the KITTI dataset \cite{uhrig2017sparsity}.

\subsection{Quantitative Evaluation of Height Completion}
\label{subsection:hmquantitative}
\newcommand{\fp}{FP}
\newcommand{\nofp}{No-\fp}
We examine \rd{} in the four relevant scenarios. Footprints (\fp) are derived by converting the ground truth height map into a binary mask.
\begin{enumerate}
    \item (\fp)
    \emph{Both} sparsity \emph{and} incompleteness on \pdata{} in \cref{tab:poznanquantitative}.
    \item (\fp) \emph{Either} sparsity \emph{or} incompleteness on BuildingNet \cite{selvaraju2021buildingnet} in \cref{tab:poznanquantitative}.
    \item (\nofp) \emph{Both} sparsity \emph{and} incompleteness in the \pdata{} in \cref{tab:nfpoznanquantitative}.
    \item (\nofp) \emph{Either} sparsity \emph{or} incompleteness on BuildingNet \cite{selvaraju2021buildingnet} in \cref{tab:nfpoznanquantitative}.
\end{enumerate}

\textbf{PoznanRD Dataset}. We conducted tests on $1$\KK\ randomly selected data from test split of \pdata, 
injected with both global and outlier noise to simulate real-world cases.
We follow \cite{yu2018pu, yuan2018pcn, li2022point2roof} to emulate global noise by incorporating Gaussian noise into all the normalized data points, with $\sigma_{\text{global}} \sim \left[0, 0.05\right]$ for each height map. 
Also, outliers are introduced by randomly assigning nonzero pixels with random values, as suggested in \cite{zeng2022lion}, where the probability for occurrence of an outlier is set to 0.01 \%.
Additionally, we introduced tree noise in 30\% of the cases, involving between 1 to 3 trees, following \cref{subsection:treepoint}.

\textbf{BuildingNet} \cite{selvaraju2021buildingnet} contains compact and noise-free building meshes spanning various categories.
These models are mostly sourced from 3D artists and often include additional elements like virtual ground, cars, and trees in the scene.
We picked 200 buildings that contain only the building structure, making it easier to infer footprints. 
We rasterized the meshes in BuildingNet \cite{selvaraju2021buildingnet} into height maps as ground truths.
It is important to note that our tests were conducted using a model trained exclusively on \pdata.

\begin{table}[t]
    \setlength{\tabcolsep}{3.0pt}
    \newcolumntype{g}{@{\hskip 2.0pt}}
    \newcolumntype{k}{@{\hskip 2pt}}
    \centering
    \scriptsize
    \resizebox{\textwidth}{!}{
    \begin{tabular}{klcgccgccgccgccgccgccgccgc}
        \toprule
        \multirow{3}{*}{Methods} & \multicolumn{8}{c}{PoznanRD} & \multicolumn{8}{c}{BuildingNet} \\
        \cmidrule(l{0.3em}r{0.3em}){2-9} \cmidrule(l{0.3em}r{0.3em}){10-17}
        & \multicolumn{2}{c}{s95 i30} & \multicolumn{2}{c}{s95 i80} & \multicolumn{2}{c}{s99 i30} & \multicolumn{2}{c}{s99 i80} 
        & \multicolumn{2}{c}{s90 i80} & \multicolumn{2}{c}{s90 i90} & \multicolumn{2}{c}{s99 i0} & \multicolumn{2}{c}{s99.75 i0} \\
        \cmidrule(l{0.3em}r{0.3em}){2-3} \cmidrule(l{0.3em}r{0.3em}){4-5} \cmidrule(l{0.3em}r{0.3em}){6-7} \cmidrule(l{0.3em}r{0.3em}){8-9}
        \cmidrule(l{0.3em}r{0.3em}){10-11} \cmidrule(l{0.3em}r{0.3em}){12-13} \cmidrule(l{0.3em}r{0.3em}){14-15} \cmidrule(l{0.3em}){16-17}
        & \tiny MAE & \tiny RMSE & \tiny MAE & \tiny RMSE & \tiny MAE & \tiny RMSE & \tiny MAE & \tiny RMSE 
        & \tiny MAE & \tiny RMSE & \tiny MAE & \tiny RMSE & \tiny MAE & \tiny RMSE & \tiny MAE & \tiny RMSE \\
        \midrule

        Linear  & 0.236 & 0.461 & 0.687 & 1.037 & 0.365 & 0.631 & 0.868 & 1.218 & 0.654 & 0.997 & 0.868 & 1.224 & 0.297 & 0.549 & 0.528 & 0.833 \\
        Spline   & 0.278 & 0.508 & 0.785 & 1.198 & 0.391 & 0.659 & 0.888 & 1.260 & 0.829 & 1.288 & 1.033 & 1.523 & 0.330 & 0.586 & 0.536 & 0.845 \\
        Nearest & 0.288 & 0.541 & 0.691 & 1.089 & 0.424 & 0.734 & 0.866 & 1.271 & 0.634 & 1.044 & 0.856 & 1.287 & 0.347 & 0.662 & 0.582 & 0.961 \\
        IDW & 0.239 & 0.449 & 0.648 & 0.984 & 0.377 & 0.619 & 0.827 & 1.172 & 0.573 & 0.897 & 0.808 & 1.159 & 0.309 & \textbf{0.540} & 0.537 & 0.822 \\
        P.M. Diff. & 0.266 & 0.473 & 1.825 & 2.311 & 0.523 & 0.739 & 3.085 & 3.548  & 1.090 & 1.448 & 1.775 & 2.136 & 0.414 & 0.706 & 0.743 & 0.979 \\
        \midrule
        Ours  & \textbf{0.162} & \textbf{0.342} & \textbf{0.430} & \textbf{0.727} & \textbf{0.253} & \textbf{0.463} & \textbf{0.603} & \textbf{0.916} & \textbf{0.508} & \textbf{0.871} & \textbf{0.705} & \textbf{1.071} & \textbf{0.280} & 0.550 & \textbf{0.447} & \textbf{0.751} \\
        \bottomrule
    \end{tabular}
    }

    \vspace{+0.5\baselineskip}
    
    \caption{Evaluation of height map completion on \pdata\ and BuildingNet \cite{selvaraju2021buildingnet} dataset, measured by Mean Absolute Error (MAE) and Root Mean Square Error (RMSE) in meters. \textbf{Bold} represents best outcome.}
    
    \label{tab:poznanquantitative}
    \vspace{-1.0\baselineskip}
\end{table}

\textbf{(\fp) 
\emph{Both} sparsity \emph{and} incompleteness.}
We evaluated combinations of sparsity at 95\% and 99\%, along with incompleteness at 30\% and 80\%, using the method shown in \cref{subsection:incompleteness}. Notably, our sparsity selection is according to USGS 3DEP LiDAR with Quality Level (QL) 2.
Note that 
QL2 only guarantees at least 2 points per square meter and
that most of the US territory LiDAR is QL 2 or lower.
In fact, it is common to find roof data with extreme sparsity, as shown as yellow bars in \cref{fig:usgsdistribution}.
While we noticed several cases whose incompleteness exceeded 95\%, the combination of this level of incompleteness with severe sparsity makes the analysis overly difficult and potentially meaningless. Therefore, we set the upper limit for incompleteness at 80\%.

In this evaluation, we benchmark against interpolation techniques commonly employed in DEM inpainting, including linear, nearest-neighbor, Spline \cite{keys1981cubic}, and IDW \cite{shepard1968two} interpolation.
In particular, we compare with the latest DSM inpainting technique \cite{biasutti2019diffusion} based on Perona-Malik Diffusion (P.M. Diff.) \cite{perona1990scale}. 
\Cref{tab:poznanquantitative} shows that our approach consistently surpasses all these methods in height restoration, regardless of the varying degrees of sparsity and incompleteness. 

\textbf{(\fp) \emph{Either} sparsity \emph{or} incompleteness.}
Here, we assess the handling of pure sparsity (s99 i0, s99.75 i0) and incompleteness (s90 i80, s90 i90) independently. 
We maintain the same noise injection settings as those used for the \pdata{} dataset but exclude tree noise.
\Cref{tab:poznanquantitative} shows RoofDiffusion outperforms most DEM interpolation methods and demonstrates generalizability to unseen BuildingNet \cite{selvaraju2021buildingnet} datasets. \rd\ demonstrates a stronger performance advantage for the incompleteness task, suggesting  particular effectiveness at restoring structural information.

\begin{table}[t]
    \setlength{\tabcolsep}{3.0pt}
    \newcolumntype{g}{@{\hskip 2.0pt}}
    \newcolumntype{k}{@{\hskip 2pt}}
    \centering
    \scriptsize
    
    \resizebox{\textwidth}{!}{
    
    \begin{tabular}{klcgccgccgccgccgccgccgccgc}
        \toprule
        \multirow{3}{*}{Methods} & \multicolumn{8}{c}{PoznanRD} & \multicolumn{8}{c}{BuildingNet} \\
        \cmidrule(l{0.3em}r{0.3em}){2-9} \cmidrule(l{0.3em}r{0.3em}){10-17}
        & \multicolumn{2}{c}{s95 i30} & \multicolumn{2}{c}{s95 i60} & \multicolumn{2}{c}{s99 i30} & \multicolumn{2}{c}{s99 i60} 
        & \multicolumn{2}{c}{s90 i60} & \multicolumn{2}{c}{s90 i80} & \multicolumn{2}{c}{s98 i0} & \multicolumn{2}{c}{s99 i0} \\
        \cmidrule(l{0.3em}r{0.3em}){2-3} \cmidrule(l{0.3em}r{0.3em}){4-5} \cmidrule(l{0.3em}r{0.3em}){6-7} \cmidrule(l{0.3em}r{0.3em}){8-9}
        \cmidrule(l{0.3em}r{0.3em}){10-11} \cmidrule(l{0.3em}r{0.3em}){12-13} \cmidrule(l{0.3em}r{0.3em}){14-15} \cmidrule(l{0.3em}){16-17}
        & \tiny MAE & \tiny RMSE & \tiny MAE & \tiny RMSE & \tiny MAE & \tiny RMSE & \tiny MAE & \tiny RMSE 
        & \tiny MAE & \tiny RMSE & \tiny MAE & \tiny RMSE & \tiny MAE & \tiny RMSE & \tiny MAE & \tiny RMSE \\
        \midrule

        Linear  & 0.778 & 1.966 & 1.330 & 2.811 & 1.091 & 2.542 & 1.969 & 3.625 & 0.568 & 1.302 & 1.517 & 2.484 & 0.325 & 0.926 & 0.426 & 1.111 \\
        pNCNN  & 1.635 & 3.016 & 1.885 & 3.169 & 2.012 & 3.298 & 2.172 & 3.378 & 0.918 & 1.605 & 1.185 & 1.822 & 0.904 & 1.632 & 1.063 & 1.765 \\
        CU-Net & 1.246 & 1.823 & 1.628 & 2.244 & 1.544 & 2.187 & 1.923 & \textbf{2.554} & 0.675 & 1.280 & 1.641 & \textbf{2.386} & 0.323 & \textbf{0.717} & 0.397 & \textbf{0.829} \\
        \midrule
        Ours (\nofp)  & \textbf{0.319} & \textbf{1.232} & \textbf{0.769} & \textbf{2.018} &\textbf{ 0.722} & \textbf{1.968} & \textbf{1.200} & 2.600 & \textbf{0.509} & \textbf{1.228} & \textbf{1.501} & 2.449 & \textbf{0.262} & 0.803 & \textbf{0.349} & 0.962 \\
        \bottomrule
    \end{tabular}
    }

    \vspace{+0.5\baselineskip}
    
    \caption{Evaluation of height map completion \emph{w/o footprint} on the PoznanRD and BuildingNet \cite{selvaraju2021buildingnet} datasets.}

    \label{tab:nfpoznanquantitative}

    \vspace{-1.5\baselineskip}
    
\end{table}

\begin{table}[t]
    \begin{minipage}[t]{0.51\textwidth}
        \vspace{0pt}
        \centering
        \scriptsize
        \resizebox{\textwidth}{!}{
        \begin{tabular}{lccccc}
            \toprule
            \multirow{2}{*}{Methods} & \multicolumn{2}{c}{s99 i30} & \multicolumn{2}{c}{s99 i80} \\
            \cmidrule(r){2-3} \cmidrule(l){4-5}
             & RMSE & \#Face & RMSE & \#Face \\
            \midrule
            City3D + IDW  & 0.352 & 124.40 & 0.708 & 105.27 \\
            City3D + P.M. Diff & 0.577 & 89.04 & 3.016 & 97.47 \\
            \midrule
            City3D + Ours    & \textbf{0.244} & \textbf{82.72} & \textbf{0.534} & \textbf{80.12} \\
            \bottomrule
        \end{tabular}
        }
        \vspace{+0.5\baselineskip}
        \caption{Evaluation of point cloud preprocessors for City3D \cite{huang2022city3d} on \pdata. City3D tested with GT point cloud achieved 0.104 RMSE and 82.68 average faces.}
        \label{tab:point2planecity3d}
    \end{minipage}
    \hfill
    \begin{minipage}[t]{0.47\textwidth}
        \vspace{0pt}
        \centering
        \scriptsize
        \resizebox{\textwidth}{!}{
        \begin{tabular}{lccccc}
            \toprule
            Methods & s95 i30 & s95 i60 & s99 i30 & s99 i60 \\
            \midrule
            Linear  & 82.18 & 68.54 & 73.81 & 51.04 \\
            pNCNN & 68.97 & 66.32 & 64.80 & 63.94 \\
            CU-Net & 82.12 & 73.88 & 75.92 & 69.51 \\
            \midrule
            Ours (No-FP) & \textbf{92.14} & \textbf{81.83} & \textbf{83.59} & \textbf{75.15} \\
            \bottomrule
        \end{tabular}
        }
        \vspace{+0.5\baselineskip}
        \caption{
        Footprint predictions measured by IoU (\%).}
         \label{tab:footprint-iou}
    \end{minipage}
    \vspace{-1.5\baselineskip}
\end{table}

\textbf{(\nofp) \emph{Both} sparsity \emph{and} incompleteness.}
We compare to the state-of-the-art unguided depth completion methods, pNCNN \cite{eldesokey2020uncertainty} and CU-Net \cite{wang2022cu}, that we selected for their exclusive use of depth data and hence close relation to height completion tasks. 
These algorithms were trained on our \pdata, using identical data augmentation settings, but no footprint information was used. We adhered to the default hyperparameter configurations specified for each method. 
To ensure a fair comparison, we also trained \rdN{}, a variant of \rd{} that does not require footprints.
We tested reconstruction accuracy on  \pdata\ (s95 i30, s95 i60, s99 i30, and s99 i60) utilizing MAE and RMSE.
The noise injection settings mirror the experiment in \cref{tab:poznanquantitative}.
Since predicting both footprint and roof simultaneously is a more difficult problem,
the sparsity and incompleteness selection in \cref{tab:nfpoznanquantitative} is lower than in \cref{tab:poznanquantitative}.

\Cref{tab:nfpoznanquantitative} shows that \rd{} achieves the most accurate reconstruction
while pNCNN \cite{eldesokey2020uncertainty} fails to accurately predict the height values.
CU-Net \cite{wang2022cu} tends to 
over-smooth heights. 
Linear interpolation struggles to recognise noise and incomplete regions.

\textbf{(\nofp) \emph{Either} sparsity \emph{or} incompleteness.}
When  tested on the BuildingNet \cite{selvaraju2021buildingnet} dataset,
No-FP \rd\ outperforms linear interpolation, CU-Net \cite{wang2022cu}, and pNCNN \cite{eldesokey2020uncertainty} in scenarios with dominant sparsity (s98 i0, s99 i0) or incompleteness (s90 i60, s90 i80).
While CU-Net’s smoother predictions reduce large errors and improve RMSE, notably at s90~i80, s98~i0, and s99~i0 in \cref{tab:nfpoznanquantitative}, \mbox{CU-Net} does not predict sharp height maps well, as indicated by a higher MAE.

\subsection{Footprint Recovery}
\label{subsection:footprint-recovery}
No-FP \rd{} can help footprint recovery. Given a corrupted height map, we can predict the complete heights and infer a binary footprint by assigning a value of one to non-zero pixel values and zero otherwise. \cref{tab:footprint-iou} displays Intersection over Union (IoU) between ground truth and the predicted footprint. \rd{} yields the best the footprint recovery  on  the \pdata\ dataset compared to linear interpolation, CU-Net \cite{wang2022cu}, and pNCNN \cite{eldesokey2020uncertainty}.

\subsection{Enhancement of 3D Reconstruction}
\label{subsection:reconstruction}
This subsection demonstrates that using our model as the point cloud preprocessor can boost the accuracy of 3D building reconstruction. 
We preprocessed point clouds for City3D \cite{huang2022city3d}, a leading algorithm for converting point clouds to compact building meshes, by 
with 
our method, IDW \cite{shepard1968two}, and Perona-Malik Diffusion \cite{biasutti2019diffusion}. 
We report the average RMSE of the distance from each ground truth point to the face of the closest reconstructed mesh\footnote{We tested City3D without preprocessing but only a few meshes are reconstructed due to the severe sparsity.}. 
To further demonstrate the advantages of using clean geometric details from \rd{} in City3D \cite{huang2022city3d} reconstructions, we evaluate the minimum number of polygonal faces required for the output while maintaining geometry accuracy. Here, we dissolve all the edges with a dihedral angle of less than 5 degrees. A lower count, while maintaining a similar level of reconstruction accuracy, is preferable as it indicates a more compact representation.

As shown in \cref{tab:point2planecity3d}, 
City3D achieves the lowest point-to-plane distance when utilizing point clouds processed by \rd.
When using \rd{} as a preprocessor for City3D, the average minimum  face count is nearly identical 
to feeding ground truth point clouds to City3D. 

\Cref{fig:city3d} illustrates that using \rd{} as a preprocessing step significantly enhances the reconstruction quality in City3D on \pdata.
RoofDiffusion enhances gable details (\cref{fig:city3d-a}), complex geometry recovery (\cref{fig:city3d-b,fig:city3d-c}), and is robust to tree noise (\cref{fig:city3d-d}).

\subsection{Qualitative Evaluation of Height Completion}
\label{subsection:hmqualitative}

\begin{figure*}[t]
    \scriptsize
    \newcommand{\repeatdots}[1]{%
      \foreach \n in {1,...,#1} {$\cdot$}%
    }
    \newcolumntype{G}{@{\hskip 1.8pt}}
    \newcolumntype{T}{m{0.015\textwidth}}
    \newcolumntype{I}{>{\centering\arraybackslash}m{\SmallImageSize}}
    \centering

\resizebox{\textwidth}{!}{
    \begin{tabular}{G T G I G I G I G I G I G I G I G I G I}

        & \multicolumn{2}{c}{\repeatdots{9} AHN3 \repeatdots{9}} & \multicolumn{2}{c}{\repeatdots{7} Dales3D \repeatdots{7}} & \multicolumn{2}{c}{\repeatdots{3} Wayne County \repeatdots{3}} & \multicolumn{3}{c}{\repeatdots{14} Cambridge \repeatdots{14}} \\[-0.3ex]

        \rotatebox{90}{Input (3D)} & \smallImg{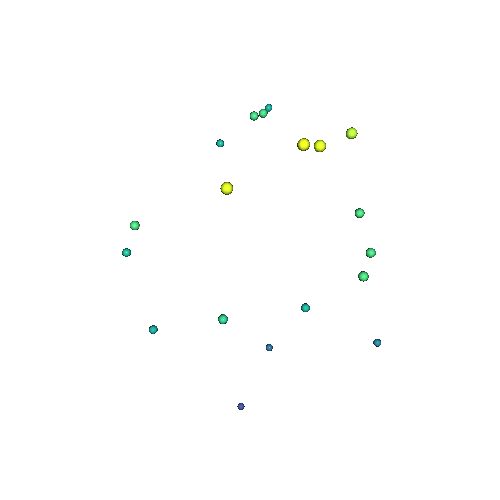} & \smallImg{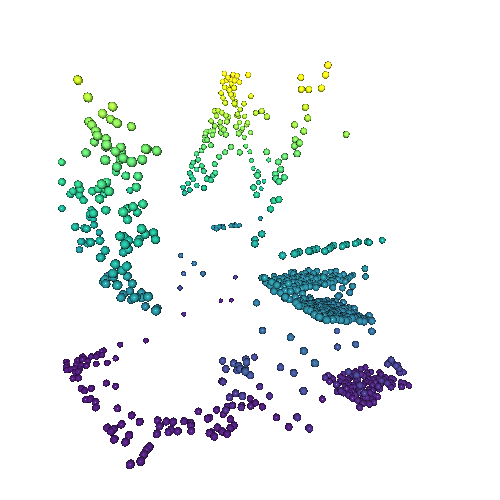} & \smallImg{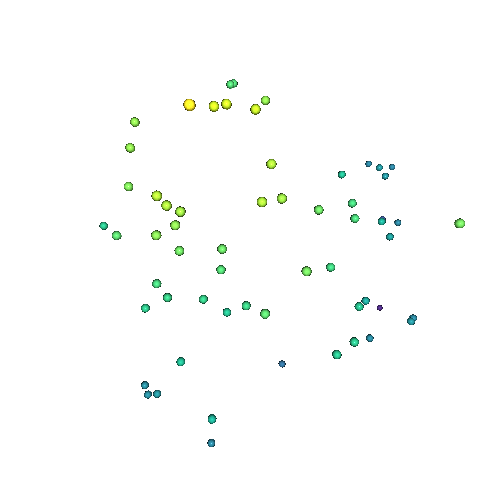} & \smallImg{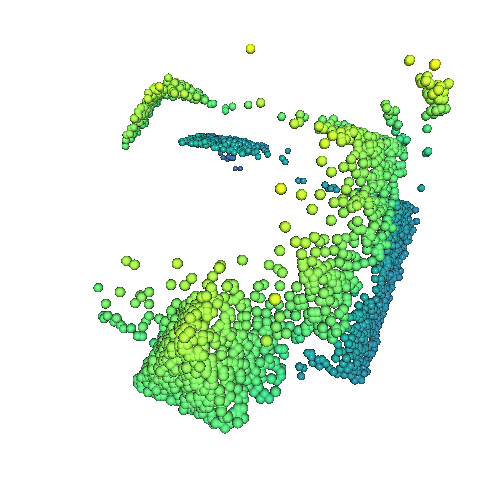} &
        \smallImg{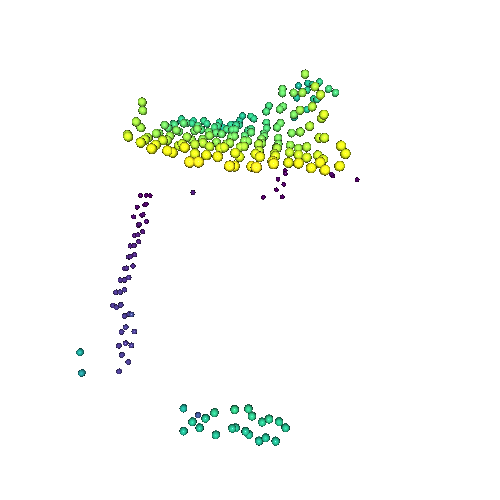} & \smallImg{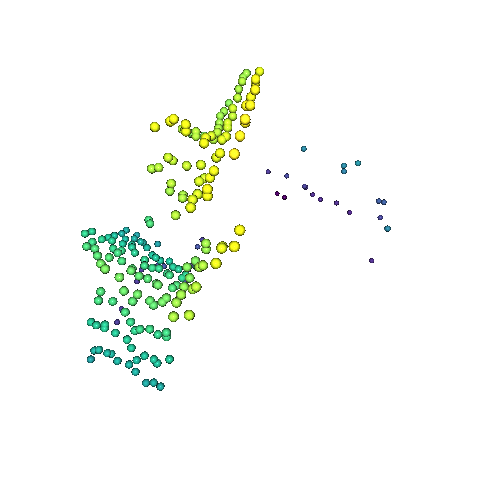} & 
        \smallImg{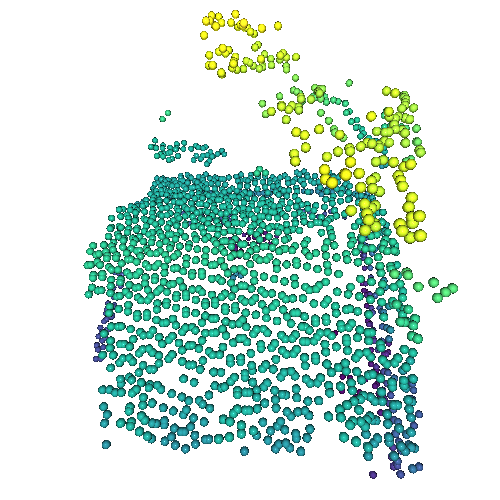} & \smallImg{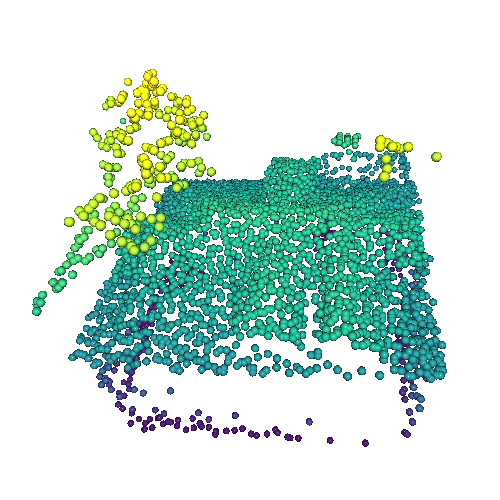} & \smallImg{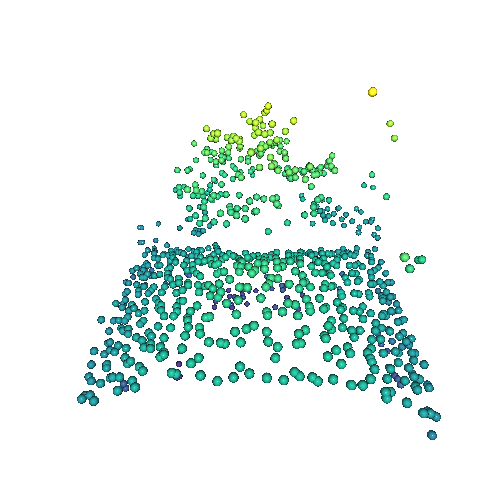} \\[-0.5ex]
    
        \rotatebox{90}{Input} & \smallImg{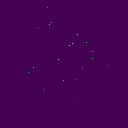} & \smallImg{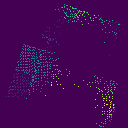} & \smallImg{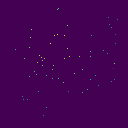} & \smallImg{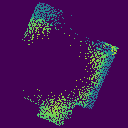} & \smallImg{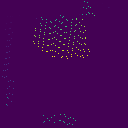} & \smallImg{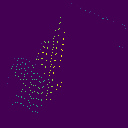} & 
        \smallImg{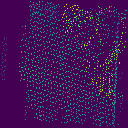} & \smallImg{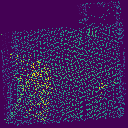} & \smallImg{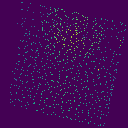} \\[-0.5ex]
        
        \rotatebox{90}{IDW} & \smallImg{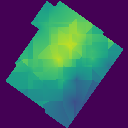} & \smallImg{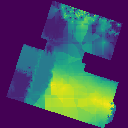} & \smallImg{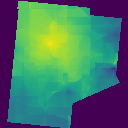} & \smallImg{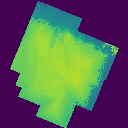} & 
        \smallImg{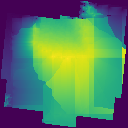} & \smallImg{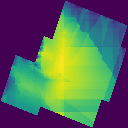} & \smallImg{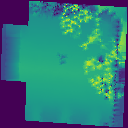} & \smallImg{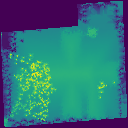} & \smallImg{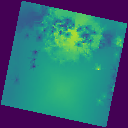} \\ [-0.5ex]  

        \rotatebox{90}{P.M. Diff.} & \smallImg{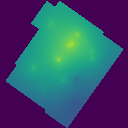} & \smallImg{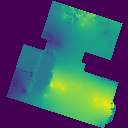} & \smallImg{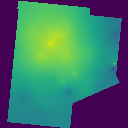} & \smallImg{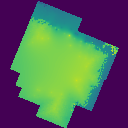} &  
        \smallImg{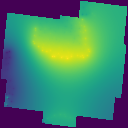} & \smallImg{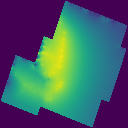} &
        \smallImg{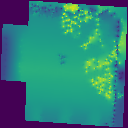} & \smallImg{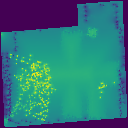} & \smallImg{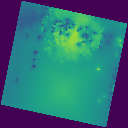} \\[-0.5ex]    

        \rotatebox{90}{Ours} & \smallImg{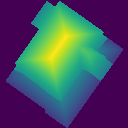} & \smallImg{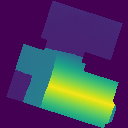} & \smallImg{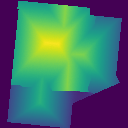} & \smallImg{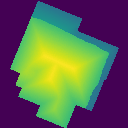} & 
        \smallImg{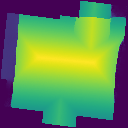} & \smallImg{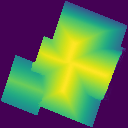} & \smallImg{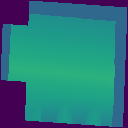} & \smallImg{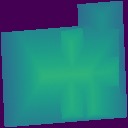} & \smallImg{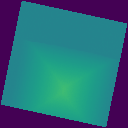} \\[-1ex]   

        \rotatebox{90}{Ours (3D)} & \smallSubImg{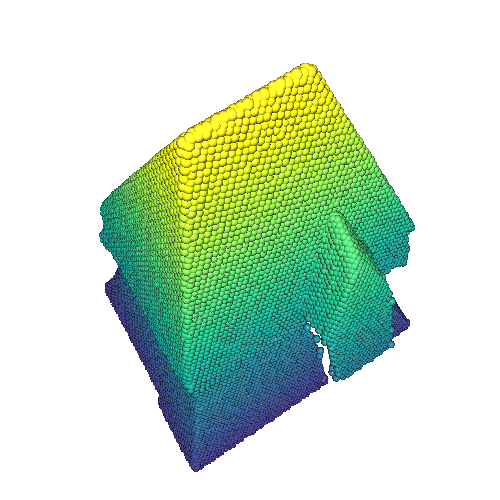}{}{fig:qualitative-a} & \smallSubImg{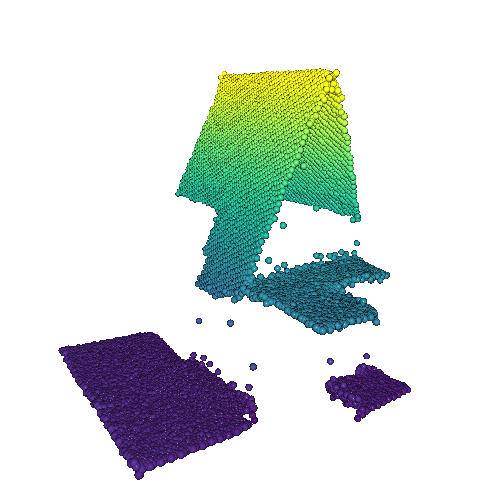}{}{fig:qualitative-b} & \smallSubImg{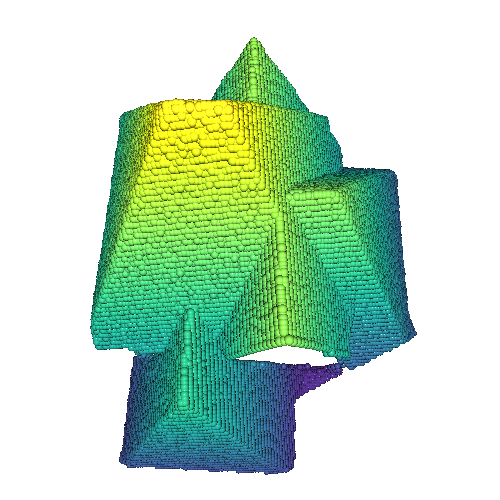}{}{fig:qualitative-c} & \smallSubImg{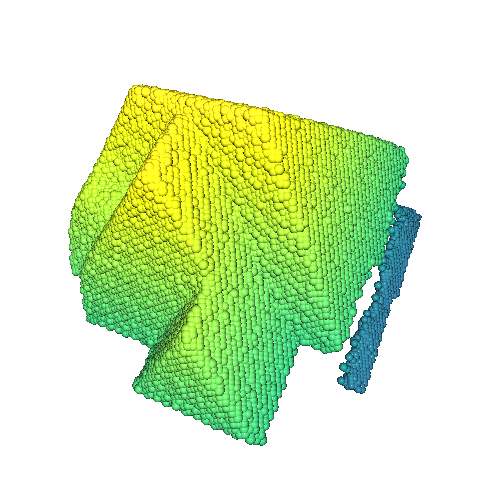}{}{fig:qualitative-d} &
        \smallSubImg{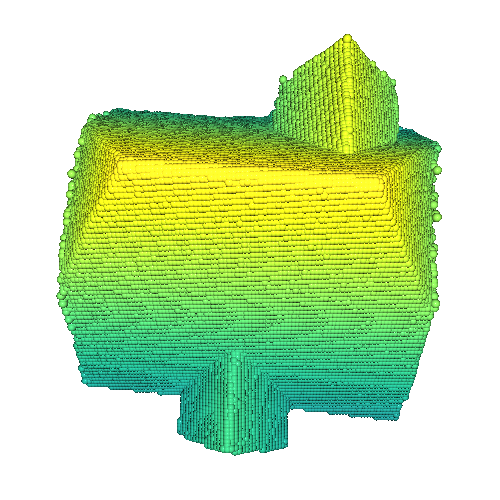}{}{fig:qualitative-e} & \smallSubImg{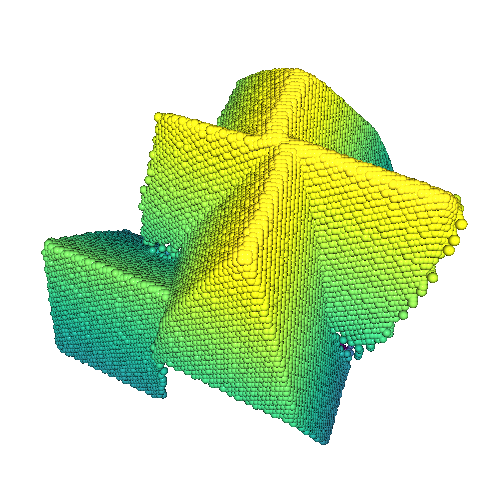}{}{fig:qualitative-f} & 
        \smallSubImg{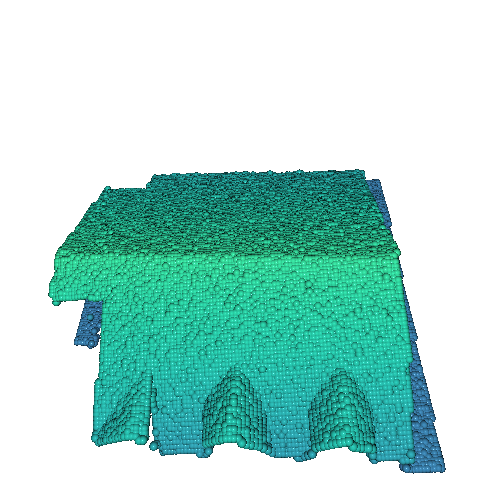}{}{fig:qualitative-g} & \smallSubImg{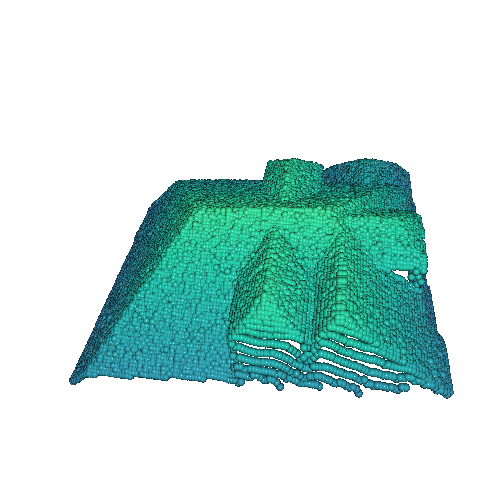}{}{fig:qualitative-h} & \smallSubImg{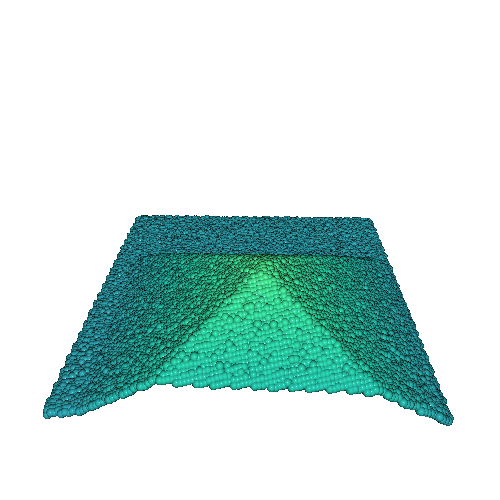}{}{fig:qualitative-i} \\ 
        
    \end{tabular}
}
    \vspace{-0.8\baselineskip} 
    \caption{Evaluation of the completion and denoising of real-world scans.}
    \label{fig:experimentResults}
    \vspace{-1.0\baselineskip}
\end{figure*}
To illustrate the capability of handling the gap between synthetic and real-world scanning conditions, we evaluated our method on datasets featuring real-world LiDAR scans. Due to the absence of ground truth, we adopt a visual comparison. Importantly, to maintain the real-world properties of the scans, we refrain from injecting any additional noise including global, outlier, and tree noise.

\textbf{Real-world Noise.}
We artificially remove points from the rasterized height maps to evaluate the resilience of our model to such corruptions, even in the presence of real-world noise and unseen roof geometries in both AHN3 \cite{huang2022city3d} and Dales3D \cite{varney2020dales}.
The 99\% sparsity (\cref{fig:qualitative-a,fig:qualitative-c}) and 70\% incompleteness (\cref{fig:qualitative-b,fig:qualitative-d}) are synthesized.
We observe that IDW \cite{shepard1968two} produces a mosaic effect, particularly in incomplete regions, and fails to capture the true geometry of roofs. 
Meanwhile, Perona-Malik Diffusion \cite{biasutti2019diffusion} yields smoother outcomes but tends to overlook sharp roof details.
In contrast, \rd{} effectively yields clean and faithful geometric reconstructions.

\begin{figure}[t]
    \newcolumntype{G}{@{\hskip 3pt}}
    \newcolumntype{T}{m{0.015\textwidth}}
    \newcolumntype{I}{>{\centering\arraybackslash}m{\LargeImageSize}}
    \begin{minipage}[t]{0.46\textwidth}
        \vspace{0pt}
        \centering
        \resizebox{\textwidth}{!}{
        \begin{tabular}{G T G I G I G I G I}
            \tiny\rotatebox{90}{Input} & \largeImg{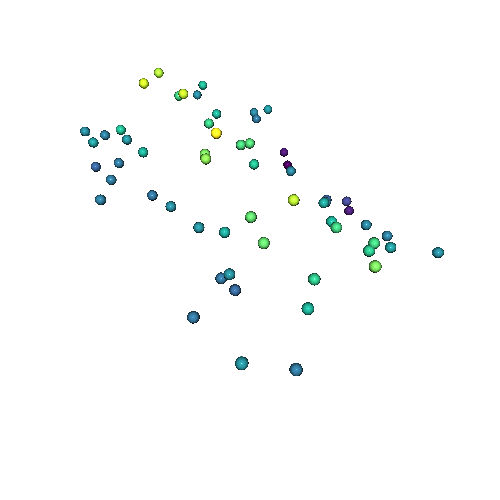} & \largeImg{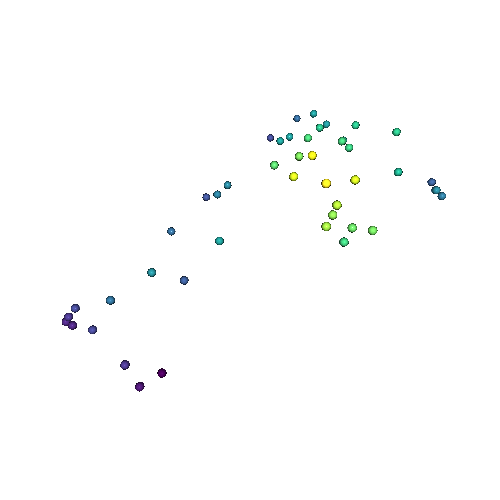} & \largeImg{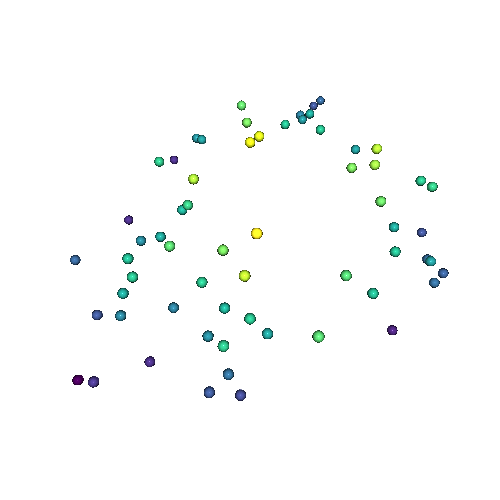} & \largeImg{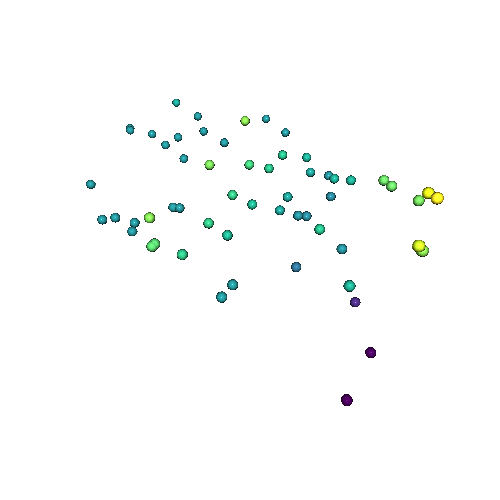} \\
        
            \tiny\rotatebox{90}{IDW} & \largeImg{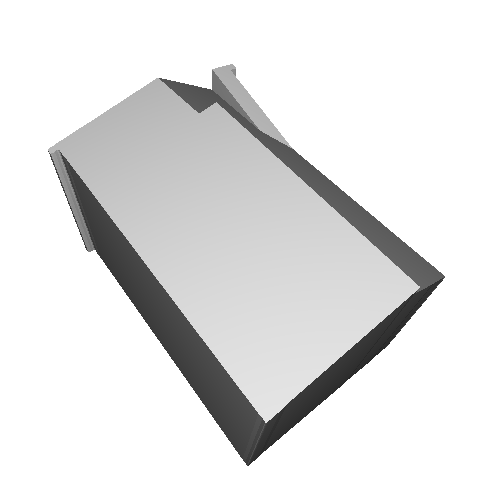} & \largeImg{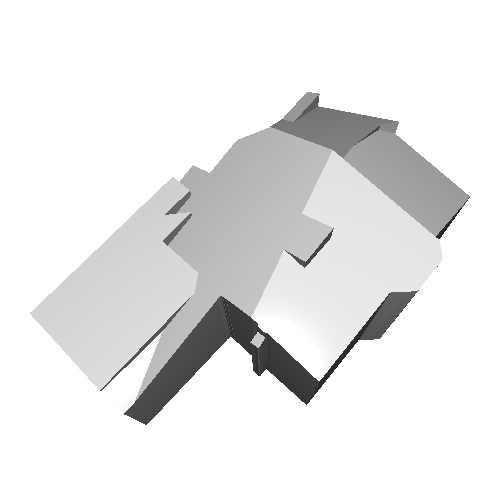} & \largeImg{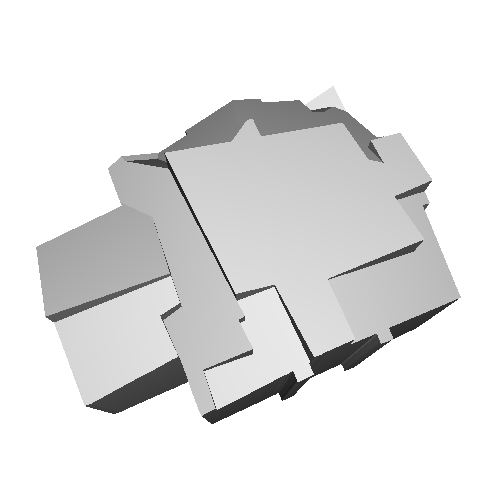} & \largeImg{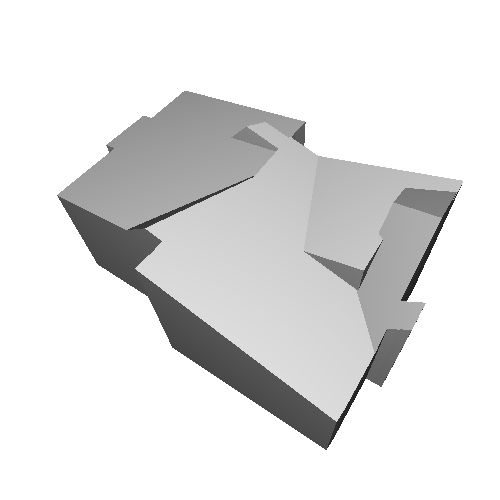} \\
            
            \tiny\rotatebox{90}{P.M. Diff.} & \largeImg{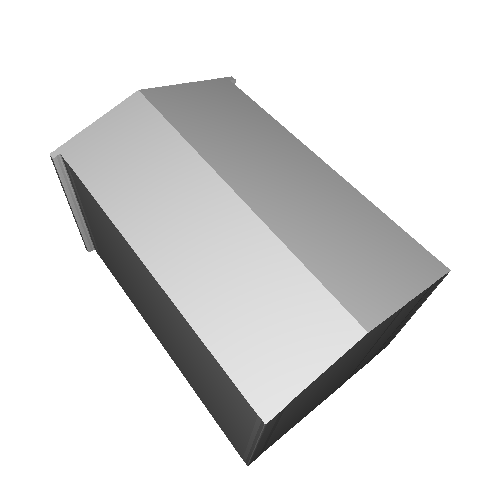} & \largeImg{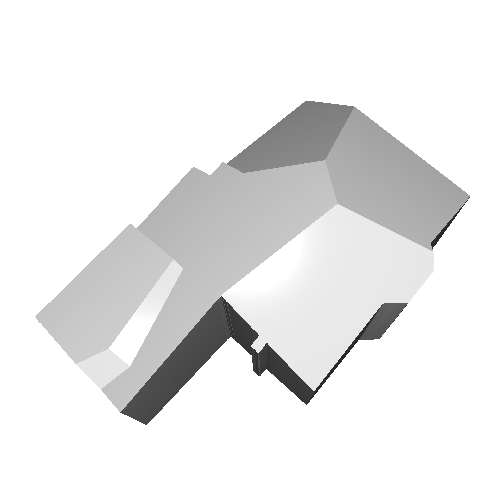} & \largeImg{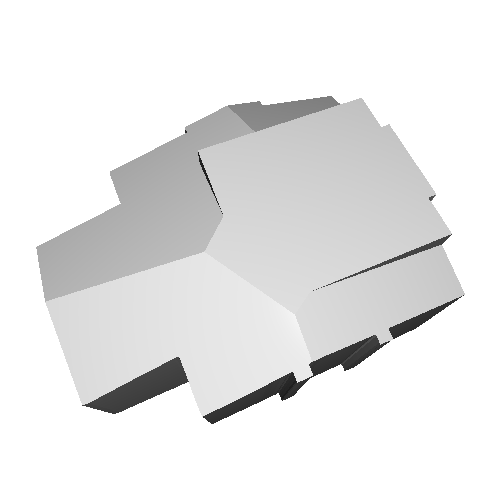} & \largeImg{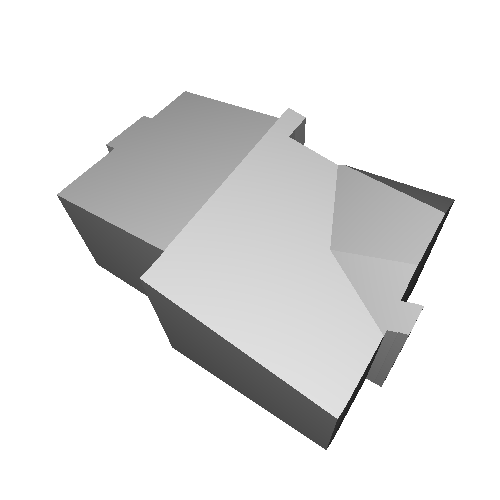} \\
            
            \tiny\rotatebox{90}{Ours} & \largeImg{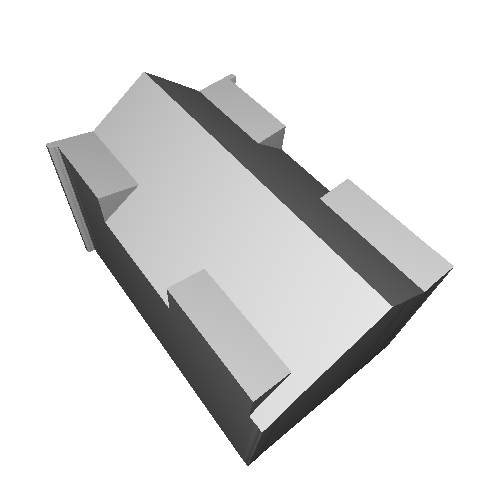} & \largeImg{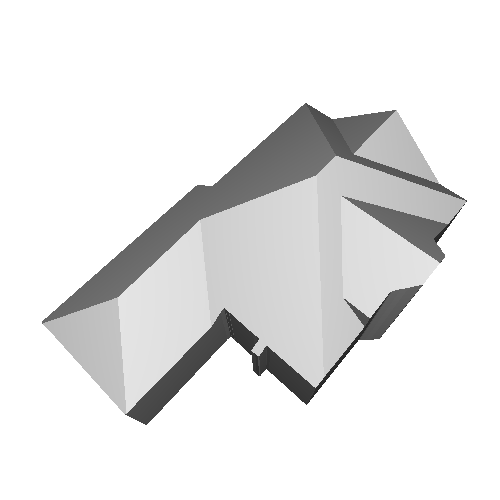} & \largeImg{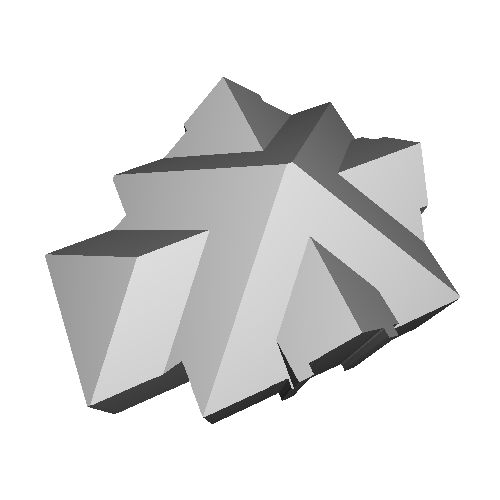} & \largeImg{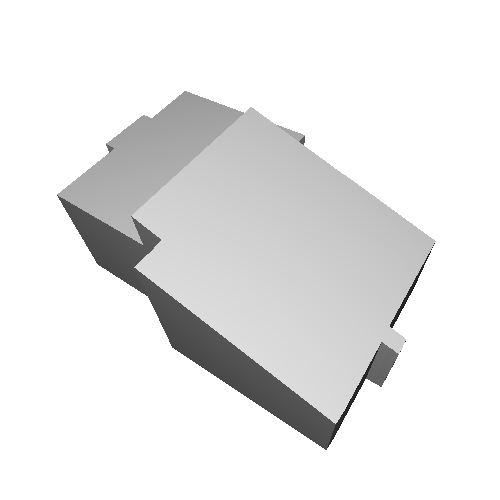} \\

            \tiny\rotatebox{90}{GT} & \largeImg{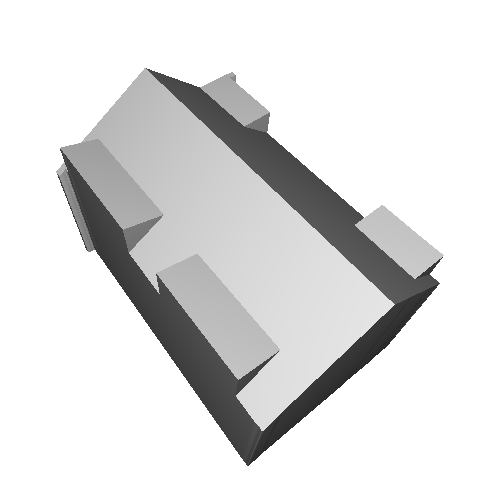} & \largeImg{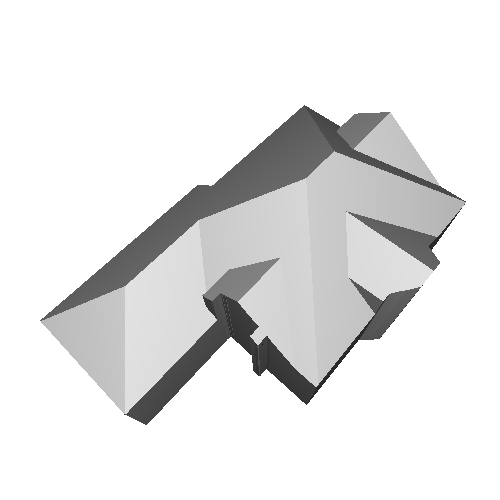} & \largeImg{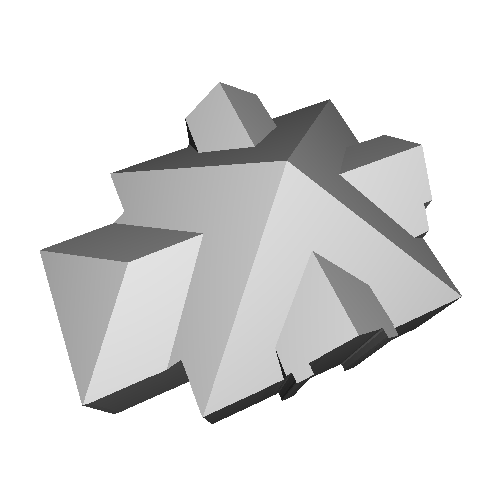}& \largeImg{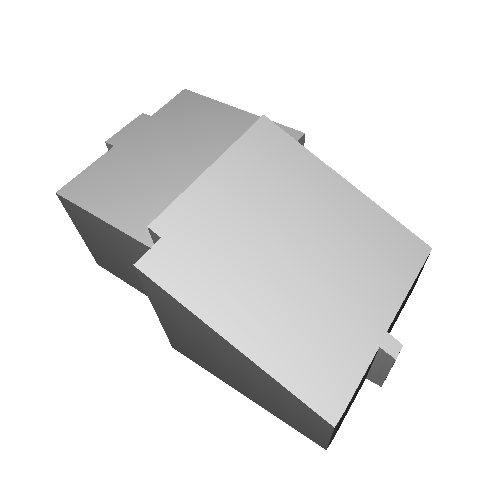} \\[-1.5ex]

            & \subcaption{\label{fig:city3d-a}} & \subcaption{\label{fig:city3d-b}} & \subcaption{\label{fig:city3d-c}} & \subcaption{\label{fig:city3d-d}}

        \end{tabular}
        }
        \vspace{-2.0\baselineskip}
        \caption{3D reconstruction 
        using different height map pre-processors.
        \rd{} matches the ground truth (GT) meshes generated by feeding GT height maps into City3D \cite{huang2022city3d}, except for (b).
        In (b)
        \rd{} omits a dormer present in GT
        due to the lack of data points in that area.}
        \label{fig:city3d}
    \end{minipage}
    \hfill
    \begin{minipage}[t]{0.5\textwidth}
        \vspace{0pt}

        \resizebox{\textwidth}{!}{
        \begin{tabular}{T G I G I G I G I}
    
            \tiny\rotatebox{90}{Input (3D)} & \largeImg{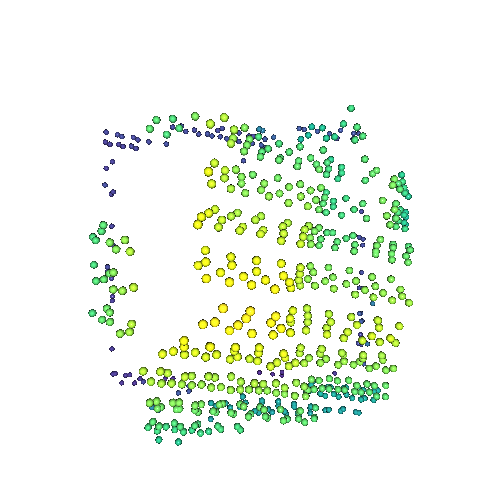} & \largeImg{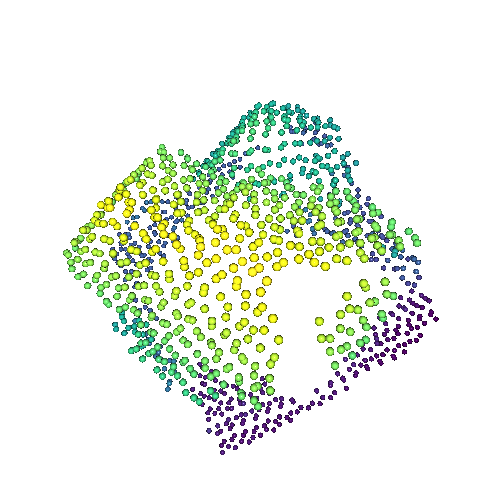} & 
            \largeImg{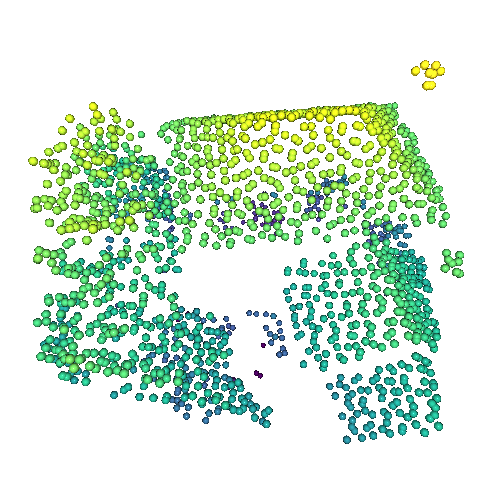} & \largeImg{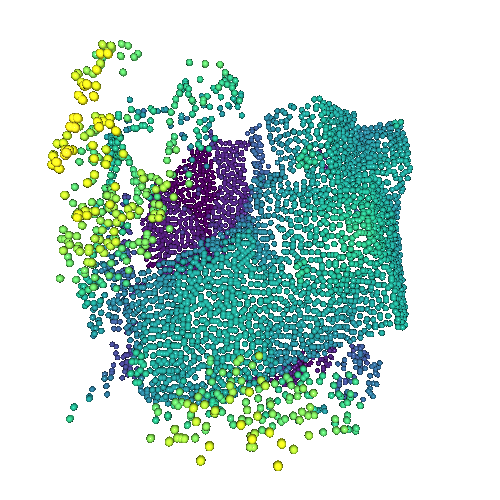} \\
        
            \tiny\rotatebox{90}{Input} & \largeImg{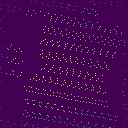} & \largeImg{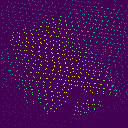} & 
            \largeImg{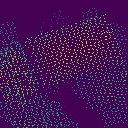} & \largeImg{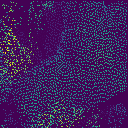} \\
            
            \tiny\rotatebox{90}{pNCNN} & 
            \largeImg{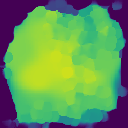} & \largeImg{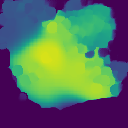} & \largeImg{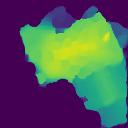} & \largeImg{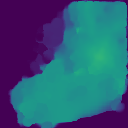} \\
    
            \tiny\rotatebox{90}{CU-Net} &
            \largeImg{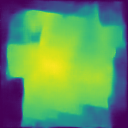} & \largeImg{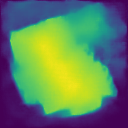} &
            \largeImg{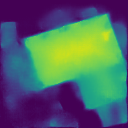} & \largeImg{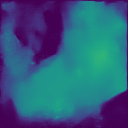} \\
    
            \tiny\rotatebox{90}{Ours} &
            \largeImg{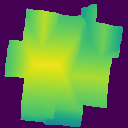} & \largeImg{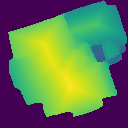} & \largeImg{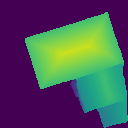} & \largeImg{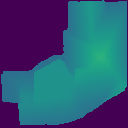} \\  

            \tiny\rotatebox{90}{Ours (3D)} &  \largeImg{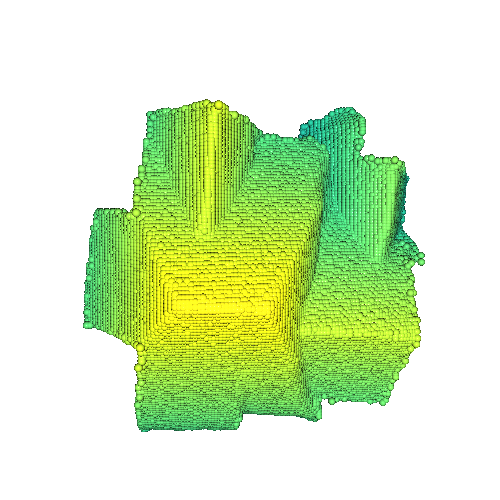} & \largeImg{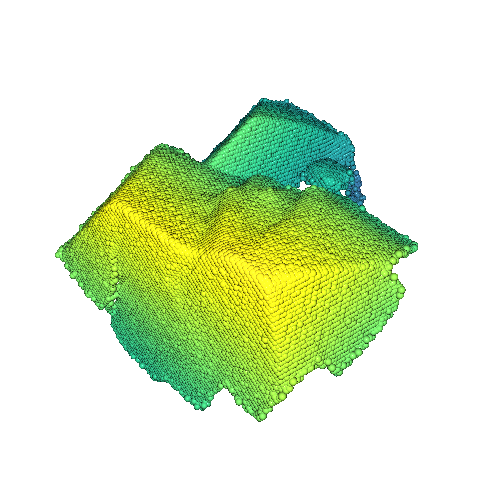} & 
            \largeImg{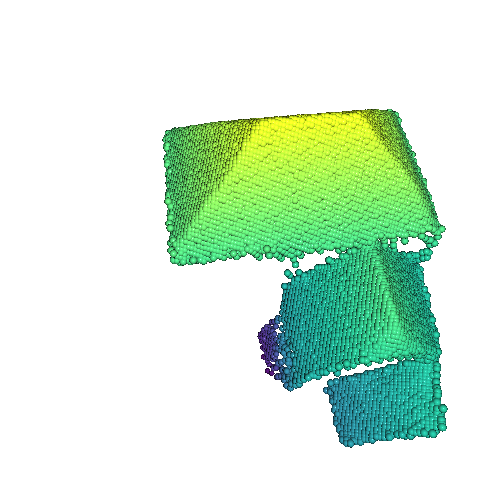} & \largeImg{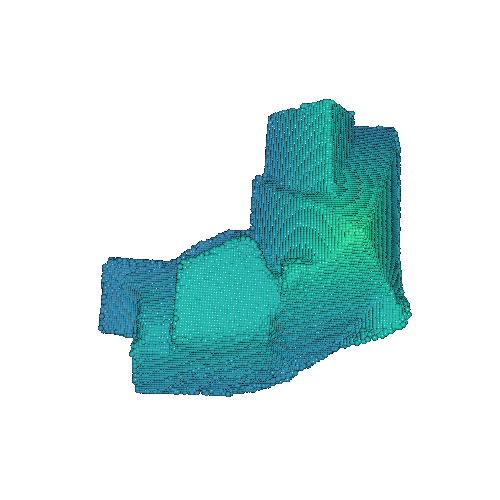} \\
            [-1.5ex]
            & \subcaption{\label{fig:nf-qualitative-a}} & \subcaption{\label{fig:nf-qualitative-b}} & \subcaption{\label{fig:nf-qualitative-c}} & \subcaption{\label{fig:nf-qualitative-d}}

        \end{tabular}
        }

        \vspace{-2.0\baselineskip}

        \caption{Evaluation of the completion and denoising in real-world scans without footprints.}
        \label{fig:nfqualitative}

    \end{minipage}
    \vspace{-1.5\baselineskip}
\end{figure}

\textbf{Real-world Incompleteness.}
We collect point clouds exhibiting incompleteness from real-world scans, sourced from the Wayne County dataset \cite{wayneco2017}. Notably, the point cloud is rated at QL2 \cite{ql2023}, which is the most common quality of LiDAR scans across the United States. 
\Cref{fig:qualitative-e,fig:qualitative-f} show that \rd{} effectively leverages available points in conjunction with footprint information to reconstruct missing parts, even when these parts are distant from the existing points. 
In contrast, IDW \cite{shepard1968two} and Perona-Malik Diffusion \cite{biasutti2019diffusion} failed in such scenarios due to a lack of prior information about roof structures.
\Cref{fig:nf-qualitative-a,fig:nf-qualitative-b} also shown \rdN{} can recover incompleteness with cleaner and sharper features.

\textbf{Real-world Tree Noise.}
We also gather point clouds containing tree noise from the Cambridge dataset \cite{cambridge2021}.
As illustrated in \cref{fig:qualitative-g,fig:qualitative-h,fig:qualitative-i}, \rd{} demonstrates its efficacy in restoring roof geometry precisely while eliminating the tree points. 
Moreover, \cref{fig:nf-qualitative-c,fig:nf-qualitative-d} highlight that \rdN{} can effectively extract and recover roof geometry from overlapping tree noise.

\subsection{Limitations}
\label{subsection:limitations}
We selected data containing tree noise from USGS 3DEP LiDAR sampled over Wayne County, MI \cite{wayneco2017}, and Cambridge, MA \cite{cambridge2021}, for analysis. 
In approximately 36\% of these cases, tree noise was mistakenly reconstructed as building structures, see examples in the supplement.
This error was due to the resemblance of tree canopies to architectural elements, such as dormers or chimneys, or due to severe occlusion by tree noise obscuring the underlying structures. 
Footprint-guided \rd{} can encounter challenges due to severe misalignment between the footprint and the height map. This problem also occurs in \cite{huang2022city3d}. Incorporating misalignment into the data augmentation pipeline and relaxing the footprint mask, $\footprint$, during the denoising process warrants further investigation.
Moreover, factors such as floor space, height from the ground, and location are useful indicators of roof geometry but are not yet used in the model. 
We leave the above as future work. 
\section{Conclusions}
\vspace{-0.8\baselineskip}
We introduced \rd{}, a diffusion model for roof height map repair.
\rd{} is capable of repairing extreme sparsity, incompleteness, and noise. 
Additionally, we unveiled a comprehensive roof dataset containing over 13\KK\ complex geometric roof structures, with complete and a clean ground truth mesh and height map.
This dataset can serve as a valuable asset for future long-tail research in remote sensing. 
To approximate real-world conditions, we also introduced methods for synthesizing tree noise and incomplete shapes. 
By generating intentionally-corrupted height maps from ground truths, these techniques not only mitigate the absence of ground truth in real-world scans but also facilitate data augmentation.
They allow for benchmark customization with varying levels of height map corruption.
Our experiments demonstrate the robustness of \rd{}
 across datasets with real \cite{varney2020dales, huang2022city3d, wayneco2017, cambridge2021} and synthetic scans \cite{selvaraju2021buildingnet}, under diverse corruption conditions. 
\rd{}
outperforms both non-learning \cite{keys1981cubic,shepard1968two,biasutti2019diffusion} and learning-based methods \cite{wang2022cu, eldesokey2020uncertainty} and significantly improves the accuracy of the 3D building reconstruction algorithm \cite{huang2022city3d}.

\section*{Acknowledgement}
We thank Oleksandr Moskalenko and Dr. Zhe Jiang for generously providing computational resources on UF HyperGator. 
We also extend our appreciation to our colleagues at Meta Platforms for their assistance in processing data and offering valuable feedback throughout this project.
We acknowledge the map services and data available from the U.S. Geological Survey, National Geospatial Program, and the Spatial Information System of the City of Poznań, Poland.
Additionally, we are grateful for the funding provided by the UF Graduate School Preeminence Award and the UF Informatics Institute Fellowship.

\bibliographystyle{splncs04}
\bibliography{main}

\title{Supplementary Material for RoofDiffusion}

\titlerunning{RoofDiffusion Supplementary}

\author{Kyle Shih-Huang Lo\inst{1} \and
Jörg Peters \inst{1} \and
Eric Spellman\inst{2}}

\authorrunning{K.~Lo et al.}

\institute{University of Florida, Gainesville FL 32611, USA
\\
\and
Meta Platforms, Inc.\\
\email{kyleshihhuanglo@ufl.edu, jorg@cise.ufl.edu, espellman@meta.com}}

\maketitle

This supplement contains:
\begin{itemize}
    \item Selection of Hyper-parameters (\cref{sec:hyper-parameters})
    \begin{itemize}
        \item Roof Height Normalization (\cref{subsec:normalization})
        \item Notations (\cref{subsec:notations})
    \end{itemize}
    \item PoznanRD Dataset (\cref{sec:pdataset})
    \begin{itemize}
        \item Data Balancing (\cref{subsec:data-balancing})
        \item Synthesizing Incompleteness (\cref{subsec:incompletedetail})
        \item Synthesizing Tree Noise (\cref{subsec:synthesizing-tree-noise})
        \item Comparison to Existing Datasets (\cref{subsec:compare-dataset})
    \end{itemize}
    \item Experiments (\cref{sec:experiments})
    \begin{itemize}
        \item Implementation Details (\cref{subsec:implementation-details})
        \item Ablation of Tree Noise Augmentation (\cref{subsec:tree-ablation})
        \item Sampling Step Analysis (\cref{subsec:step})
        \item LiDAR Scan Patterns (\cref{subsec:lidar-pattern})
        \item Variety of the Roofs (\cref{subsec:roof-variety})
        \item How Prone is the Model to Hallucination based on Footprints? (\cref{subsec:heightmap-availability})
        \item Failure Cases (\cref{subsec:failure-cases})
        \item Depth Completion on the KITTI Dataset (\cref{subsec:kitti})
        \item Additional Qualitative Experimental Results (\cref{subsec:additional-qualitative})
    \end{itemize}
\end{itemize}

\newpage

\section{Selection of Hyper-parameters}
\label{sec:hyper-parameters}
\subsection{Roof Height Normalization}
\label{subsec:normalization}
\begin{figure}[h]
    \centering
    \includegraphics[width=0.5\linewidth]{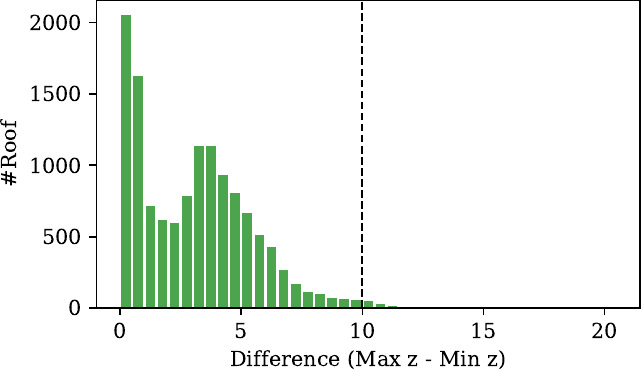}
    \vspace{-0.7\baselineskip}
    \caption{Distribution of height differences and roof counts.}
    \label{fig:zdiff}
    \vspace{-1.5\baselineskip}
\end{figure}

Our choice of 10-meters as the upper bound on
the difference between maximum minus minimum roof height,
for mapping into the range of $\left[-1,1\right]$,
is justified by the distribution of roof height differences in the \pdata, see  \cref{fig:zdiff}:
99\% of the height differences are less or equal to
10 meters, marked as the black dash line.
For the buildings with height differences exceeding 10 meters, we scale them to 10 meters and record the scaling factor for recovering the true height.
\subsection{Notations}
\label{subsec:notations}
As in the main document, in this supplement, ``s'' represents Sparsity (\%), the ratio of randomly removed to total pixels in the footprint. 
Furthermore, ``i'' denotes the Incompleteness ratio (\%), the proportion of pixels removed due to incompleteness in the footprint. 
Lastly, ``t'' signifies the Tree count used to synthesize tree noise.

\section{PoznanRD Dataset}
\label{sec:pdataset}

\subsection{Data Balancing}
\label{subsec:data-balancing}
We started with collecting 16\KK\ compact and high-detail LoD 2.2 \cite{biljecki2016improved} roof meshes from the city of Poznan in Poland \cite{poznan3d}. 
To match our focus on complex roof structures, we rebalanced the dataset by excluding 3\KK\ flat roofs. 
This resulted in our \pdata{} comprising 13\KK\ buildings. 
The original longitude, latitude, height, and scaling for each roof are recorded in a Comma-Separated Values (CSV) file for easy access.

\newpage

\begin{figure}[t]
    \centering
    \begin{minipage}{0.8\linewidth}
        \begin{algorithm}[H]
            \caption{Incompleteness Mask (Training)}
            \begin{algorithmic}[1]
                \State \textbf{set} $\incompleteMask$ \textbf{to} \Call{zeros}{$\imageH$, $\imageW$}
                \For {$g = 1 \ldots G$} \Comment{Index of Gauss distribution.}
                    \For {$i = 1 \ldots H$} \Comment{Image height.}
                        \For {$j = 1 \ldots W$} \Comment{Image width.}
                            \State $p \sim \mathcal{U}(0,1)$
                            \If{$p_{\text{Gauss}, g}(i,j) > p$}
                                \State $\incompleteMask(i,j) = 1$
                            \EndIf
                        \EndFor
                    \EndFor
                \EndFor
                \State $\incompleteMask = \footprint \odot \incompleteMask$ \Comment{Set outside footprint to zero.}
            \end{algorithmic}
            \label{algo:incompletetraining}
        \end{algorithm}
        \vspace{-3.0\baselineskip}
    \end{minipage}
\end{figure}

\begin{figure}
    \centering
    \begin{minipage}{0.8\linewidth}
        \begin{algorithm}[H]
            \caption{Incompleteness Mask (Benchmark)}
            \begin{algorithmic}[1]
                \State \textbf{set} $\incompleteMask$ \textbf{to} \Call{zeros}{$\imageH$, $\imageW$}
                \While {\Call{sum}{M} $< N_{\text{thres}}$}
                        \State $\left(i, j\right) \sim \gaussMix(i, j)$
                        \If {$\footprint(i, j) == 1$} \Comment{Should be within footprint.}
                            \State $\incompleteMask(i, j) = 1$
                        \EndIf
                \EndWhile
            \end{algorithmic}
            \label{algo:incompletebenchmark}
        \end{algorithm}
        \vspace{-3.0\baselineskip}
    \end{minipage}
\end{figure}

\subsection{Synthesizing Incompleteness}
\label{subsec:incompletedetail}
We show the detail of generating a binary mask for synthesizing incompleteness, denoted as $\incompleteMask$. In this mask, a value of 1 indicates the pixels that will be removed, thereby yielding incompleteness, otherwise, 0.
First, we employ a Gaussian Mixture Model (GMM) \cite{reynolds2009gaussian} to establish a probability density function, 
\begin{gather}
    \gaussMix(i, j) := \sum_{g=1}^G \frac{1}{G} \ \underbrace{e^{-\frac{\|(i, j) - \gaussMean\|^2}{2\boldsymbol{\sigma}_g^2}}}_{p_{\text{Gauss}, g}},
\end{gather}
where $i$ and $j$ are the pixel coordinate in $x$ and $y$ direction of right-handed coordinate system, respectively, and
\begin{gather}
    \mu_{g,x} \sim \mathcal{U}(1, \imageH)  \quad \mu_{g,y} \sim \mathcal{U}(1, \imageW),
    \\
    \sigma_{g,x} \sim \mathcal{U}(\varMin, \varMax), \quad \sigma_{g,y} \sim \mathcal{U}(\varMin, \varMax),
    \\
    \gaussMean = \left(\mu_{g,x}, \mu_{g,y}\right),
    \\
    \boldsymbol{\sigma}_g = \left(\sigma_{g,x}, \sigma_{g,y}\right),
\end{gather}
and $G$ denotes the number of Gaussian distributions utilized in the construction of the GMM. 
Furthermore, $\mu_{g,x}$ and $\mu_{g,y}$ represent the mean in $x$ and $y$ direction, respectively, for \mbox{$g$-th} Gaussian distribution in the model. 
Similarly, $\sigma_{g,x}$ and $\sigma_{g,y}$ represent the standard deviation in $x$ and $y$ direction, respectively, for \mbox{$g$-th} Gaussian distribution. 
Also, $\varMin$,  $\varMax$ are hyper-parameters. 
$ \imageH$ and $\imageW$ are the height and width of the image. 
Note that a greater value of $G$ can yield a more intricate GMM, which, in turn, can synthesize a more complex shape of incompleteness.

During training, there is no requirement to generate an incomplete mask with a precise number of missing points. Therefore, for more efficient mask generation within the dataloader, we propose a method outlined in \cref{algo:incompletetraining}.

For benchmarking, we can sample an incompleteness mask, $\incompleteMask$, with a specific number of pixels to be removed, $N_{\text{thres}}$, using \cref{algo:incompletebenchmark}.

\begin{figure}[t]
    \newcolumntype{G}{@{\hskip 0.007\columnwidth}}
    \newcolumntype{I}{>{\centering\arraybackslash}m{\TinyImageSize}}
    \setlength\tabcolsep{0pt}
    \centering
    \resizebox{0.7\textwidth}{!}{
    \begin{tabular}{IGIGIGIGIGI}
        \tinyImg{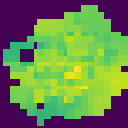} & \tinyImg{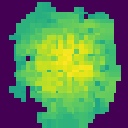} & \tinyImg{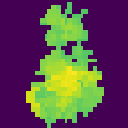} & \tinyImg{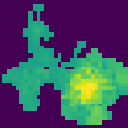} & \tinyImg{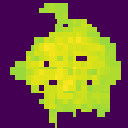} & \tinyImg{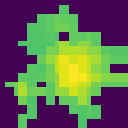} \\
    \end{tabular}
    }

    \vspace{-0.7\baselineskip}
    
    \caption{Examples of tree height map in \pdata dataset. (The examples are color mapped and resized for clear visualization. The height increases as the color becomes more yellow.)}
    \label{fig:treeHeightMap}
\end{figure}
\begin{figure}[t]
    \centering
    \begin{minipage}{\linewidth}
        \begin{algorithm}[H]
            \caption{Tree Noise Injection} 
            \begin{algorithmic}[1]
                \State \textbf{Given} $\gtImage \sim q( \gtImage )$ \textbf{and} $\treeImage \sim q( \treeImage )$
                \State $\treeCount \sim \mathcal{U}_\mathbb{Z}( \treeCountMin, \treeCountMax)$
                \For {$k = 1 \ldots N_{\text{tree}}$}
                    \State $d \sim \mathcal{U}(1, 360)$ \Comment{Tree rotation degree.}
                    \State $\xyScale \sim \mathcal{U}(\xyScaleMin, \xyScaleMax)$ \Comment{Tree coverage area scaler.}
                    \State $\zScale \sim \mathcal{U}(\zScaleMin, \zScaleMax)$ \Comment{Tree height scaler.}
                    \State $\treeImage$ = \Call{rotate}{$\treeImage$, $d$}
                    \State $\treeImage$ = \Call{resize}{$\treeImage$, $\xyScale$}
                    \State $\treeImage = \zScale\treeImage$
                    \Repeat
                        \State $\left(i, j\right) \sim \{(i, j) \mid \footprint(i, j) = 0\}$ \Comment{Sample coordinates outside of footprint.}
                        \State \textbf{set} $\treeImage'$ \textbf{to} \Call{zeros}{$\imageH$, $\imageW$}
                        \State \textbf{merge} $\treeImage$ \textbf{into} $\treeImage'$ \textbf{at center} \((i, j)\)
                        \State $\condImage$, $c_{\text{replaced}}$ = \Call{max}{$\gtImage$, $\treeImage'$} \Comment{Get occluded height map and pixel count.}
                    \Until{$c_{\text{replaced}} > 0$} \Comment{Enforce tree occlusion.}
                \EndFor
            \end{algorithmic}
            \label{algo:tree}
        \end{algorithm}
    \end{minipage}
\end{figure}

\subsection{Synthesizing Tree Noise}
\label{subsec:synthesizing-tree-noise}

\Cref{fig:treeHeightMap} shows examples of tree height maps in our dataset.
\cref{algo:tree} outlines injecting tree noise into ground truth height maps, $\gtImage$, to generate a corrupted height map, $\condImage$.

\newpage

\begin{table}[h]
    \newcolumntype{g}{@{\hskip 2.8pt}}
    \centering
    \resizebox{\textwidth}{!}{
    \small
    \begin{tabular}{lcccccc}
        \toprule
        Dataset       & Roof Variety & Point Cloud & Mesh Property & Roof Only & \#Roof (\KK) \\
        \midrule
        RoofGAN \cite{qian2021roof}          & Limit & -                 & Compact (Noise Free)   & \checkmark & 0.5   \\
        Ren \etal \cite{ren2021intuitive}    & Limit & -                 & Compact (Noise Free)   & \checkmark & 3 \\
        RoofN3D \cite{wichmann2018roofn3d}   & Limit & Real-world Scan   & Compact (Noise Free)   & \checkmark & 118 \\
        UrbanScene3D \cite{lin2022capturing} & High  & Real-world Scan   & Dense (Reconstructed)  &   & -   \\
        STPLS3D \cite{chen2022stpls3d}       & High  & Real-world Scan   & Dense (Reconstructed)  &   & -   \\
        City3D  \cite{huang2022city3d}       & High  & Real-world Scan   & Coarse (Reconstructed) &   & 20  \\
        Building3D \cite{wang2023building3d} & High  & Real-world Scan   & Coarse (Reconstructed) & \checkmark & 160 \\
        BuildingNet \cite{selvaraju2021buildingnet} & High  & Sample from Mesh  & Compact (Noise Free)   &  & 2 \\
        \midrule
        PoznanRD (Ours)  & High  & Sample from Mesh  & Compact (Noise Free)   & \checkmark & 13  \\
        \bottomrule
    \end{tabular}
    }

    \vspace{0.5\baselineskip}
    
    \caption{Comparison of Building Datasets.}
    \label{tab:buildingdataset}
\end{table}

\subsection{Comparison to Existing Datasets}
\label{subsec:compare-dataset}
\Cref{tab:buildingdataset} provides the comparison between our \pdata{} and the existing building or roof datasets.
RoofGAN \cite{qian2021roof}, RoofN3D \cite{wichmann2018roofn3d}, and the dataset \cite{ren2021intuitive} provide compact, noise-free roof meshes but offer only a limited range of roof types. 
Specifically, RoofGAN \cite{qian2021roof} encompasses 0.5\KK\ roofs, constructed with 2 to 5 hip roof primitives.
The dataset \cite{ren2021intuitive} presents 3\KK\ more complex meshes compared to RoofGAN \cite{qian2021roof}, yet it primarily features buildings constructed with hip-based primitives.
Moreover, most roofs in \cite{ren2021intuitive} lack detailed structures like dormers, which are central to our focus. 
RoofN3D \cite{wichmann2018roofn3d} is confined to pyramid, saddleback, and two-sided hip roofs.

UrbanScene3D \cite{lin2022capturing} and STPLS3D \cite{chen2022stpls3d} offer a diverse range of roof types but provide ground truth meshes with scan noise and dense triangles, reconstructed from large-coverage real-world scanning. City3D \cite{huang2022city3d} and Building3D \cite{wang2023building3d} provide datasets featuring coarser ground truths, utilizing plane partition-based reconstruction algorithms and artist-assisted refinement methods, respectively.
These datasets can contribute to research in compact mesh reconstruction.
However, the meshes, are reconstructed from real-world scans and may contain scan noise and algorithmic errors, which do not align with our requirement for clean, error-free data.

BuildingNet \cite{selvaraju2021buildingnet} provides a wide variety of compact meshes but of only 2\KK\ buildings. 
Furthermore, direct usage is often impractical due to the inclusion of non-relevant elements like humans, virtual ground, trees, cars, and landscaping. 
Although BuildingNet \cite{selvaraju2021buildingnet} provides the classifying labels, we have encountered several misclassifications that can adversely affect the accuracy of roof-only ground truth extraction.

\newpage
\section{Experiments}
\label{sec:experiments}

\subsection{Implementation Details}
\label{subsec:implementation-details}
\textbf{Network Architecture.} We followed Palette \cite{saharia2022palette} by employing a U-Net \cite{ronneberger2015u} architecture with an attention mechanism \cite{vaswani2017attention} in its deeper layers to construct our conditional diffusion models. 
Specifically, the input size is $2\times128\times128$, one channel for the corrupted height map, $\condImage$, and another for the estimated height map at step $t$, $\hat{\image}_t$. 
The output is the predicted noise in $\hat{\image}_t$ and is of size $1\times128\times128$. 
Our network contains four down-sampling modules and four up-sampling modules. 
These modules operate at resolutions of $128 \times 128$, $64 \times 64$, $32 \times 32$, and $16 \times 16$, with channel dimensions set to 64, 128, 256, and 512, respectively. 
Each module includes two residual blocks \cite{he2016deep,dhariwal2021diffusion} and the attention mechanism \cite{vaswani2017attention} is integrated into the modules when the resolution reaches $32 \times 32$ and $16 \times 16$.

\textbf{Data Augmentation.} We augment each height map by rotating 90, 180, and 270 degrees. Outlier noise occurs with a probability of 0.01\%. 
Global noise is synthesized by sampling $\sigma_{\text{global}}$ from a uniform distribution $\mathcal{U}(0, 0.05)$. 
For each pixel, a Gaussian distribution is constructed with the height as the mean and $\sigma_{\text{global}}$ as the variance; subsequently, the height value is re-sampled using this distribution. 
We also augment the data with varying sparsity including 99, 98, 90, 80, 50, and 25\%.

For synthesizing incompleteness, we follow \cref{subsec:incompletedetail} and use \cref{algo:incompletetraining} with five Gaussian distributions, $G=5$, to synthesize various types of incompleteness. For each Gauss distribution, the minimum variance, $\varMin$, and maximum variance, $\varMax$, are set to 0 and 0.3, respectively.

The tree noise is injected according to  \cref{algo:tree} using a 30\% probability. We set the minimum and maximum tree count, $\treeCountMin$ and $\treeCountMax$, to 1 and 3, respectively. Furthermore, the tree coverage scaling parameters, $\xyScaleMin$ and $\xyScaleMax$, are set to 0.5 and 2.0. Lastly, $\zScaleMin$ and $\zScaleMax$ are set to 2.0 and 4.0 for height scaling.

\textbf{Training} is conducted on 8 NVIDIA Ampere A100 GPUs with a batch size of 512 for 260 epochs. 
This took approximately 2.5 days.
The settings for variance scheduling and the use of exponential moving averages are adopted from \cite{saharia2022palette}. 
The learning rate is set at $7\times10^{-5}$, and a warm-up learning rate is employed for the first 10\KK\ steps, starting with a factor of 0.2.

\textbf{Training \rdN.} During data augmentation, we retain the entire tree shape and do not remove tree noise outside the footprint.
While \rd\ uses a footprint image, $\footprint$,
that is 0 for inside pixels and 1 outside,
$\footprint$ is an array of 1s
when training  \rdN .
This
 guarantees that the model possesses no prior footprint information.

\newpage

\begin{table}[t]
    \centering
    \small
    \begin{tabular}{lcccc}
        \toprule
        \multirow{2}{*}{Methods} & \multicolumn{4}{c}{Tree Counts}  \\
        \cmidrule{2-5}
         & 0 & 1 & 3 & 5 \\
        \midrule
        w/o tree aug. & 0.203 & 0.363 & 0.523 & 0.768 \\
        w/ tree aug.  & 0.208 & \textbf{0.278} & \textbf{0.356} & \textbf{0.504} \\
        \bottomrule
    \end{tabular}
    \vspace{0.5\baselineskip}
    
    \caption{Evaluation of tree noise resilience in training with tree augmentation (RMSE in meters).}
    \label{tab:treeaug}
\end{table}

    

\begin{table}
    \centering
    \small
    \begin{tabular}{l@{\hskip 10pt}c@{\hskip 10pt}c@{\hskip 10pt}c@{\hskip 10pt}c@{\hskip 10pt}c@{\hskip 10pt}c}
        \toprule
        \multirow{2}{*}{Test Set} & \multicolumn{6}{c}{Steps}  \\
        \cmidrule{2-7}
        & 60 & 125 & 250 & 500 & 1000 & 2000 \\
        \midrule
        Easy (s95 i30) & 0.444 & \textbf{0.343} & 0.339 & 0.337 & 0.338 & 0.344 \\
        Hard (s99 i80) & 1.123 & 0.940 & 0.919 & \textbf{0.893} & 0.899 & 0.922 \\
        \bottomrule
    \end{tabular}
    \vspace{0.5\baselineskip}
    
    \caption{Impact of inference steps on height completion quality (RMSE in meters). \textbf{Bold} denotes the most efficient in terms of achieving acceptable quality with the fewest steps.}
    \label{tab:inferencestep}
\end{table}

\subsection{Ablation of Tree Noise Augmentation}
\label{subsec:tree-ablation}

We investigated the impact of tree augmentation during model training. 
We constructed multiple test sets featuring varying numbers of tree intrusions per building. 
Specifically, tree count specifies the number of trees that will 100\% appear in each building. 
\Cref{tab:treeaug} compares height map restoration quality for models trained with and without tree augmentation. 
The results clearly indicate that tree augmentation during training effectively enhances the resilience of the model to tree noise.

\subsection{Sampling Step Analysis}
\label{subsec:step}

We analyze the relationship between height completion quality and the number of inference steps. 
The model was trained using a fixed 2k steps, and we tested it by changing the number of inference steps, as detailed in \cref{tab:inferencestep}.
For easier datasets with 95\% sparsity and 30\% incompleteness, reducing the inference steps to 250 resulted in only a minor loss of quality. 
However, for harder datasets, 500 inference steps were needed for satisfactory results. 
We observed that for more difficult tasks more steps are needed.  
Also, we attempted to train the model directly with fewer steps such as 1\KK. This led to failure to converge.

\newpage

\begin{figure}[t]
    \setlength{\tabcolsep}{4pt}
    \centering
    \begin{tabular}{cccc}
        \largeSubImg{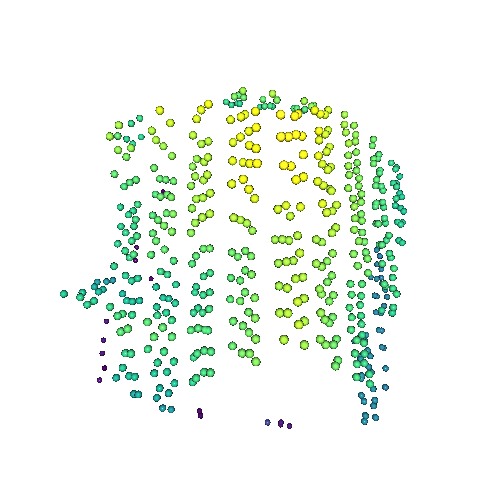}{Input 1}{} & \largeSubImg{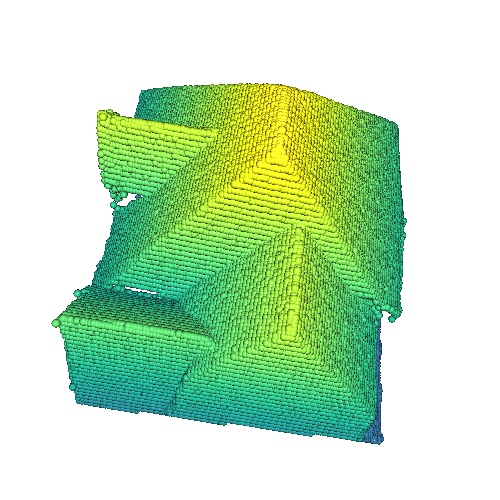}{Predicted 1}{} & \largeSubImg{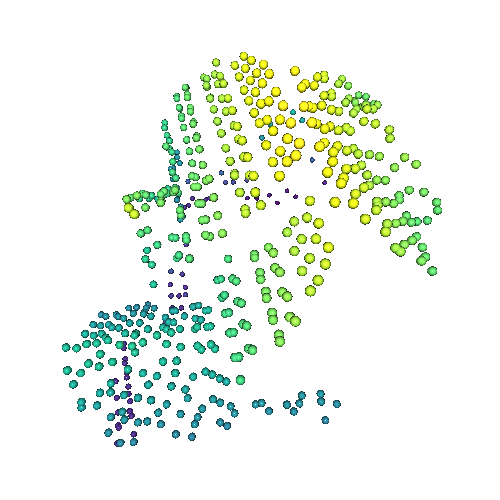}{Input 2}{} & \largeSubImg{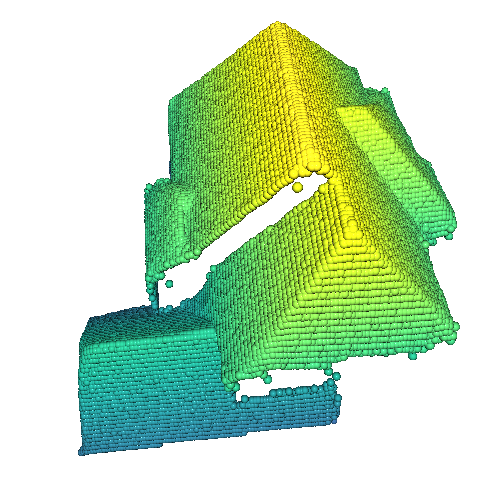}{Predicted 2}{} \\
    \end{tabular}

    \vspace{-0.7\baselineskip}
    
    \caption{\rd\ reconstruction of height maps corrupted 
    by scan line strip patterns.}
    \label{fig:linestrip}
\end{figure}

\begin{figure}[t]
    \newcolumntype{G}{@{\hskip 0.007\columnwidth}}
    \newcolumntype{I}{>{\centering\arraybackslash}m{\TinyImageSize}}
    \setlength\tabcolsep{0pt}
    \centering
    \resizebox{\textwidth}{!}{
    \begin{tabular}{IGIGIGIGIGIGIGIGIGI}
        \tinyImg{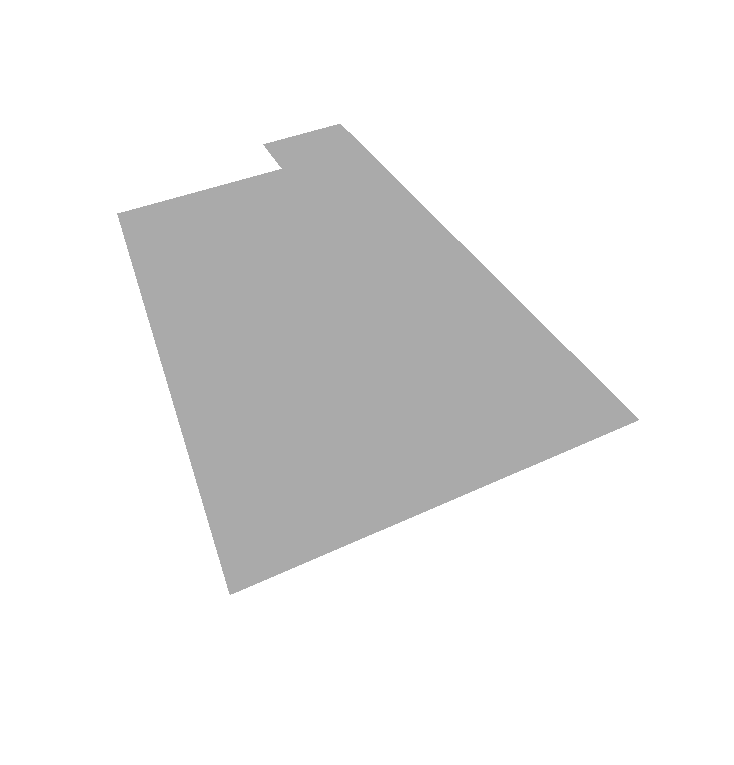} &
        \tinyImg{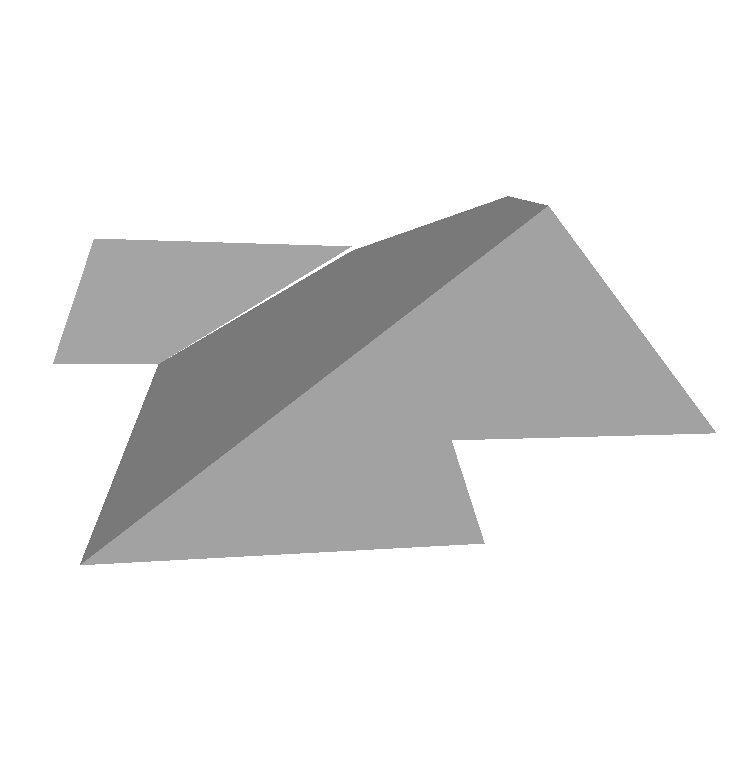} &
        \tinyImg{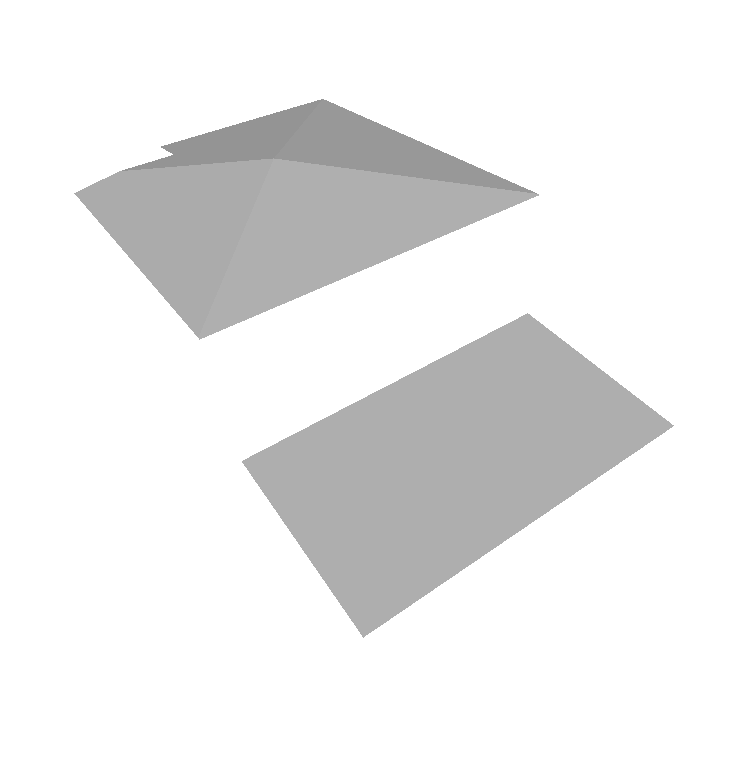} &
        \tinyImg{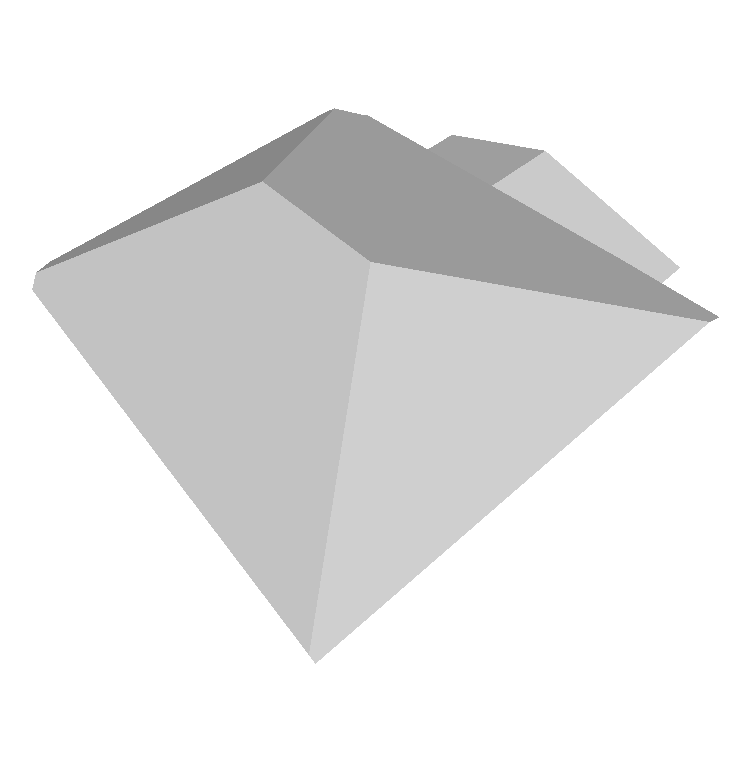} & 
        \tinyImg{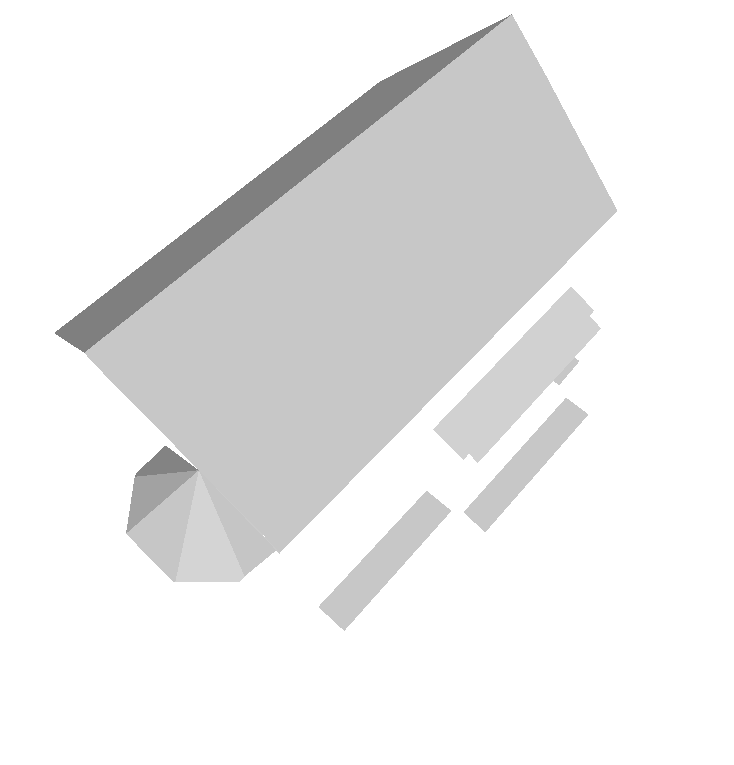} & 
        \tinyImg{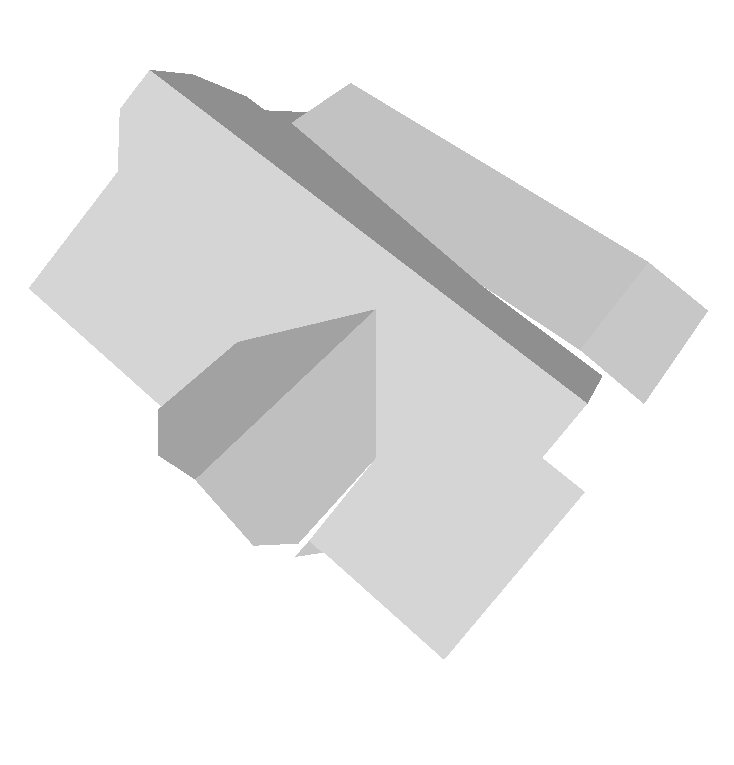} & 
        \tinyImg{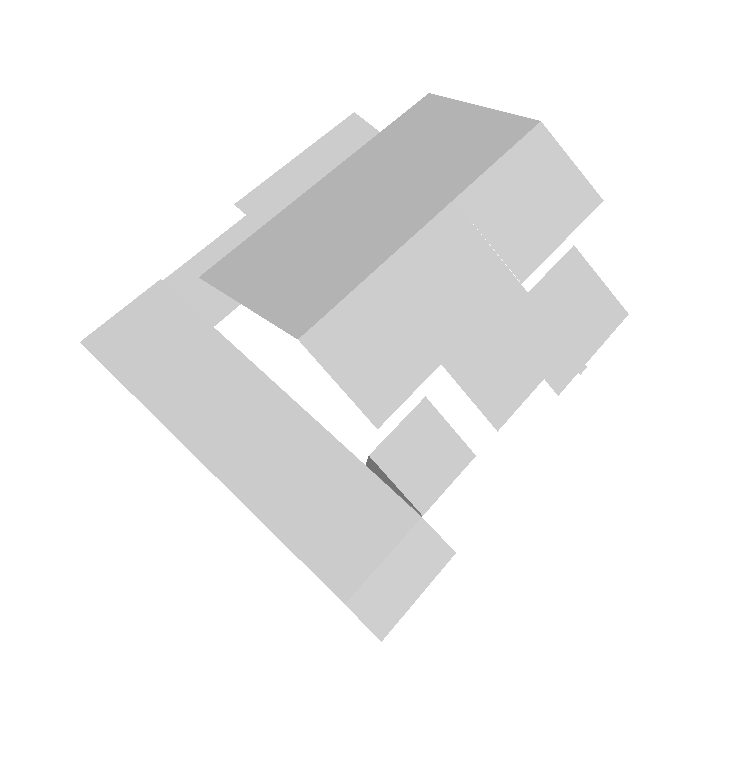} & 
        \tinyImg{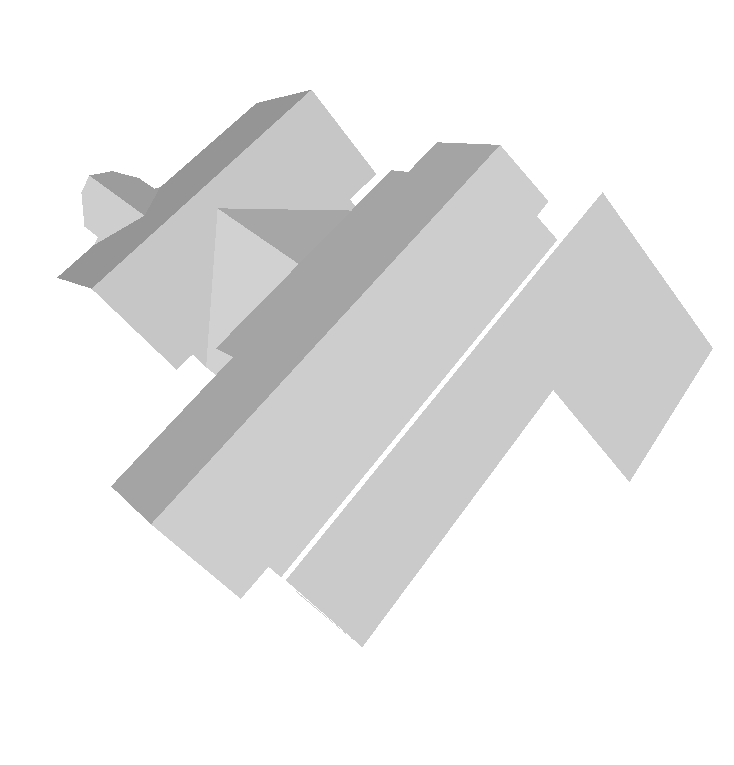} & 
        \tinyImg{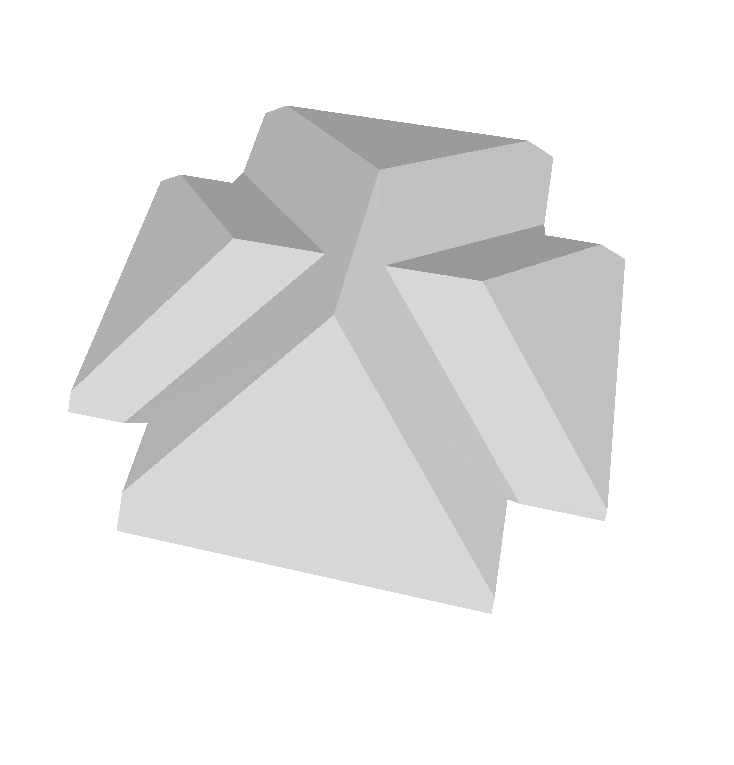} & 
        \tinyImg{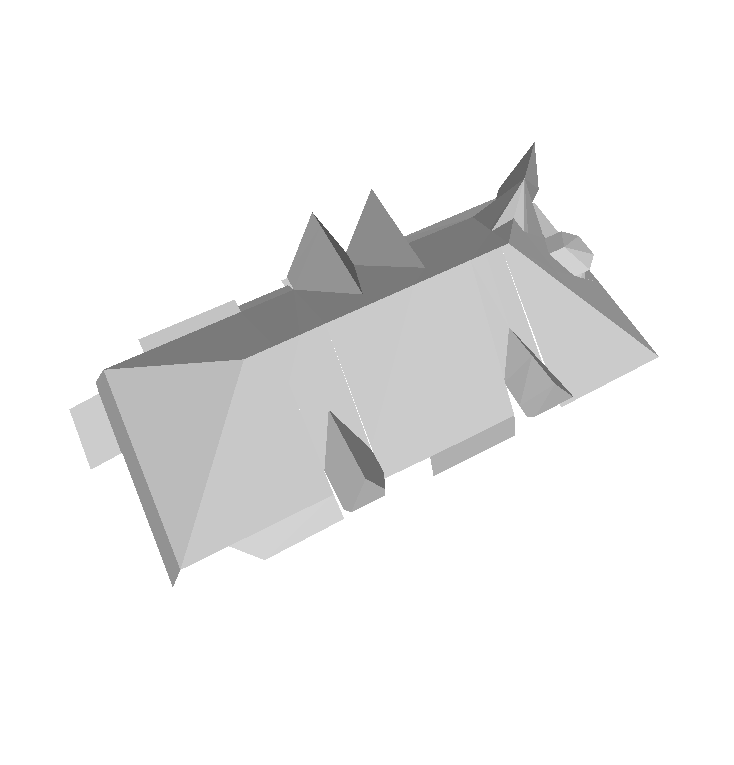} \\
        \tinyImg{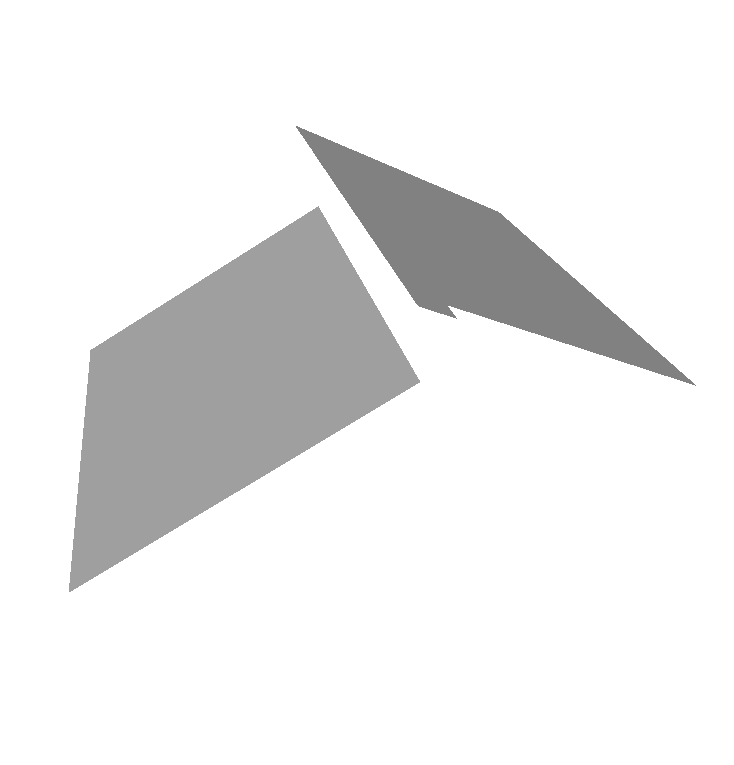} &
        \tinyImg{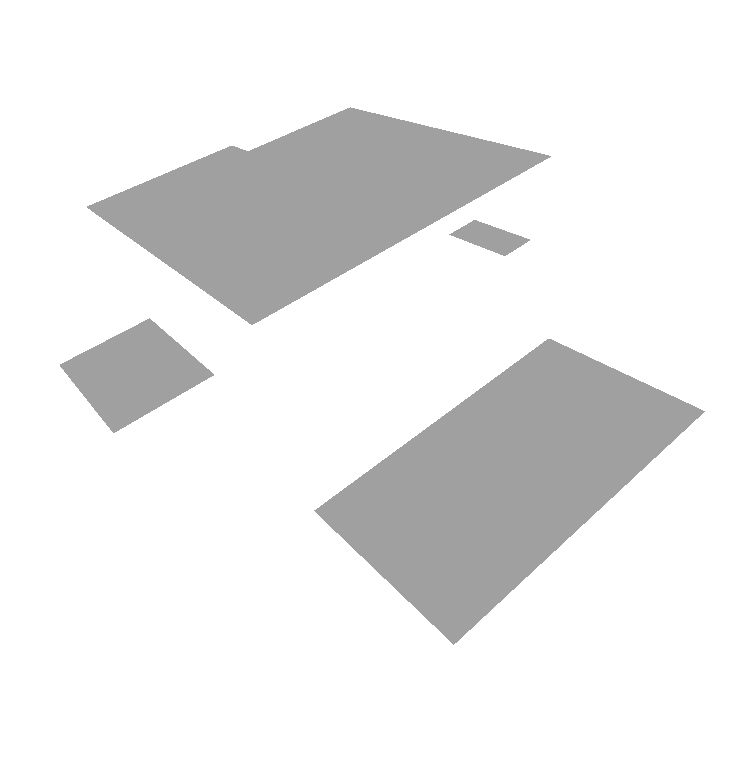} &
        \tinyImg{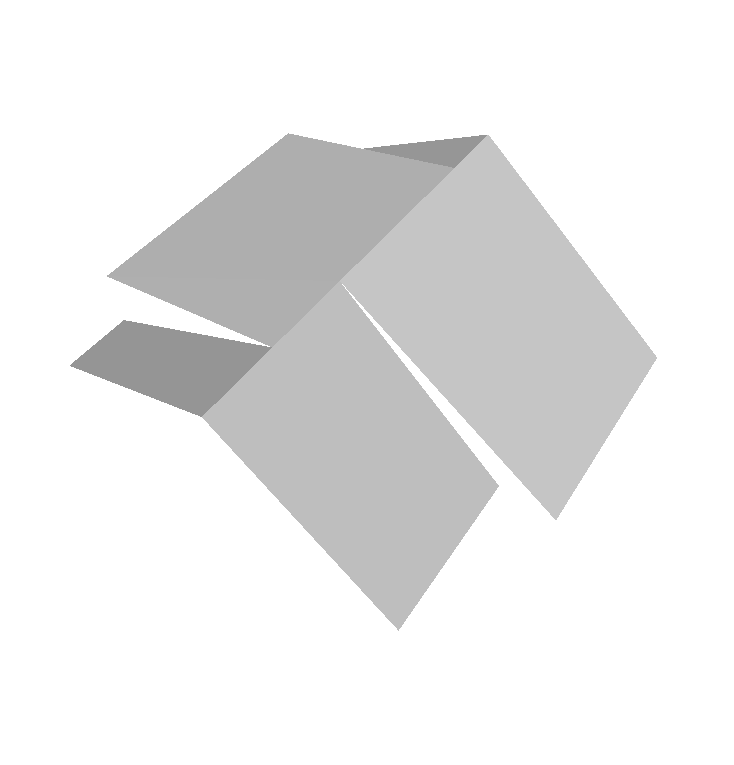} &
        \tinyImg{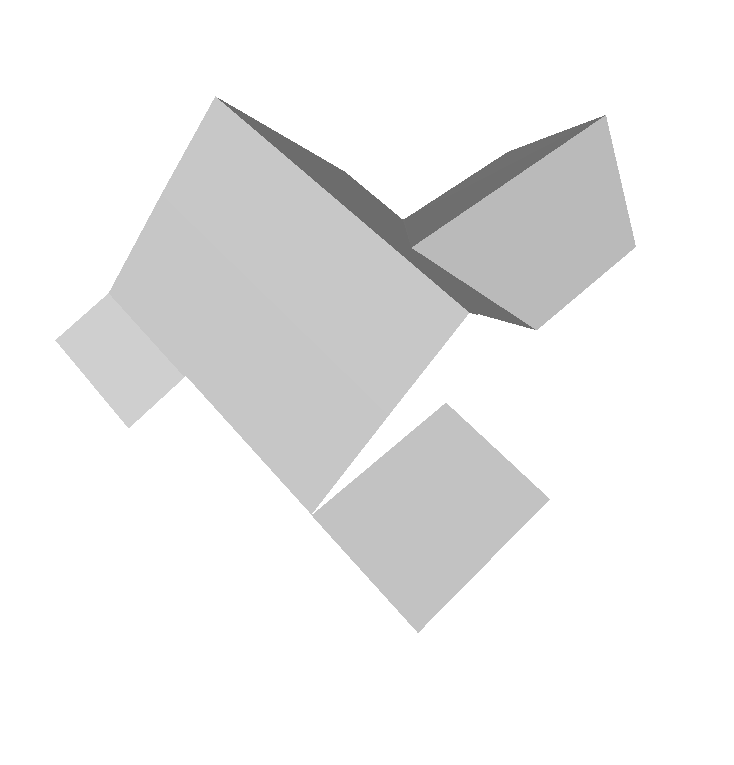} & 
        \tinyImg{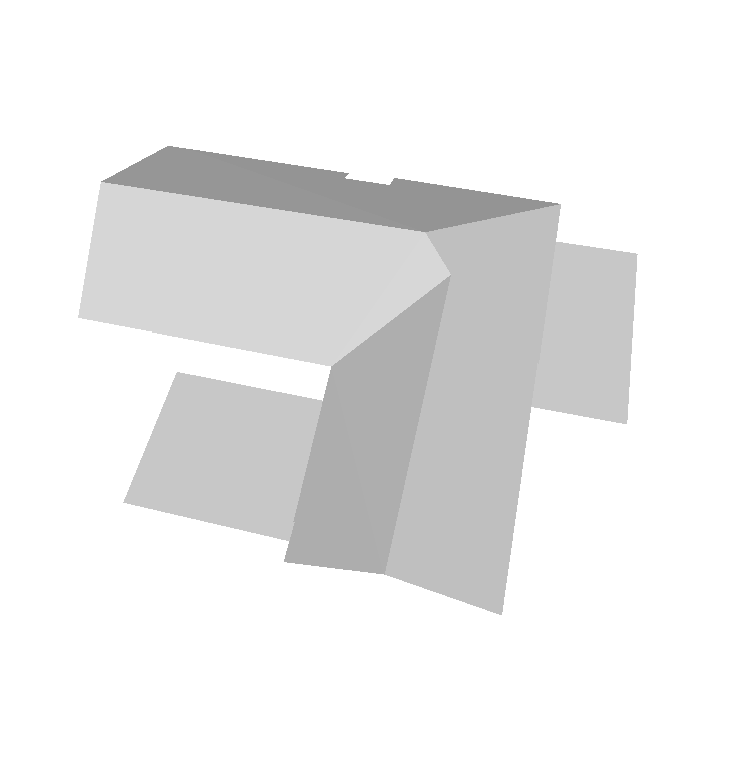} & 
        \tinyImg{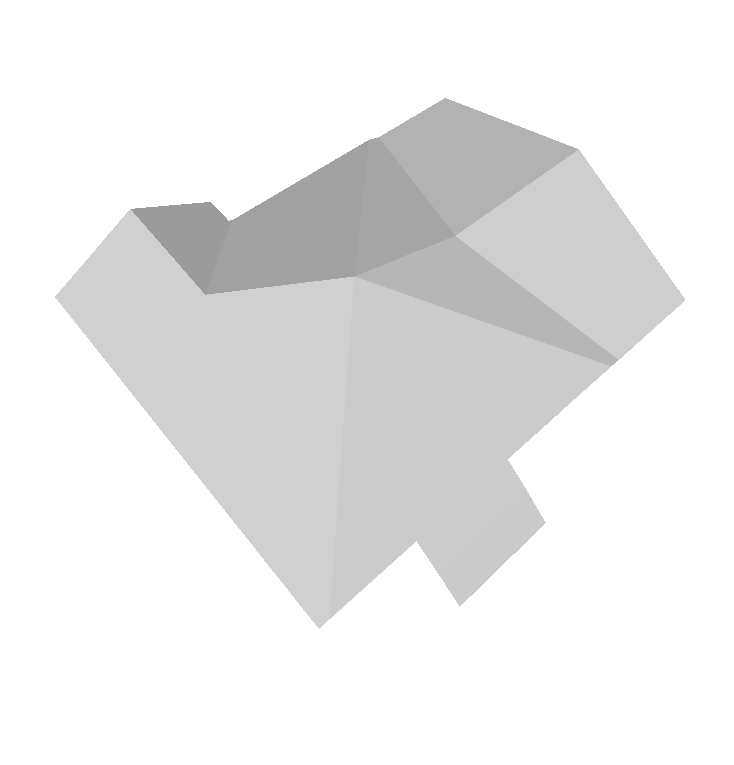} & 
        \tinyImg{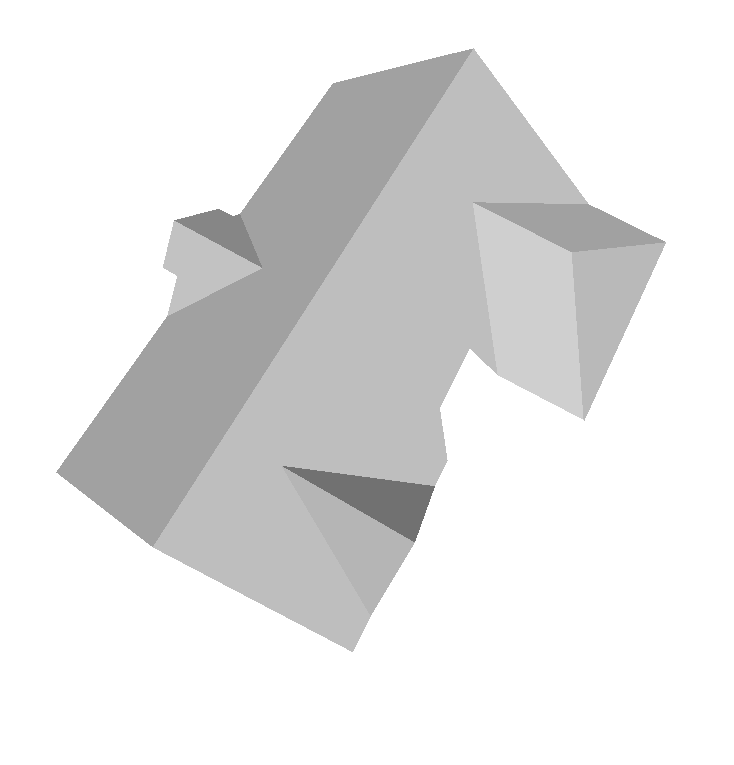} & 
        \tinyImg{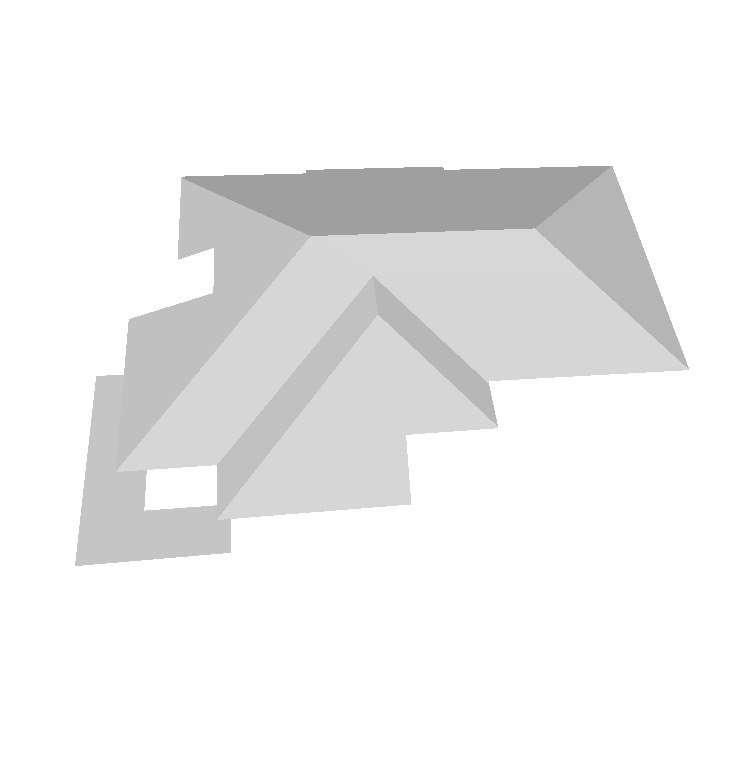} & 
        \tinyImg{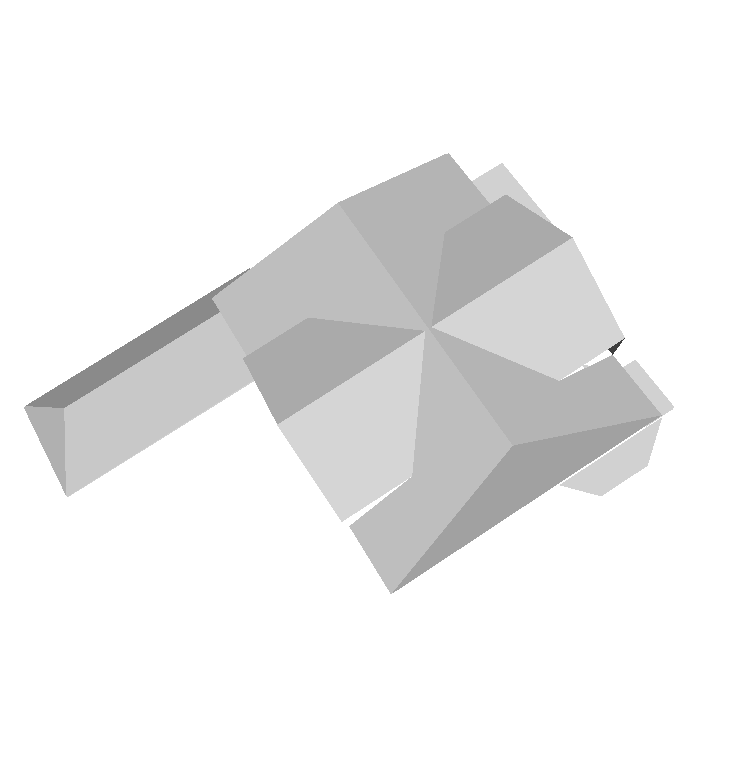} & 
        \tinyImg{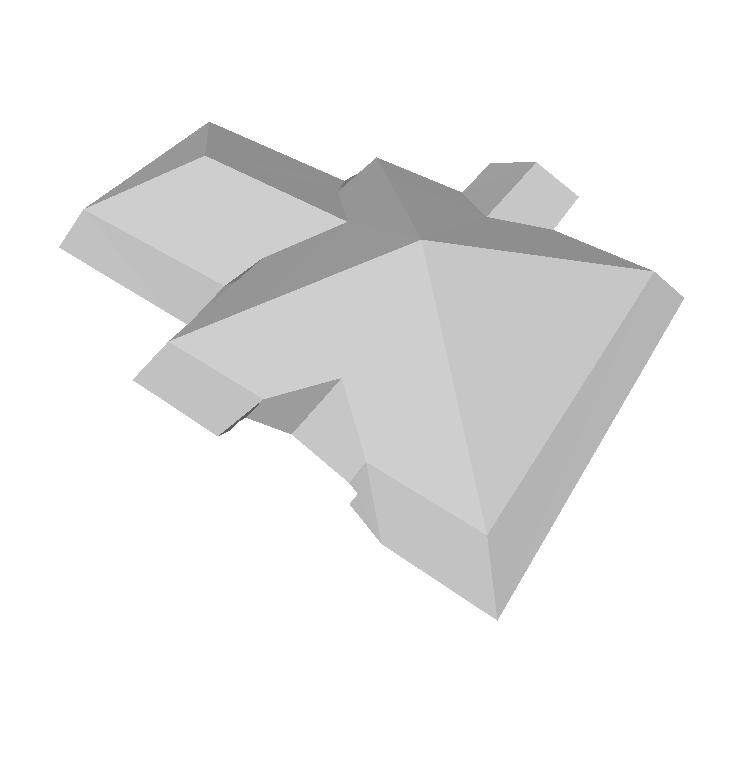} \\
        \tinyImg{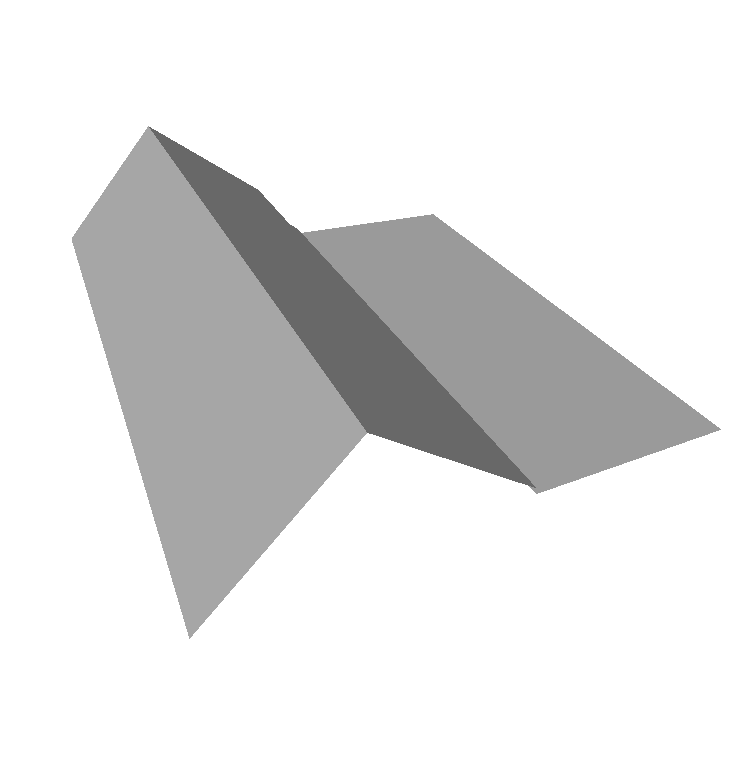} &
        \tinyImg{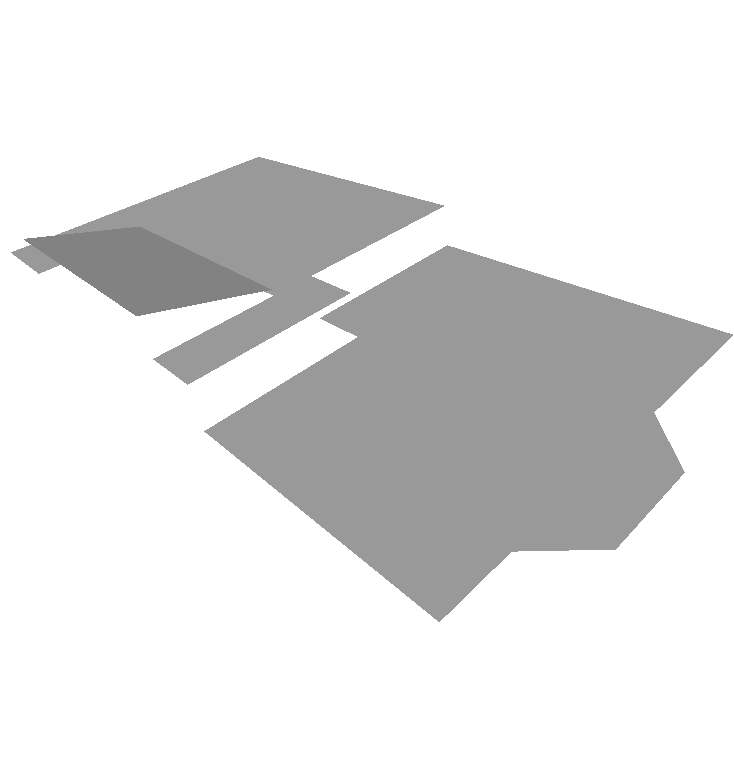} &
        \tinyImg{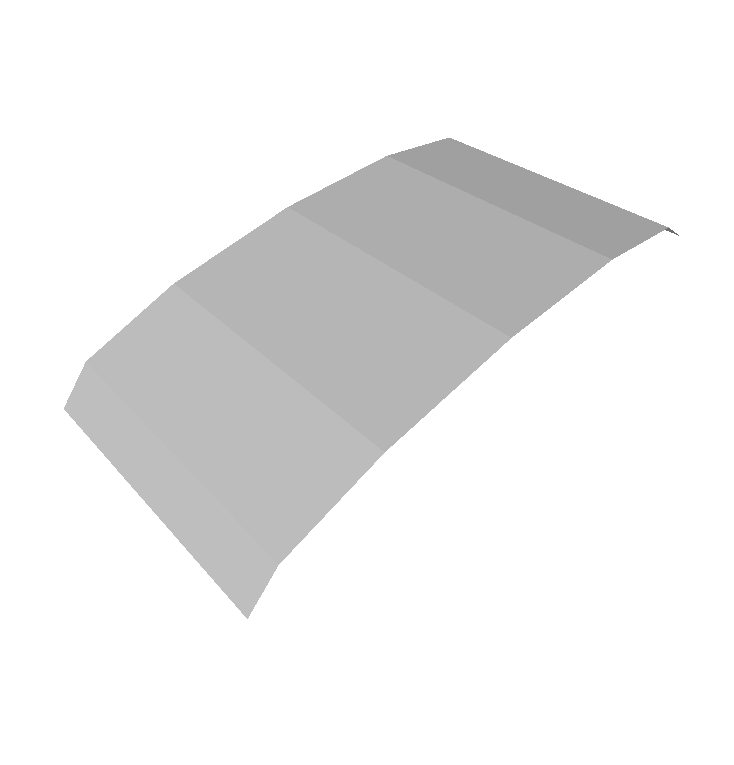} &
        \tinyImg{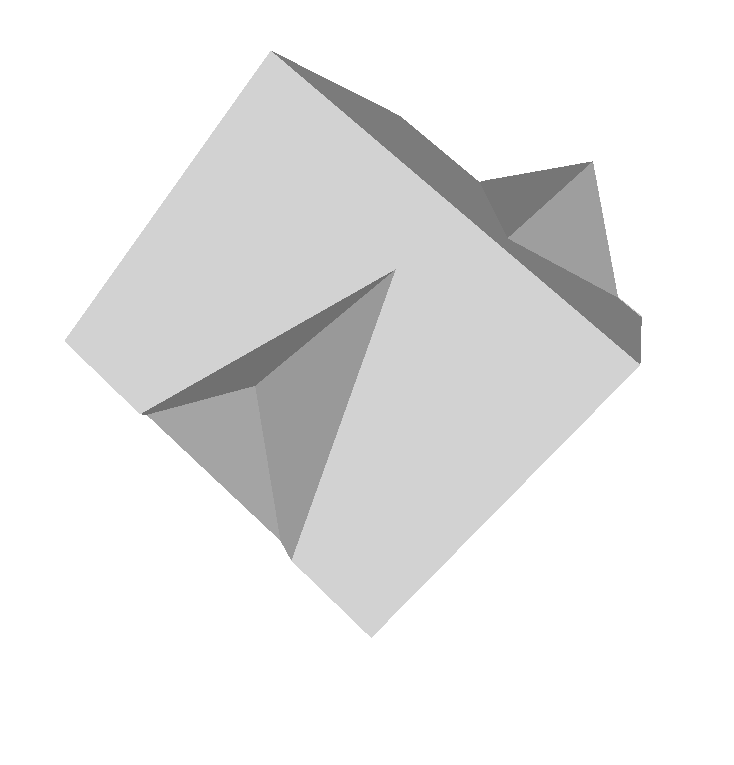} & 
        \tinyImg{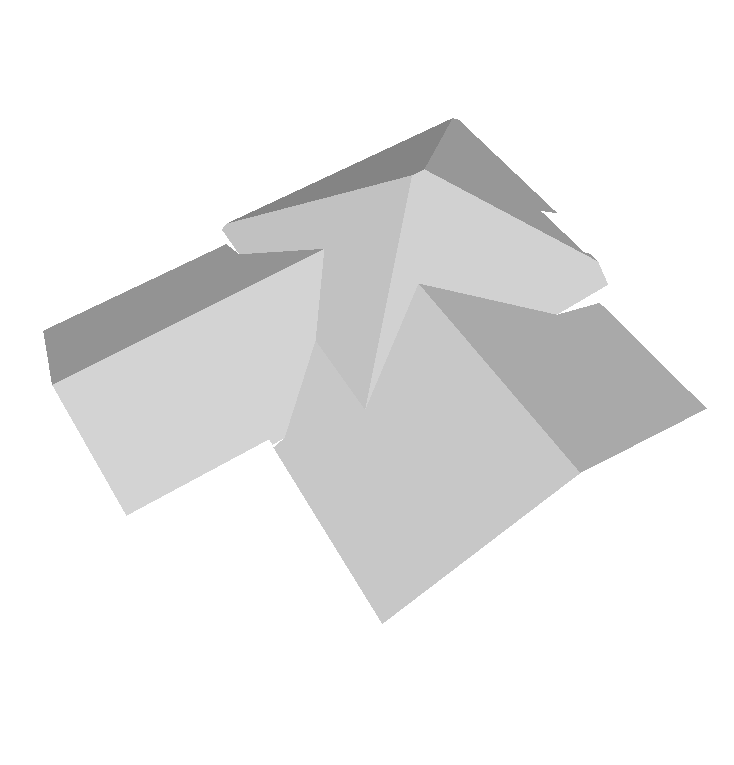} & 
        \tinyImg{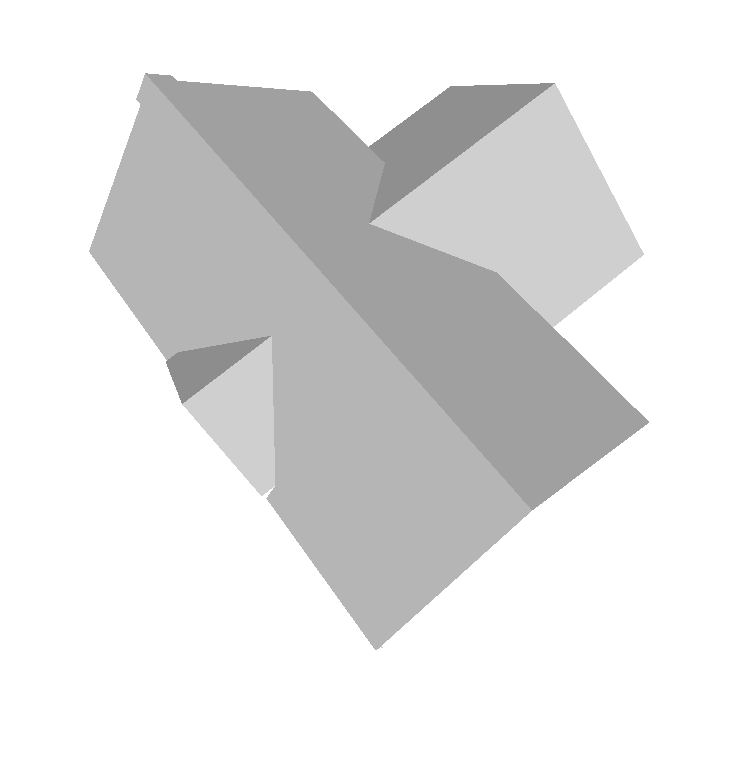} & 
        \tinyImg{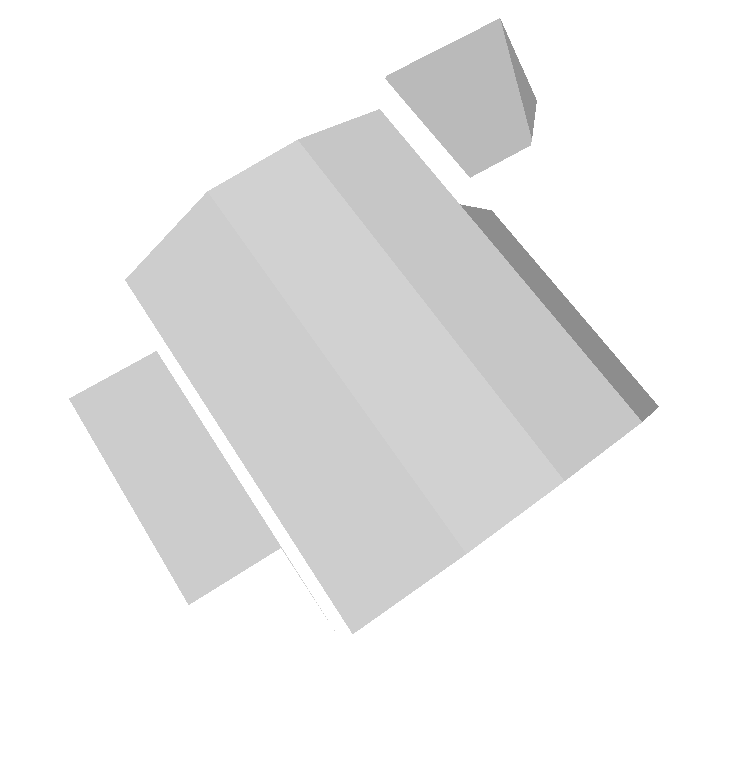} & 
        \tinyImg{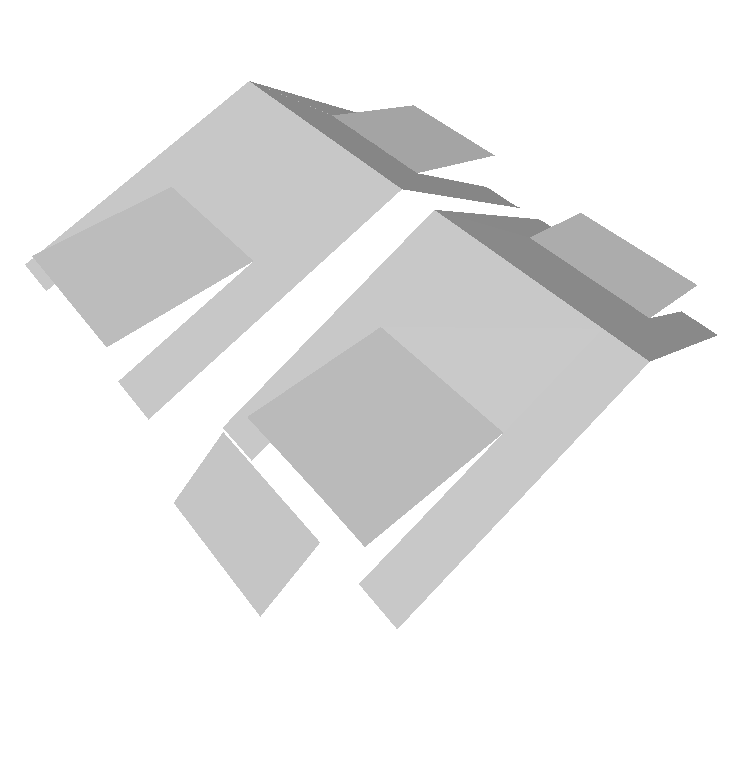} & 
        \tinyImg{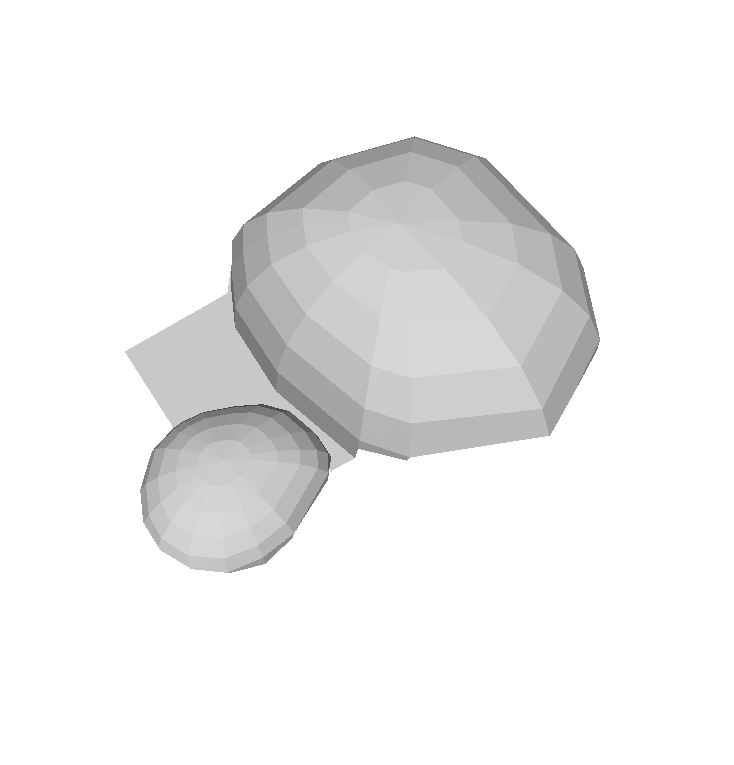} & 
        \tinyImg{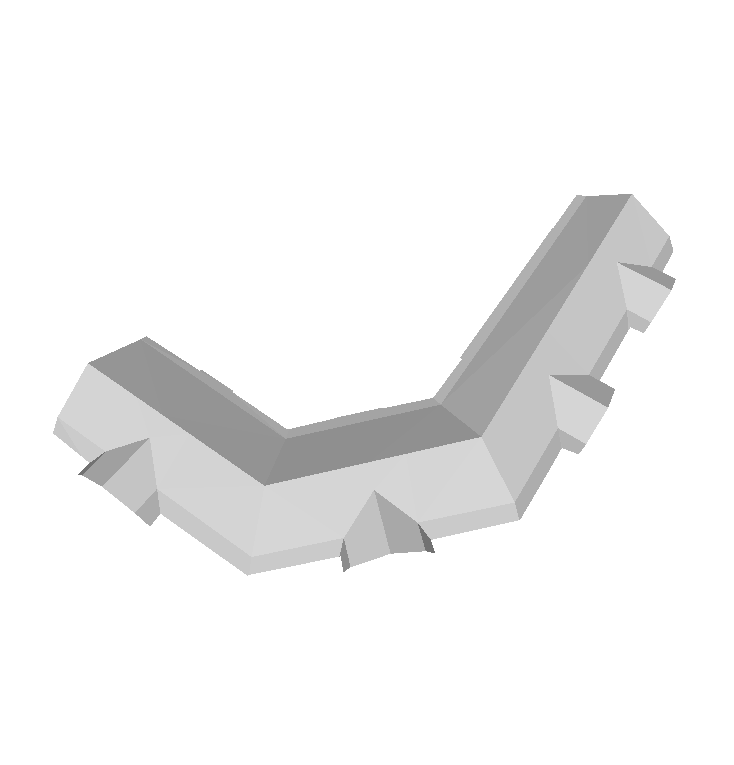} \\
    \end{tabular}
    }

    \vspace{-0.7\baselineskip}
    
    \caption{Samples in the \pdata\ (Poznan Roof Dataset).}
    \label{fig:poznandataset}
    \vspace{-1.0\baselineskip}
\end{figure}

\subsection{LiDAR Scan Patterns}
\label{subsec:lidar-pattern}
Our observations indicate that simulating sparse LiDAR points through random point removal yields results without noticeable gaps, closely resembling real-world scans. 
We found that most real-world point clouds \cite{varney2020dales, huang2022city3d, wayneco2017, cambridge2021} appear similar to those obtained via random sampling. 
While a few examples exhibit line strip patterns, these do not hinder the ability of \rd\ to restore roofs. \Cref{fig:linestrip} showcases the corrupted height maps featuring line strip patterns, and \rd\ effectively reconstructs complete, noise-free height maps from these.

\subsection{Roof Variety}
\label{subsec:roof-variety}
\Cref{fig:poznandataset} illustrates the variety of 
roof types in the 
\pdata\ dataset.  
Constructions of valid roofs of different types 
and with a variety of features have been shown for 
\rd{} and \rdN{}:
 gables (\cref{fig:morecambridge-e}), hips (\cref{fig:moreDales3d-a,fig:moreDales3d-h}), shads (\cref{fig:moreDales3d-e,fig:morecambridge-d}), leans (\cref{fig:moreAHN3-b,fig:moreAHN3-i}), and flats (\cref{fig:morecambridge-a}).

\subsection{How Prone is the Model to Hallucination based on Footprints?}
\label{subsec:heightmap-availability}
With only the footprint available, and no height data provided, we conducted tests using a height map set to all zeros, and varied the footprints. The output was predominantly zeros or values less than one meter. This suggests that the model uses footprints to enhance height prediction but does not by itself cause the diffusion model to hallucinate roof shapes.

\begin{figure}[t]
    \vspace{-0.5\baselineskip}
    \setlength{\tabcolsep}{0.1pt}
    \centering
    \begin{tabular}{ccccc}
        \ExtralargeImg{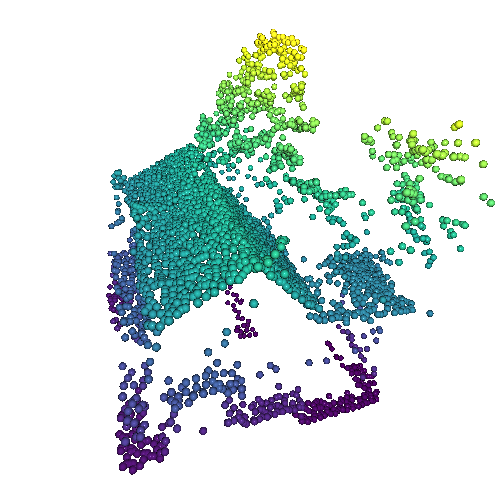} & \ExtralargeImg{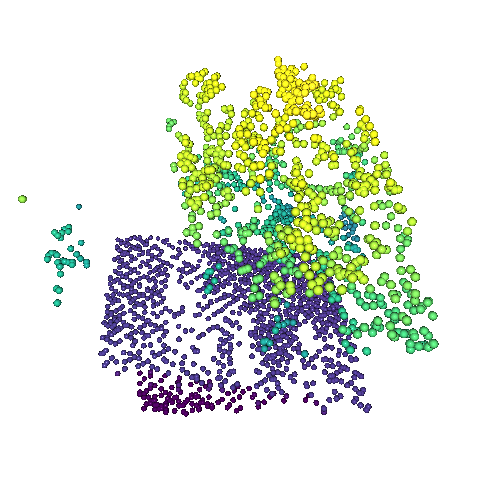} & \ExtralargeImg{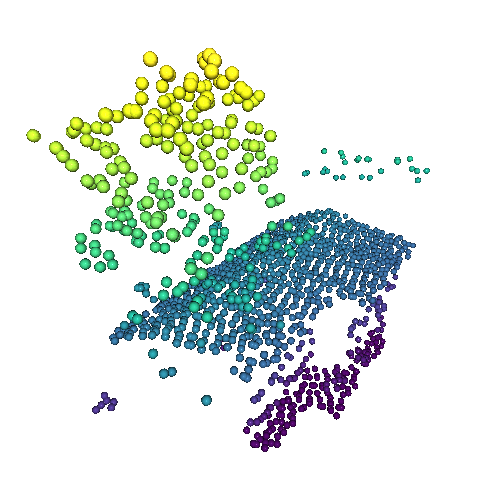} & \ExtralargeImg{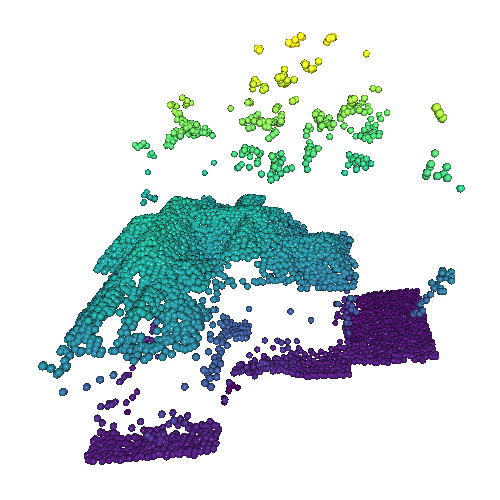} & \ExtralargeImg{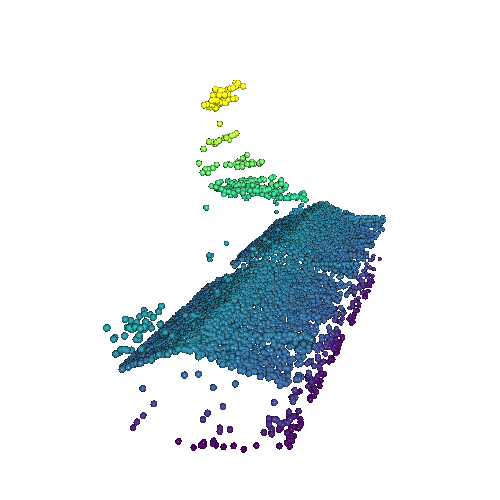} \\ [-3ex]
        \ExtralargeImg{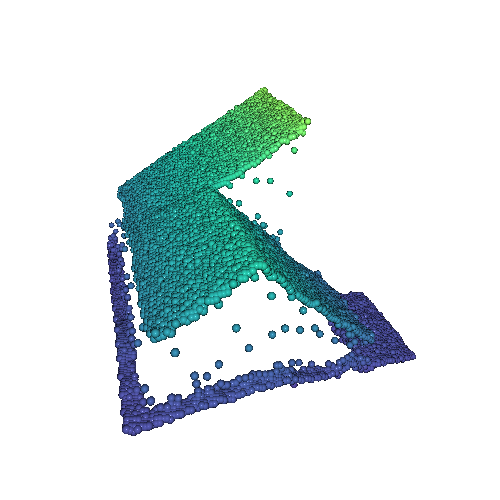} & \ExtralargeImg{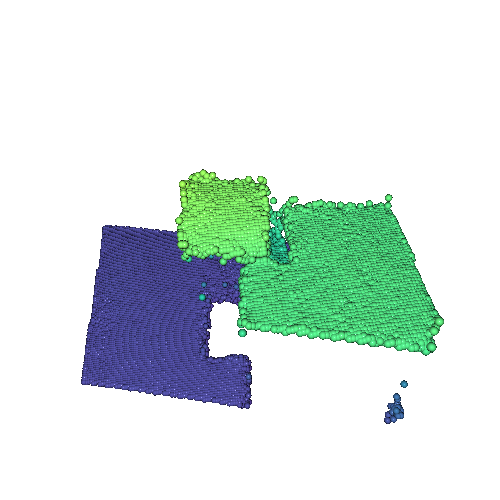} & \ExtralargeImg{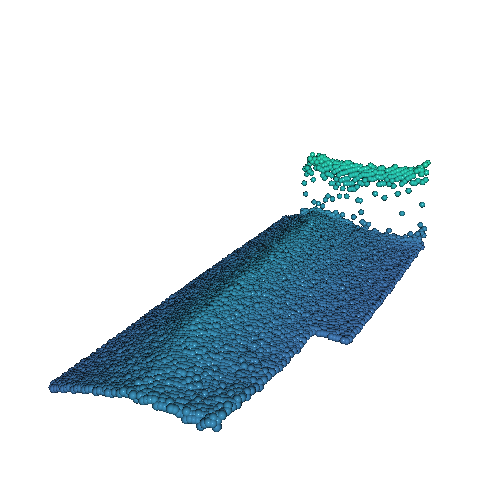} & \ExtralargeImg{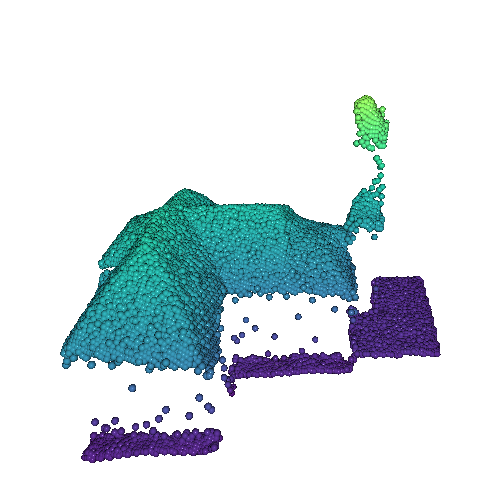} & \ExtralargeImg{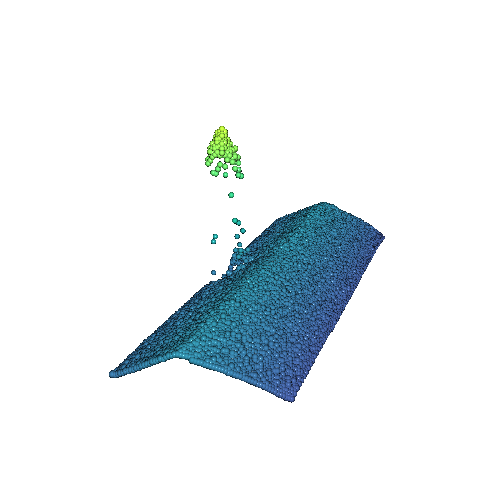} \\ [-4.5ex]
        \subcaptionbox{\label{fig:fail-a}}{} & \subcaptionbox{\label{fig:fail-b}}{} & \subcaptionbox{\label{fig:fail-c}}{} & \subcaptionbox{\label{fig:fail-d}}{} & \subcaptionbox{\label{fig:fail-e}}{} \\
    \end{tabular}

    
    \caption{Cases of tree noise 
    misinterpreted as roof structures.}
    \label{fig:fail}
\end{figure}

\subsection{Failure Cases}
\label{subsec:failure-cases}
\Cref{fig:fail} provides some examples of failure cases where \rd\ misclassifies tree noise as roof structures. 
\Cref{fig:fail-a,fig:fail-b} illustrate erroneous reconstructions of unlikely slope and flat planes, respectively, caused by extensive occlusion from tree points. 
\Cref{fig:fail-c} incorrectly identifies tree points as thin wall structures.
\Cref{fig:fail-d,fig:fail-e} mistakenly interpret tree noise as non-existent chimney structures.

\begin{table}[t]
    \centering
    \small
    \begin{tabular}{lcccc}
        \toprule
        Methods & iRMSE & iMAE & RMSE & MAE \\
        \midrule
        SparseConvs \cite{uhrig2017sparsity} & 4.94 & 1.78 & 1601.33 & 481.27 \\
        IP\_Basic \cite{ku2018defense}        & 3.78 & 1.29 & 1288.46 & 302.60 \\
        ADNN \cite{chodosh2019deep}       & 59.39 & 3.19 & 1325.37 & 439.48 \\
        Nconv\_CNN \cite{eldesokey2018propagating} & 4.67 & 1.52 & 1268.22 & 360.28 \\
        S2D (depth-only) \cite{ma2019self} & 3.21 & 1.35 & 954.36 & 288.64 \\
        HMS-Net \cite{huang2019hms}    & 2.93 & 1.14 & 937.27 & 258.48 \\
        pNCNN \cite{eldesokey2020uncertainty}   & 3.37 & 1.05 & 960.05 & 251.77 \\
        Physical\_Surface \cite{zhao2021surface} & 3.76 & 1.21 & 1239.84 & 298.30 \\
        DTP \cite{zhao2021distance}           & 2.94 & 1.07 & 937.27 & 247.81 \\
        CU-Net \cite{wang2022cu}          & \textbf{2.69} & \textbf{1.04} & \textbf{917.76} & \textbf{244.36} \\
        \midrule
        Ours (KITTI)     & 5.69 & 2.66 & 1641.41 & 491.61 \\
        Ours + M.O. (KITTI) & 4.84 & 1.84 & 1631.98 & 448.97 \\
        \bottomrule
    \end{tabular}
    \vspace{0.5\baselineskip}
    
    \caption{Evaluation of depth completion methods on KITTI dataset \cite{uhrig2017sparsity}. M.O. stands for applying mean shift offset.}
    \label{tab:depthcompletion}
\end{table}

\subsection{Depth Completion on the  KITTI Dataset}
\label{subsec:kitti}
We extend \rd\ to address the unguided depth completion task on the KITTI dataset \cite{uhrig2017sparsity}. 
Given the original KITTI depth map resolution of $352 \times 1216$, training with such size is impractical due to the excessive processing time required by the diffusion model. 
Therefore, we downsampled the depth maps to a resolution of $176 \times 608$ and trained our model using randomly cropped sections of size $176 \times 304$. 
We trained on the official training split with 86\KK\ samples. We conducted tests on the 1\KK\ validation depth maps \cite{uhrig2017sparsity}, where the predicted results with the size of $176 \times 608$ are upsampled back to $352 \times 1216$ for evaluation.

We compared to the results of state-of-the-art and representative unguided depth completion methods, reported in \cite{wang2022cu}. 
\Cref{tab:depthcompletion} combines these results with those newly reported by our method. 
We observed a mean shift phenomenon in our method.
To address this, we aligned the predicted depth map with the input sparse depth map by offsetting it with the difference between their respective means. 
This adjustment resulted in a significant performance boost.

Nevertheless, our results do not improve on
prior methods, likely due to the differences in KITTI depth completion v.s. roof completion.
Available pixels in the KITTI dataset \cite{uhrig2017sparsity}
are more uniformly distributed, and regional incompleteness size is smaller compared to roof completion tasks. 
Consequently, KITTI depth completion resembles an interpolation task more than inpainting for roof completion.

From this we conclude: 
diffusion models are more adept at tasks requiring the inpainting of large missing regions, while for smaller interpolation tasks, their performance is on par with other methods.
Our conclusion is supported by two observations. 
First, in our \pdata\ benchmark (see Table 2 in the main paper), the performance margin of \rd{} over existing methods \cite{keys1981cubic,biasutti2019diffusion,shepard1968two} is greater in tasks of restoring incompleteness compared to sparsity. Secondly, SpAgNet \cite{conti2023sparsity}, a diffusion model-based depth completion method, does not outperform existing methods \cite{eldesokey2020uncertainty,park2020non,guizilini2021sparse,hu2021penet} when completing sparsely scanned data with 64 lines. However, as the scan lines become sparser, 32, 16, 8, and 4 lines, SpAgNet \cite{conti2023sparsity} outperforms others.

\subsection{Additional Qualitative Experimental Results}
\label{subsec:additional-qualitative}
\begin{figure*}[t]
    \newcolumntype{G}{@{\hskip 3pt}}
    \newcolumntype{T}{m{0.015\textwidth}}
    \newcolumntype{I}{>{\centering\arraybackslash}m{\SmallImageSize}}

    
    \centering
    \begin{tabular}{G T G I G I G I G I G I G I G I G I}
        \scriptsize\rotatebox{90}{Input} & \smallImg{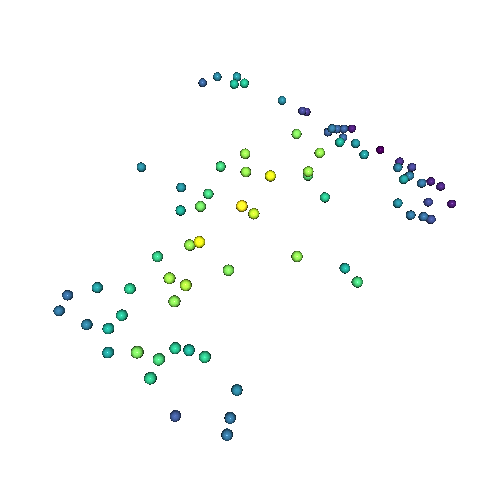} & \smallImg{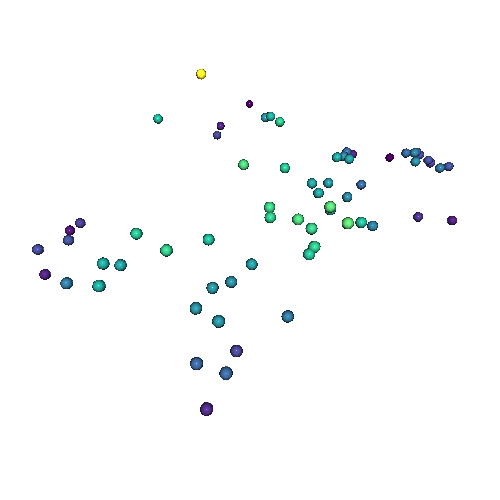} & \smallImg{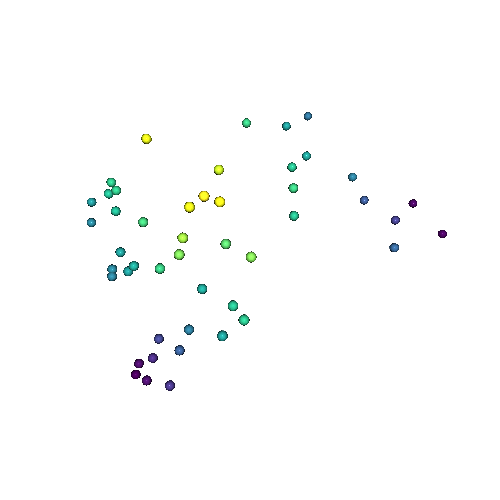} & \smallImg{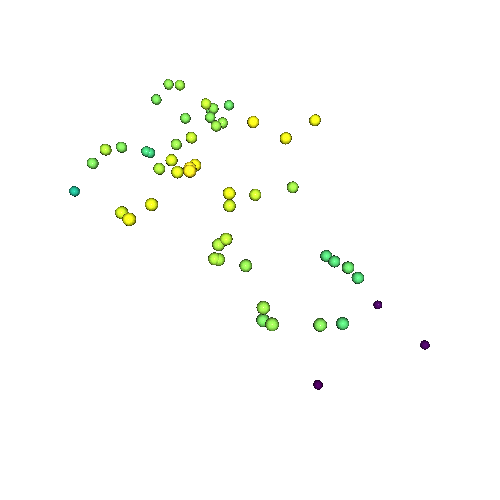} & \smallImg{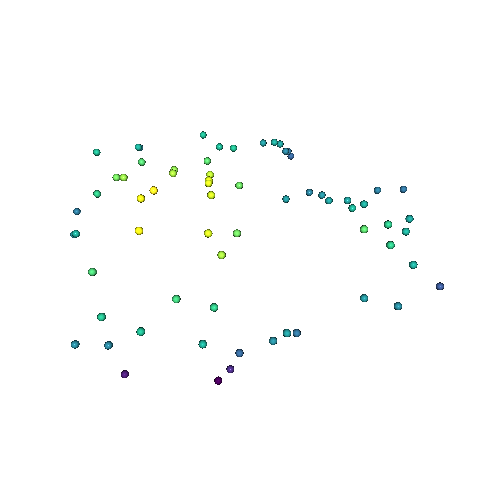} & \smallImg{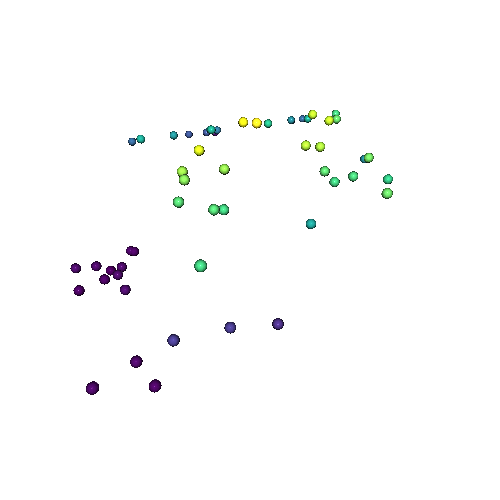} & \smallImg{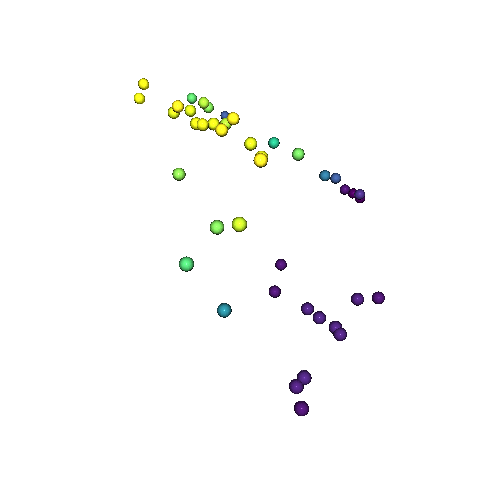} & \smallImg{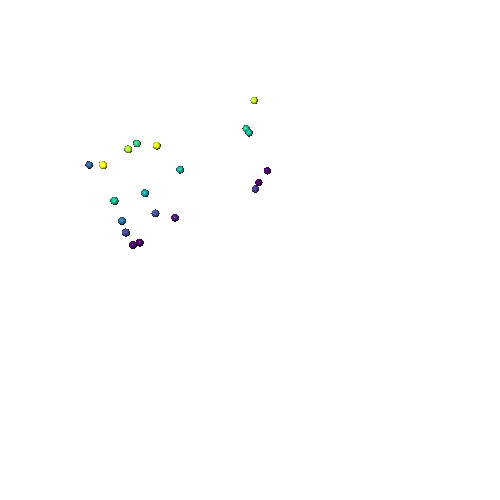} \\[-1.4ex]
    
        \scriptsize\rotatebox{90}{IDW} & \smallImg{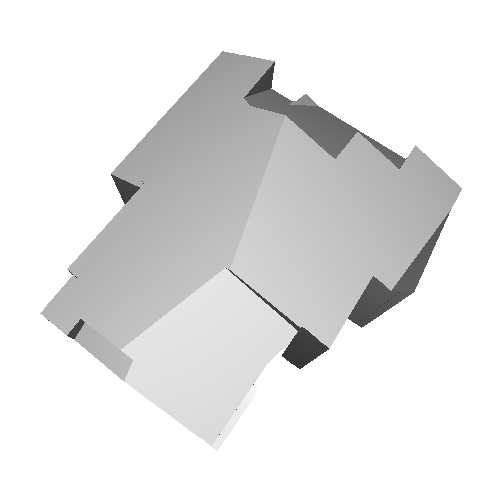} & \smallImg{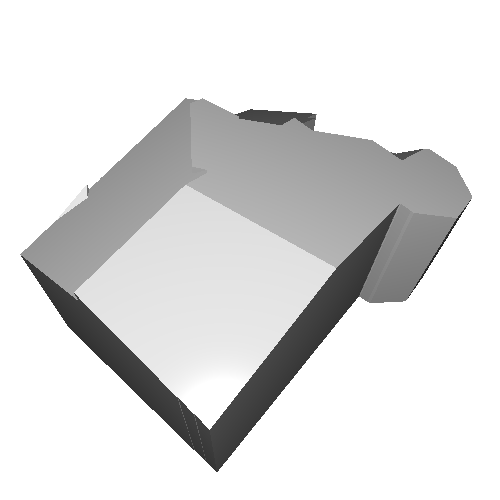} & \smallImg{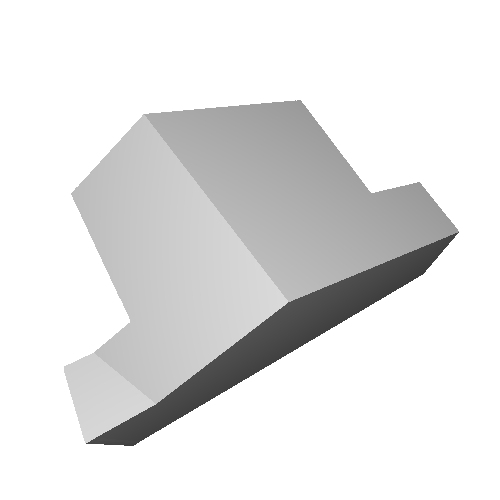} & \smallImg{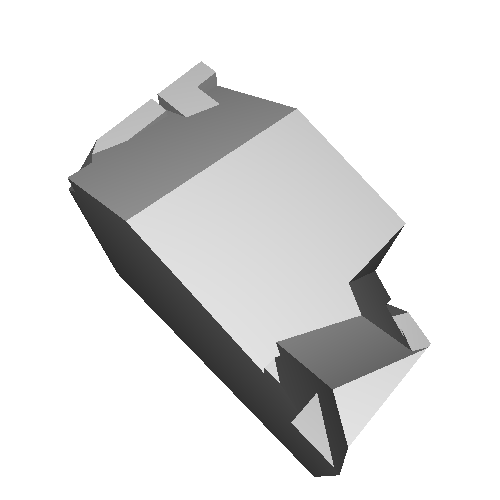}  & \smallImg{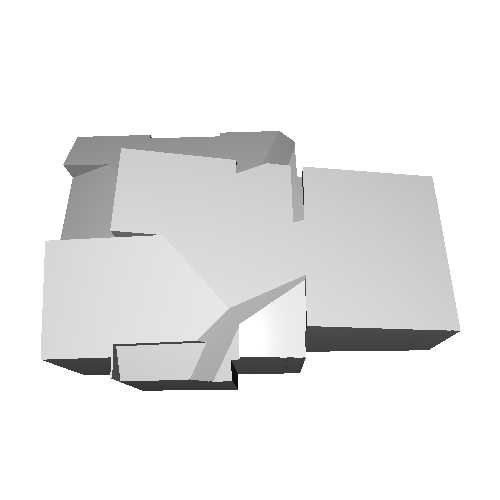}  & \smallImg{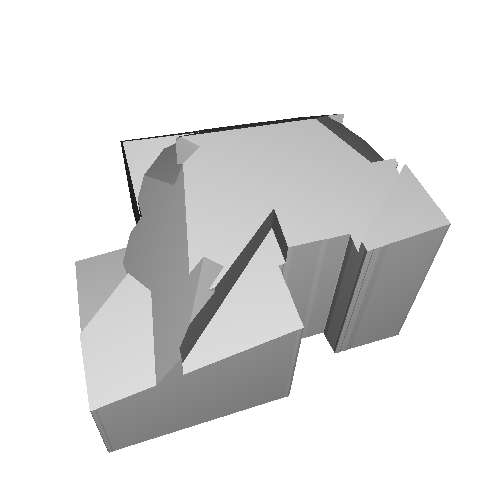}  & \smallImg{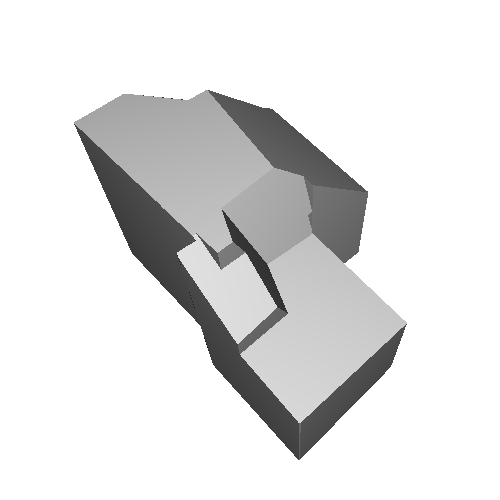}  & \smallImg{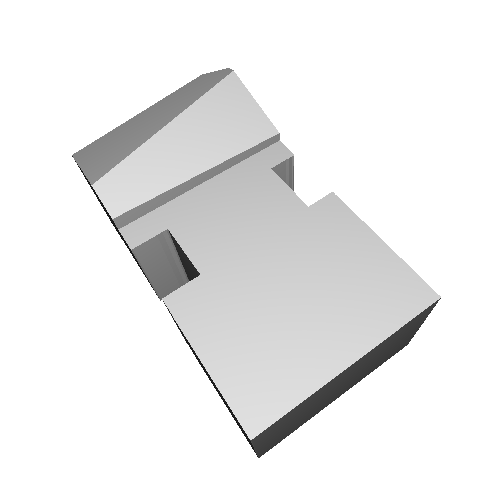} \\[-1.4ex]
        
        \scriptsize\rotatebox{90}{P.M. Diff.} & \smallImg{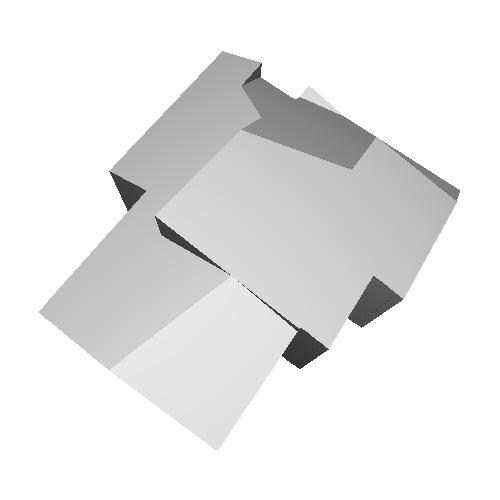} & \smallImg{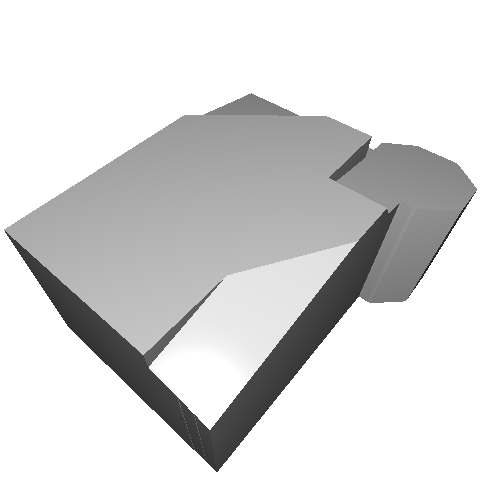} & \smallImg{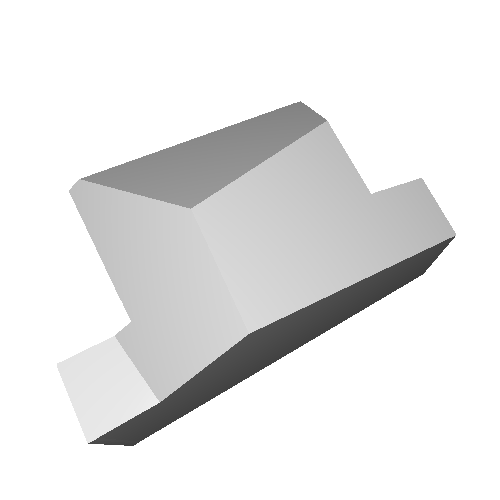} & \smallImg{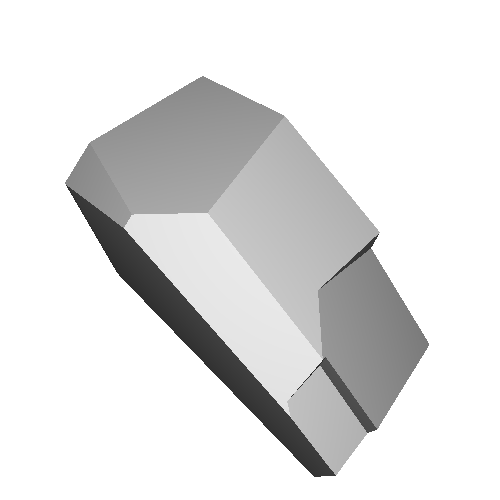} & \smallImg{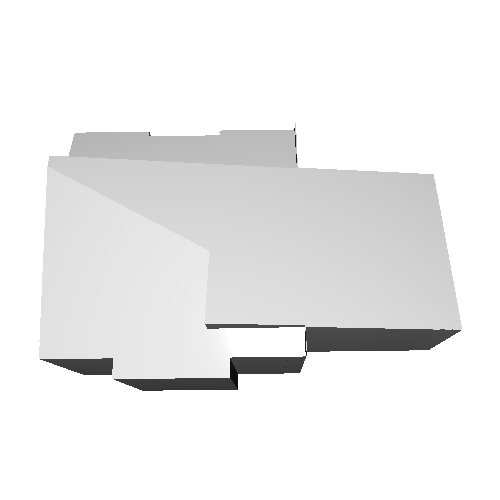} & \smallImg{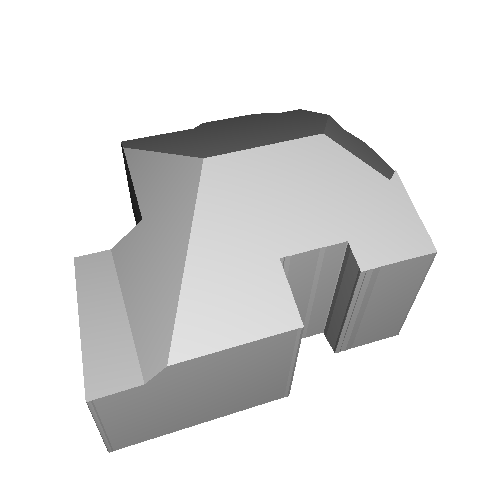} & \smallImg{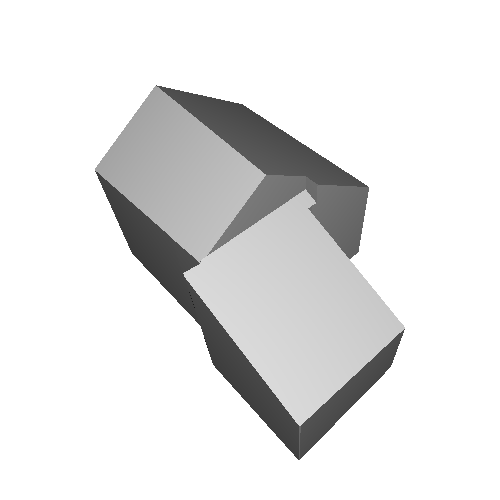} & \smallImg{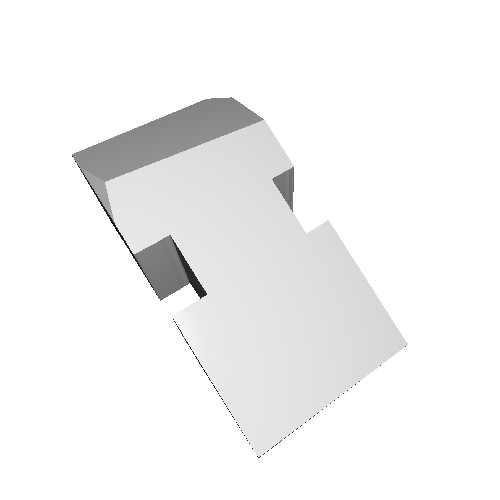} \\[-1.4ex]
        
        \scriptsize\rotatebox{90}{Ours} & \smallImg{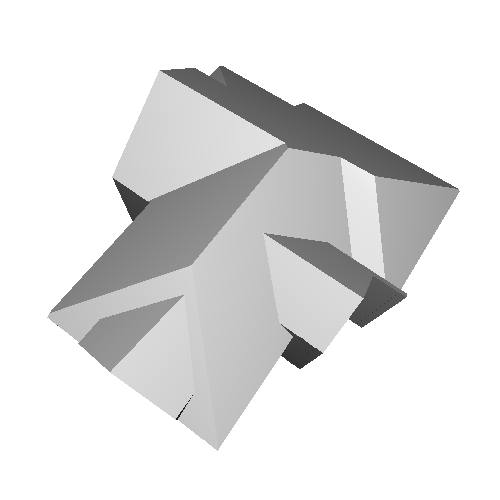} & \smallImg{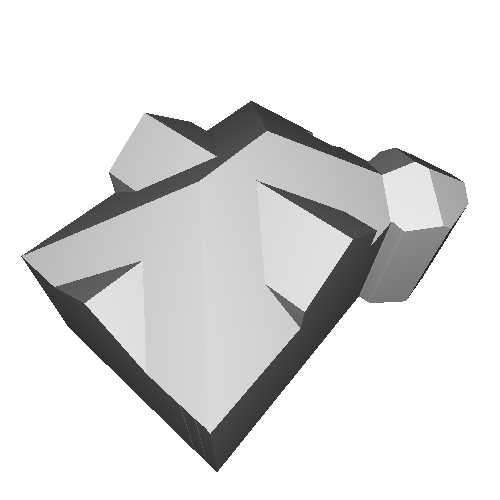} & \smallImg{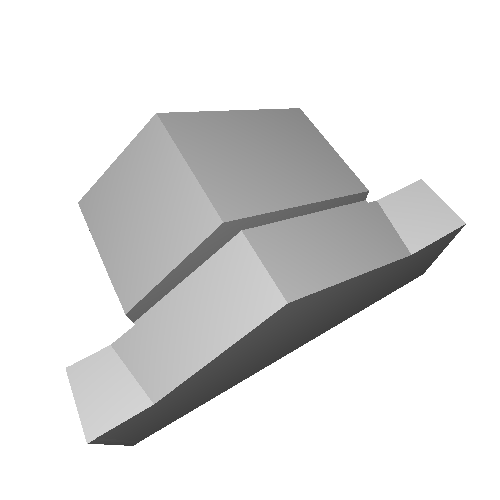} & \smallImg{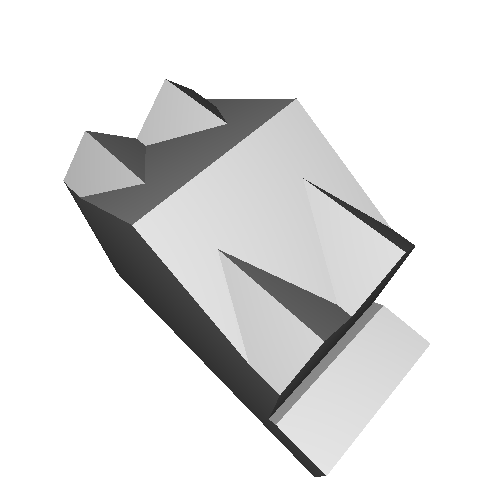} & \smallImg{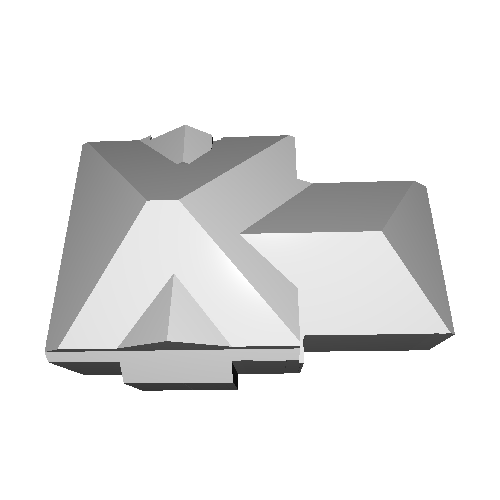} & \smallImg{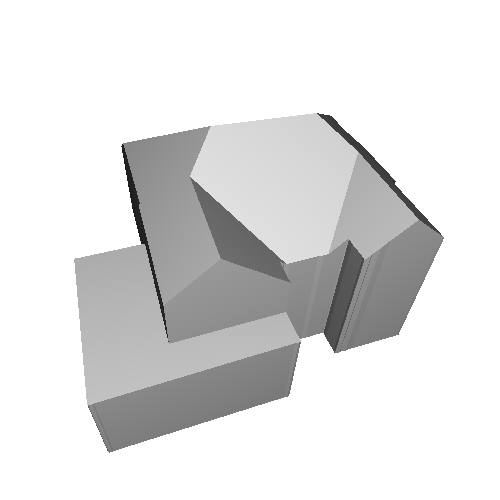} & \smallImg{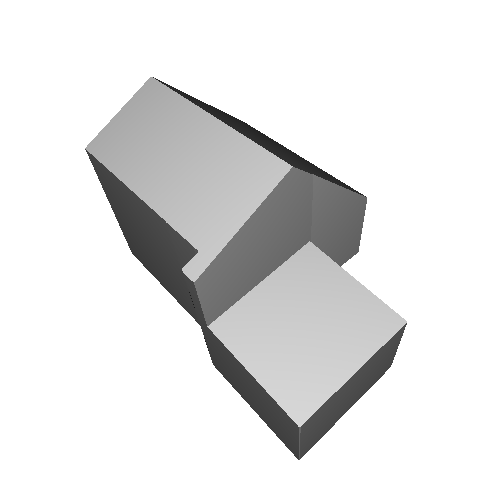} & \smallImg{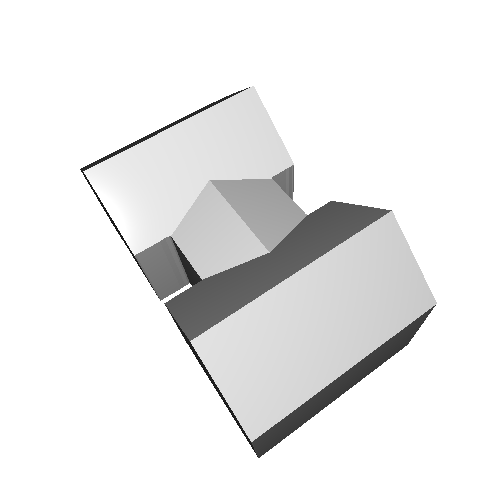} \\[-1.4ex]

        \scriptsize\rotatebox{90}{GT} & \smallSubImg{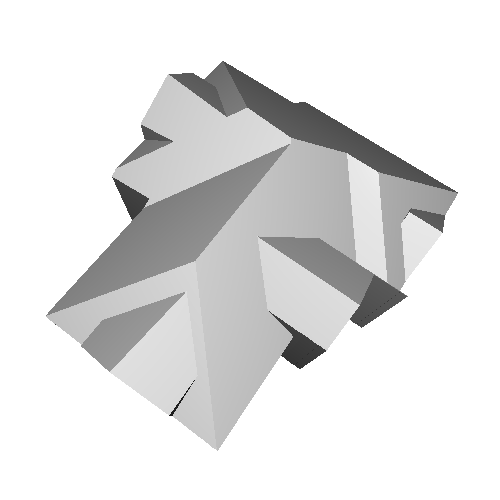}{}{fig:city3d-s-a} & \smallSubImg{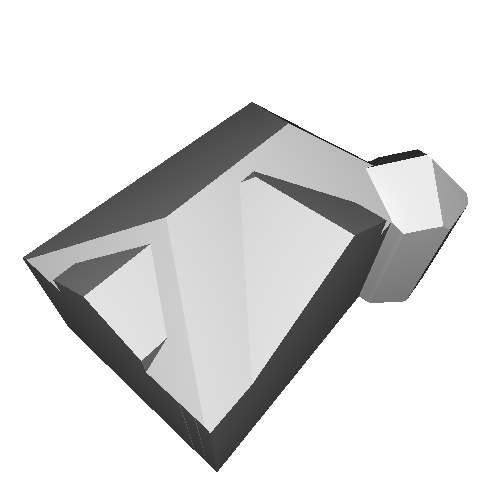}{}{fig:city3d-s-b} & \smallSubImg{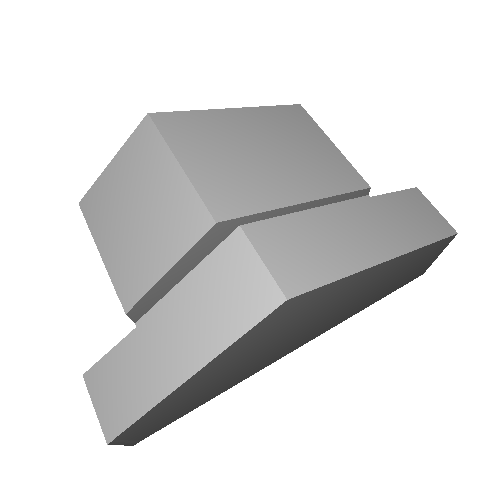}{}{fig:city3d-s-c} & \smallSubImg{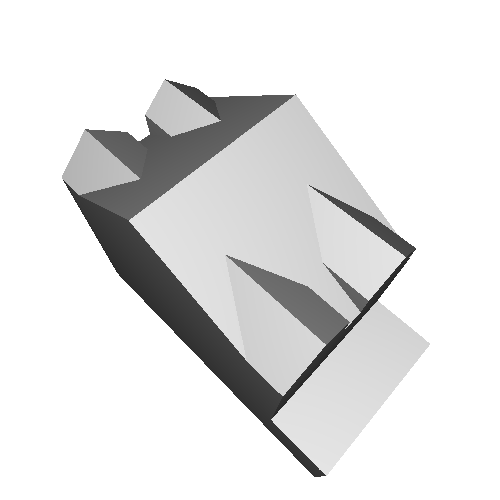}{}{fig:city3d-s-d} & \smallSubImg{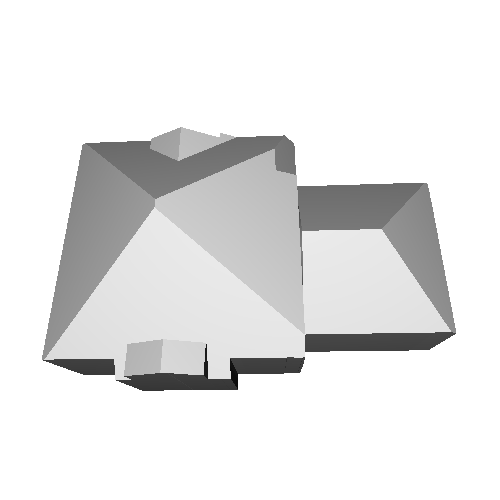}{}{fig:city3d-s-e} & \smallSubImg{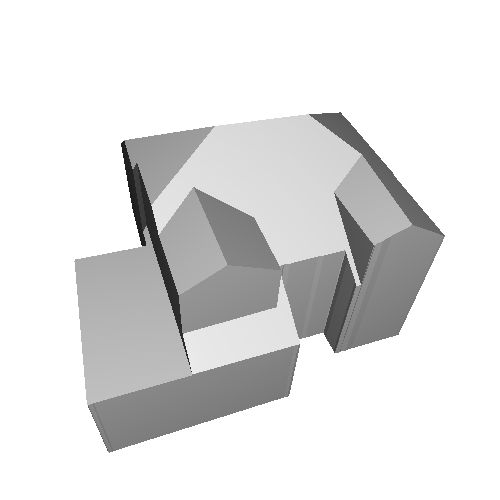}{}{fig:city3d-s-f} & \smallSubImg{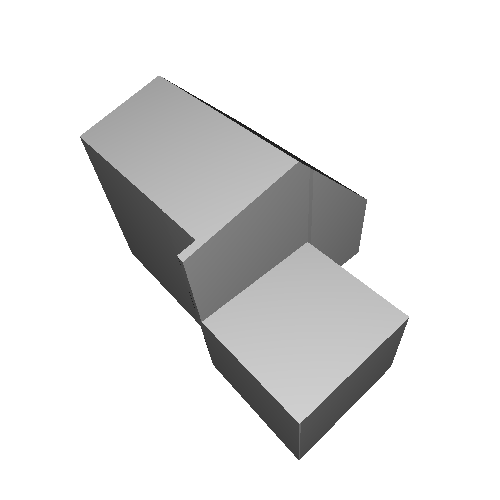}{}{fig:city3d-s-g} & \smallSubImg{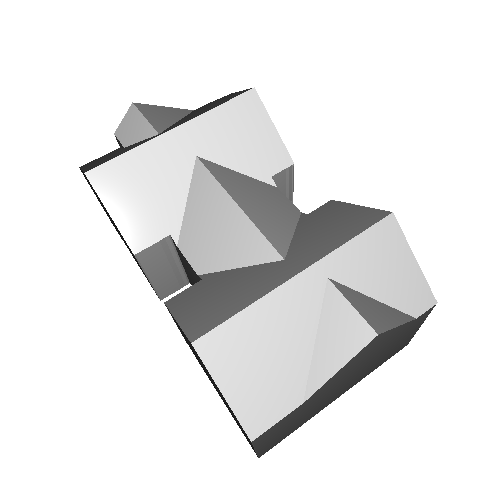}{}{fig:city3d-s-h} \\
    \end{tabular}

    \vspace{-0.7\baselineskip}
    
    \caption{3D reconstruction using different height map pre-processors.  \rd{} closely matches the ground truth (GT).}
    \label{fig:morecity3d}
\end{figure*}

\begin{figure*}[h]
    \newcommand{\repeatdots}[1]{%
      \foreach \n in {1,...,#1} {$\cdot$}%
    }
    \newcolumntype{G}{@{\hskip 1.9pt}}
    \newcolumntype{T}{m{0.015\textwidth}}
    \newcolumntype{I}{>{\centering\arraybackslash}m{\SmallImageSize}}
    \vspace{-\baselineskip}
    \centering

    \begin{tabular}{G T G I G I G I G I G I G I G I G I G I}

        \scriptsize\rotatebox{90}{Input (3D)} & \smallImg{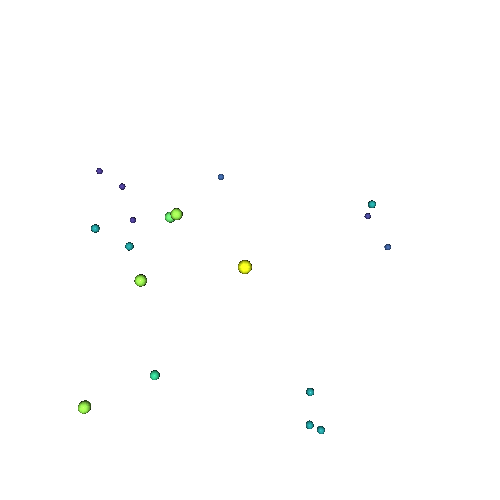} & \smallImg{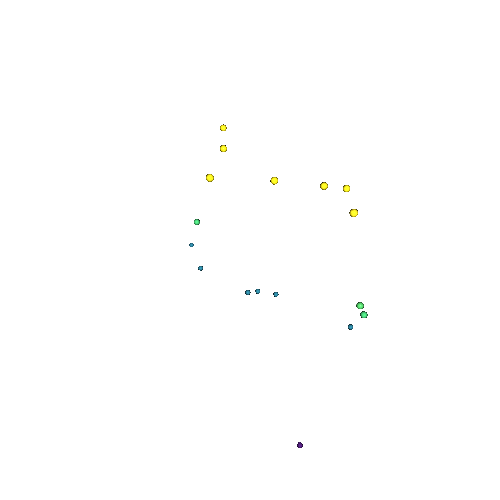} & \smallImg{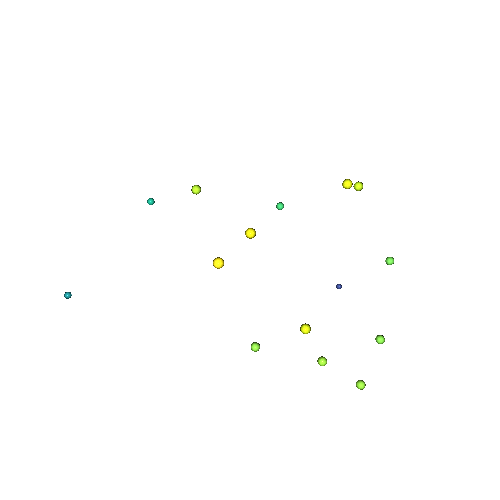} & \smallImg{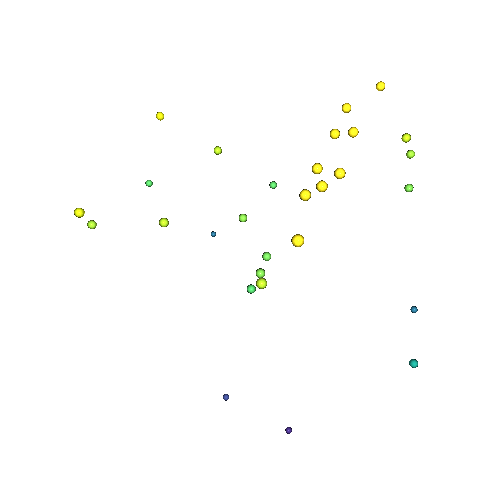} & \smallImg{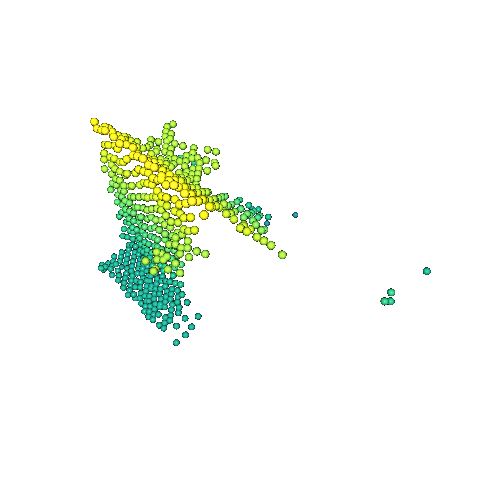} & \smallImg{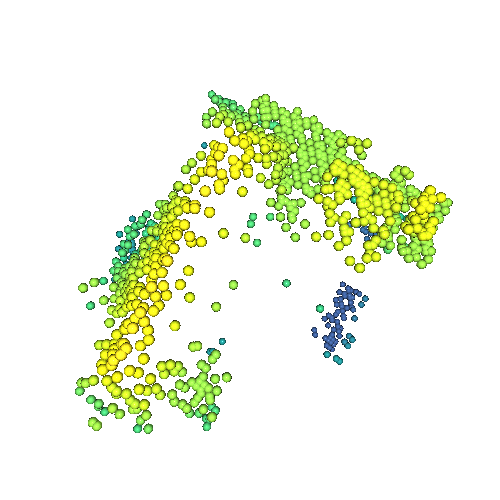} & 
        \smallImg{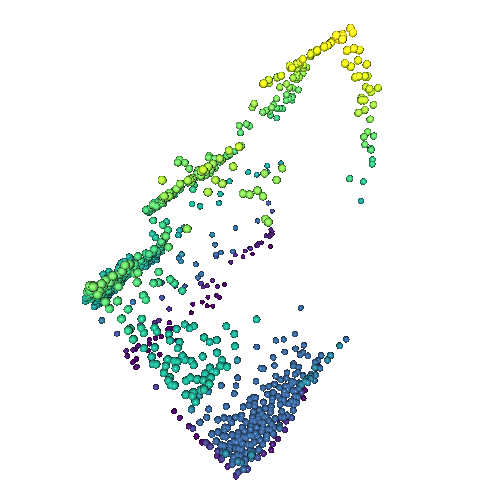} & \smallImg{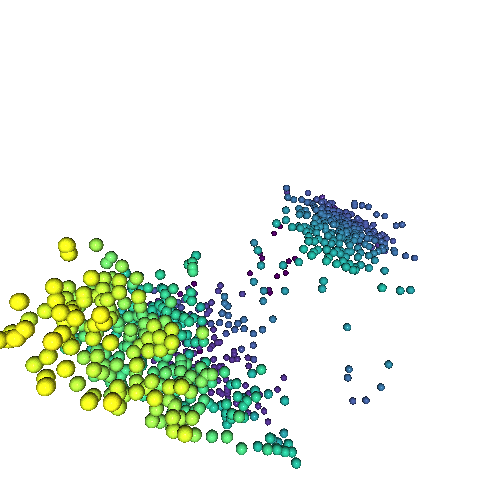} & \smallImg{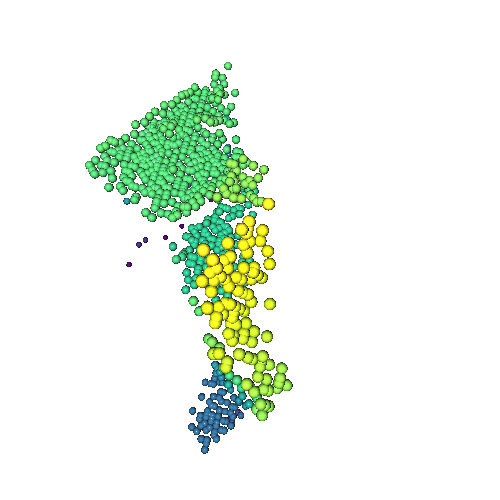} \\[-0.5ex]
    
        \scriptsize\rotatebox{90}{Input} & \smallImg{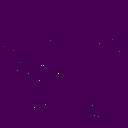} & \smallImg{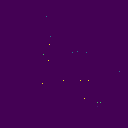} & \smallImg{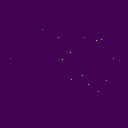} & \smallImg{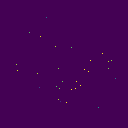} & \smallImg{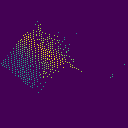} & \smallImg{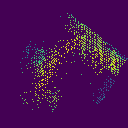} & 
        \smallImg{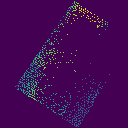} & \smallImg{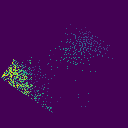} & \smallImg{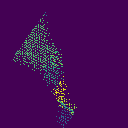} \\[-0.5ex]

        \scriptsize\rotatebox{90}{Linear} & \smallImg{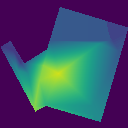} & \smallImg{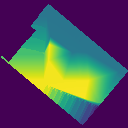} & \smallImg{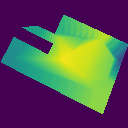} & \smallImg{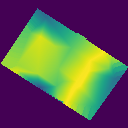} & \smallImg{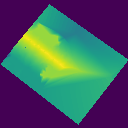} & \smallImg{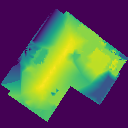} & 
        \smallImg{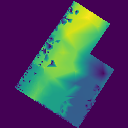} & \smallImg{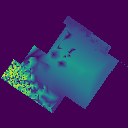} & \smallImg{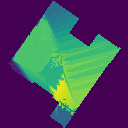} \\[-0.5ex]

        \scriptsize\rotatebox{90}{Nearest} & \smallImg{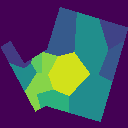} & \smallImg{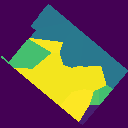} & \smallImg{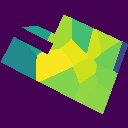} & \smallImg{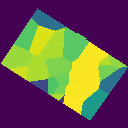} & \smallImg{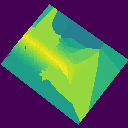} & \smallImg{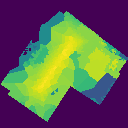} & 
        \smallImg{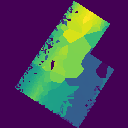} & \smallImg{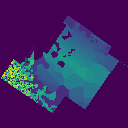} & \smallImg{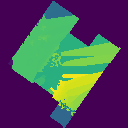} \\[-0.5ex] 

        \scriptsize\rotatebox{90}{Spline} & \smallImg{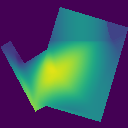} & \smallImg{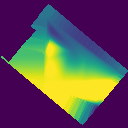} & \smallImg{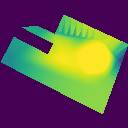} & \smallImg{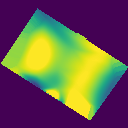} & \smallImg{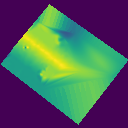} & \smallImg{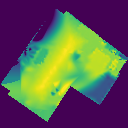} & 
        \smallImg{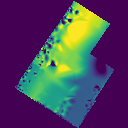} & \smallImg{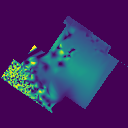} & \smallImg{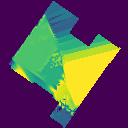} \\[-0.5ex] 
        
        \scriptsize\rotatebox{90}{IDW} & \smallImg{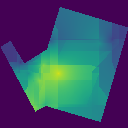} & \smallImg{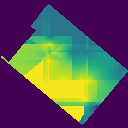} & \smallImg{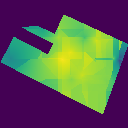} & \smallImg{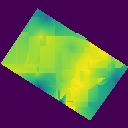} & \smallImg{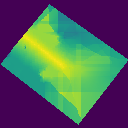} & \smallImg{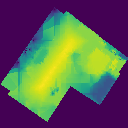} & 
        \smallImg{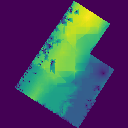} & \smallImg{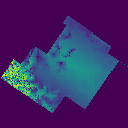} & \smallImg{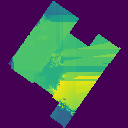} \\[-0.5ex]   

        \scriptsize\rotatebox{90}{P.M. Diff.} & \smallImg{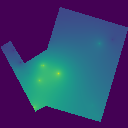} & \smallImg{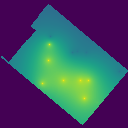} & \smallImg{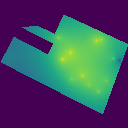} & \smallImg{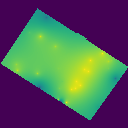} & \smallImg{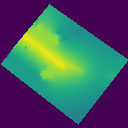} & \smallImg{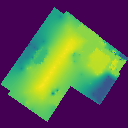} & 
        \smallImg{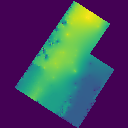} & \smallImg{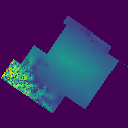} & \smallImg{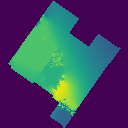} \\[-0.5ex] 

        \scriptsize\rotatebox{90}{Ours} & \smallImg{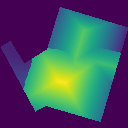} & \smallImg{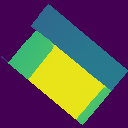} & \smallImg{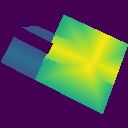} & \smallImg{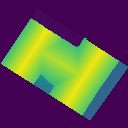} & 
        \smallImg{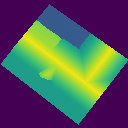} & \smallImg{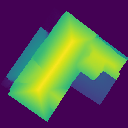} & \smallImg{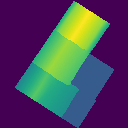} & \smallImg{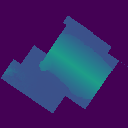} & \smallImg{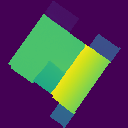} \\[-1ex]   

        \scriptsize\rotatebox{90}{Ours (3D)} & 
        \smallSubImg{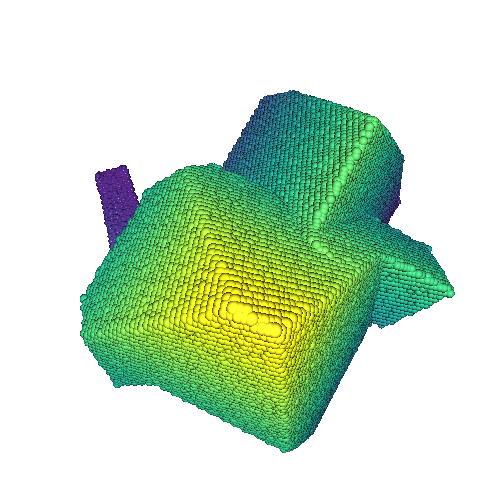}{}{fig:moreAHN3-a} & 
        \smallSubImg{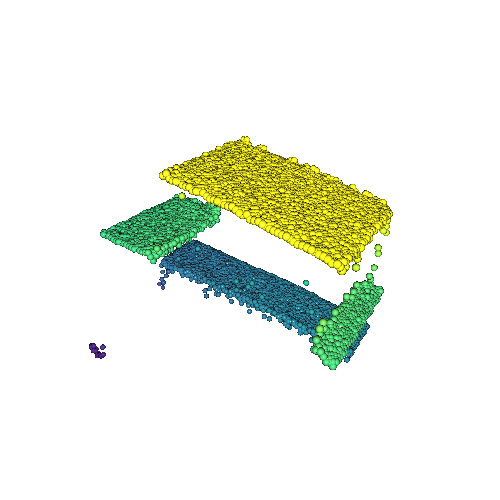}{}{fig:moreAHN3-b} & 
        \smallSubImg{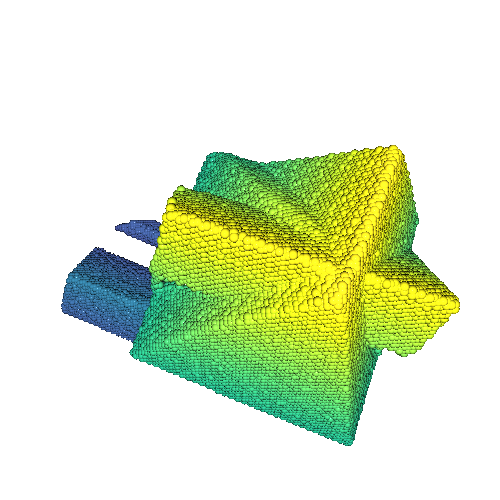}{}{fig:moreAHN3-c} & 
        \smallSubImg{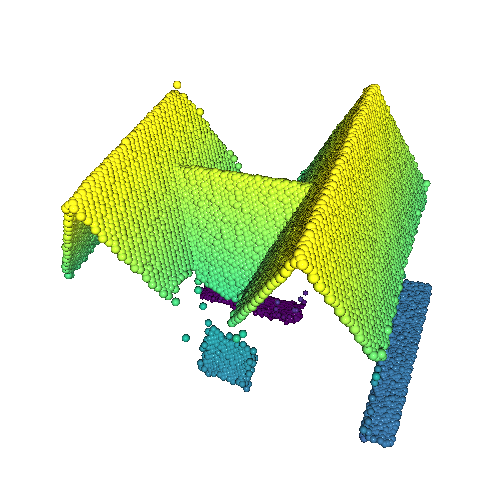}{}{fig:moreAHN3-d} &
        \smallSubImg{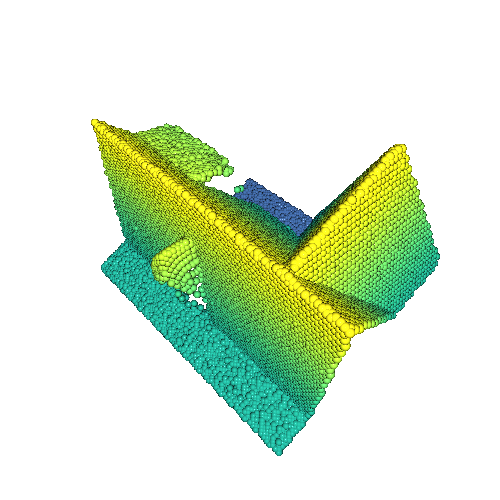}{}{fig:moreAHN3-e} & 
        \smallSubImg{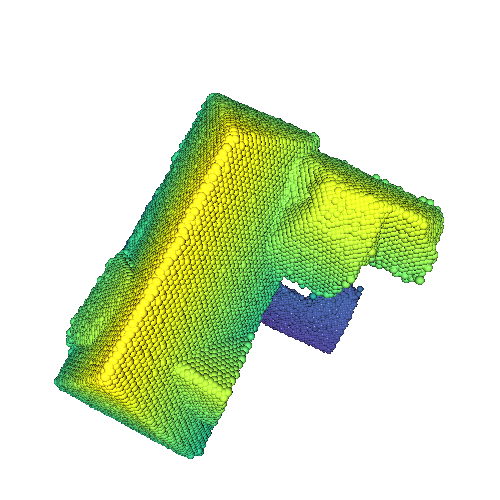}{}{fig:moreAHN3-f} & 
        \smallSubImg{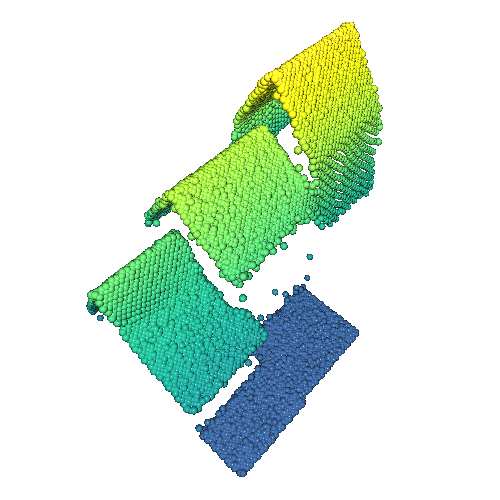}{}{fig:moreAHN3-g} & 
        \smallSubImg{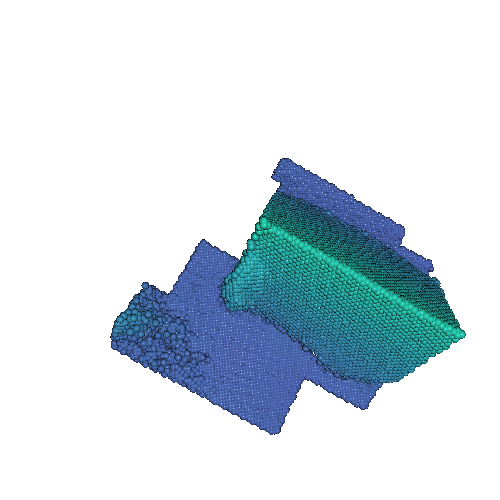}{}{fig:moreAHN3-h} & 
        \smallSubImg{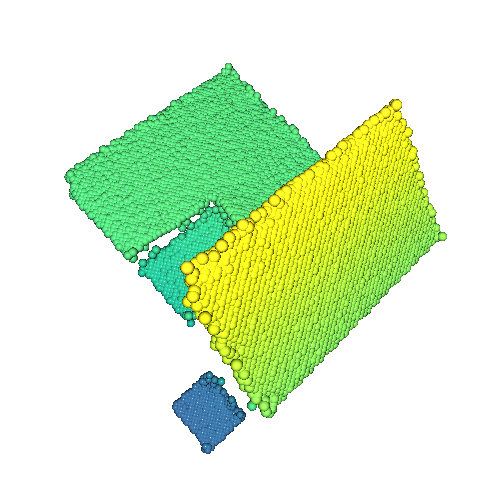}{}{fig:moreAHN3-i} \\ 
        
    \end{tabular}
    \vspace{-0.8\baselineskip} 
    
    \caption{Evaluation of completion and denoising on 
    the AHN3 dataset \cite{huang2022city3d}. Compared with Linear, Nearest, Spline \cite{keys1981cubic}, Inverse Distance Weighting (IDW) \cite{shepard1968two}, and Perona-Malik Diffusion (P.M. Diff.) \cite{perona1990scale,biasutti2019diffusion} interpolation methods.}
    \label{fig:moreAHN3}
    \vspace{-0.5\baselineskip}
\end{figure*}

\begin{figure*}[t]
    \newcommand{\repeatdots}[1]{%
      \foreach \n in {1,...,#1} {$\cdot$}%
    }
    \newcolumntype{G}{@{\hskip 1.9pt}}
    \newcolumntype{T}{m{0.015\textwidth}}
    \newcolumntype{I}{>{\centering\arraybackslash}m{\SmallImageSize}}
    \vspace{-\baselineskip}
    \centering

    \begin{tabular}{G T G I G I G I G I G I G I G I G I G I}

        \scriptsize\rotatebox{90}{Input (3D)} & \smallImg{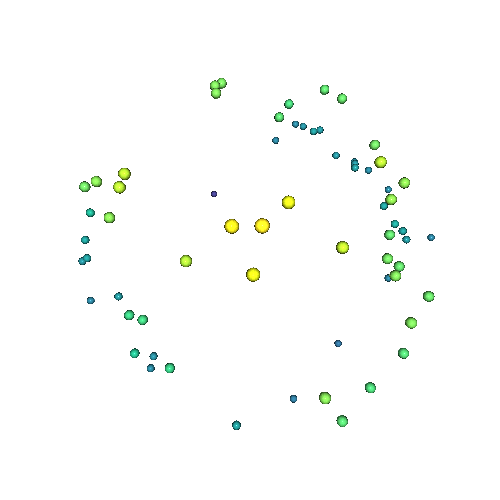} & \smallImg{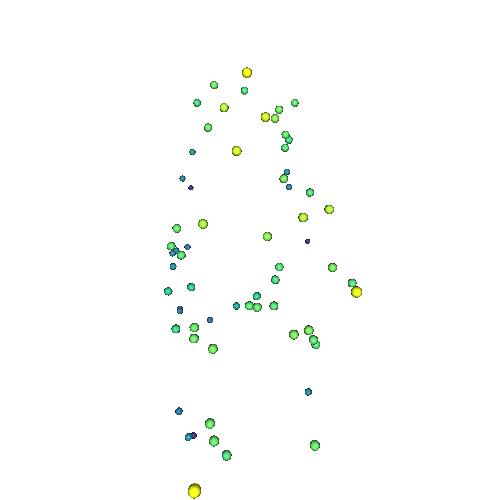} & \smallImg{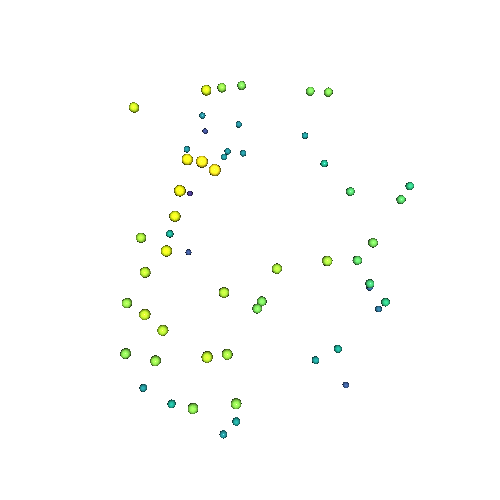} & \smallImg{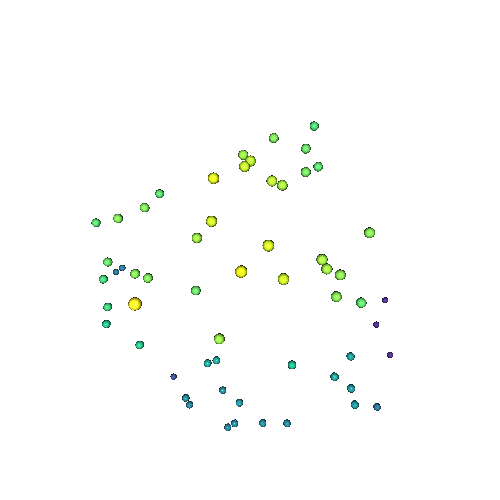} & \smallImg{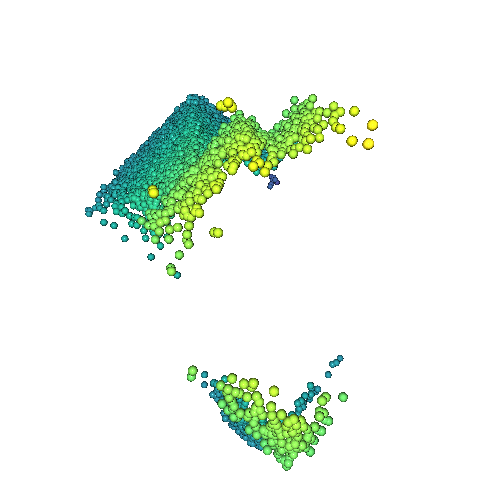} & \smallImg{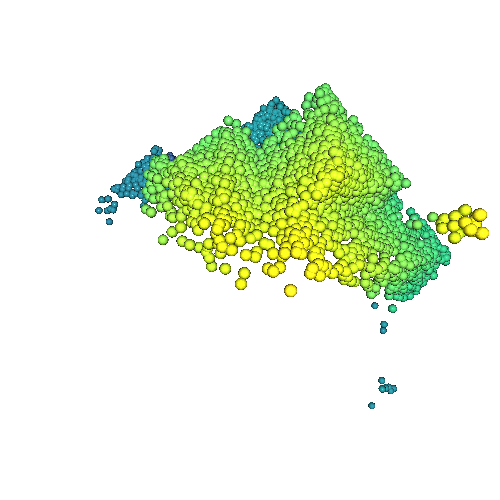} & 
        \smallImg{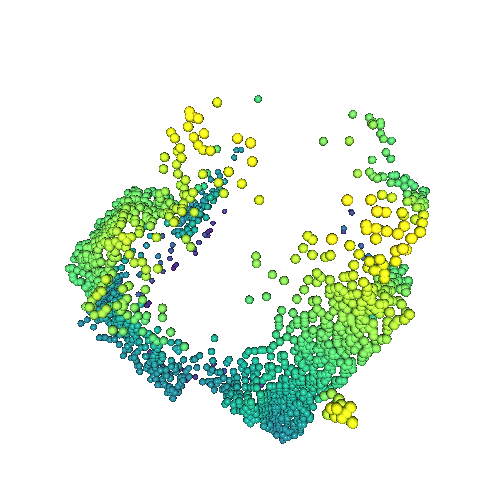} & \smallImg{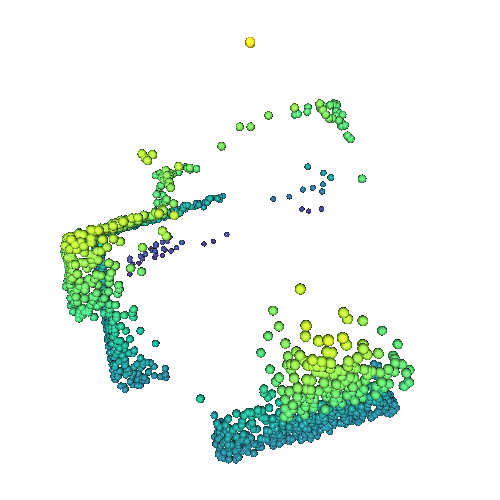} & \smallImg{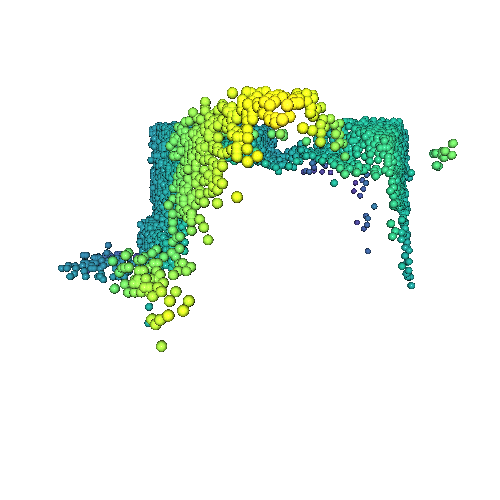} \\[-0.5ex]
    
        \scriptsize\rotatebox{90}{Input} & \smallImg{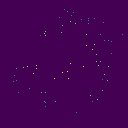} & \smallImg{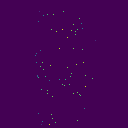} & \smallImg{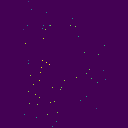} & \smallImg{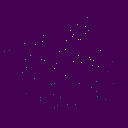} & \smallImg{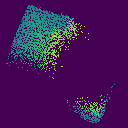} & \smallImg{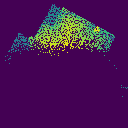} & 
        \smallImg{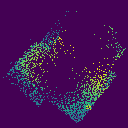} & \smallImg{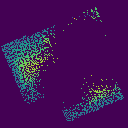} & \smallImg{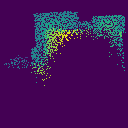} \\[-0.5ex]

        \scriptsize\rotatebox{90}{Linear} & \smallImg{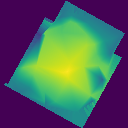} & \smallImg{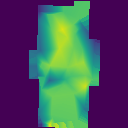} & \smallImg{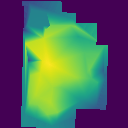} & \smallImg{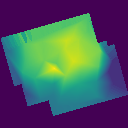} & \smallImg{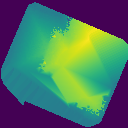} & \smallImg{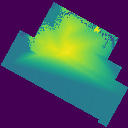} & 
        \smallImg{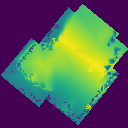} & \smallImg{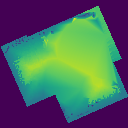} & \smallImg{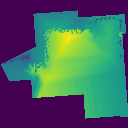} \\[-0.5ex]

        \scriptsize\rotatebox{90}{Nearest} & \smallImg{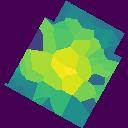} & \smallImg{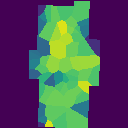} & \smallImg{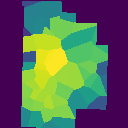} & \smallImg{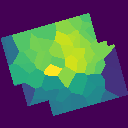} & \smallImg{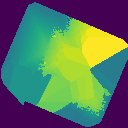} & \smallImg{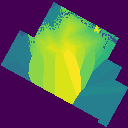} & 
        \smallImg{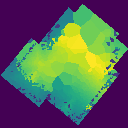} & \smallImg{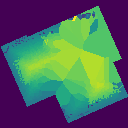} & \smallImg{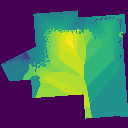} \\[-0.5ex]

        \scriptsize\rotatebox{90}{Spline} & \smallImg{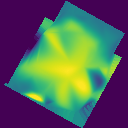} & \smallImg{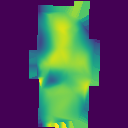} & \smallImg{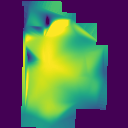} & \smallImg{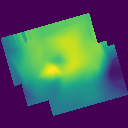} & \smallImg{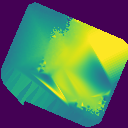} & \smallImg{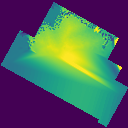} & 
        \smallImg{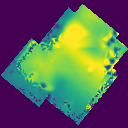} & \smallImg{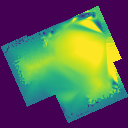} & \smallImg{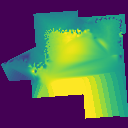} \\[-0.5ex]
        
        \scriptsize\rotatebox{90}{IDW} & \smallImg{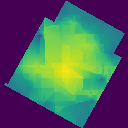} & \smallImg{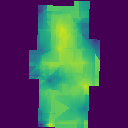} & \smallImg{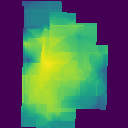} & \smallImg{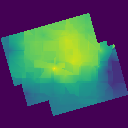} & \smallImg{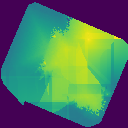} & \smallImg{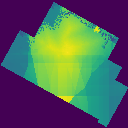} & 
        \smallImg{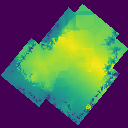} & \smallImg{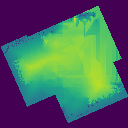} & \smallImg{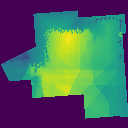} \\[-0.5ex]

        \scriptsize\rotatebox{90}{P.M. Diff.} & \smallImg{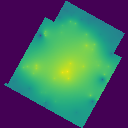} & \smallImg{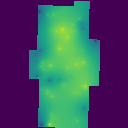} & \smallImg{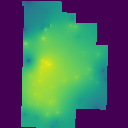} & \smallImg{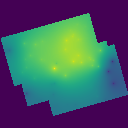} & \smallImg{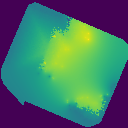} & \smallImg{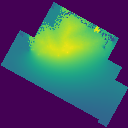} & 
        \smallImg{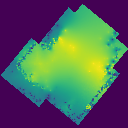} & \smallImg{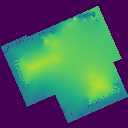} & \smallImg{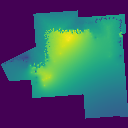} \\[-0.5ex]
        
        \scriptsize\rotatebox{90}{Ours} & \smallImg{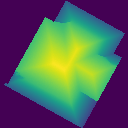} & \smallImg{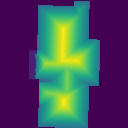} & \smallImg{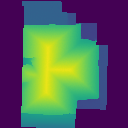} & \smallImg{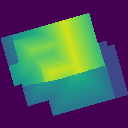} & \smallImg{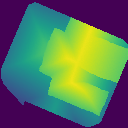} & \smallImg{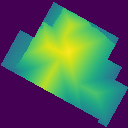} & 
        \smallImg{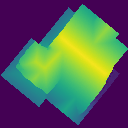} & \smallImg{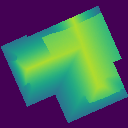} & \smallImg{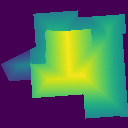} \\[-0.5ex]

        \scriptsize\rotatebox{90}{Ours (3D)} & \smallSubImg{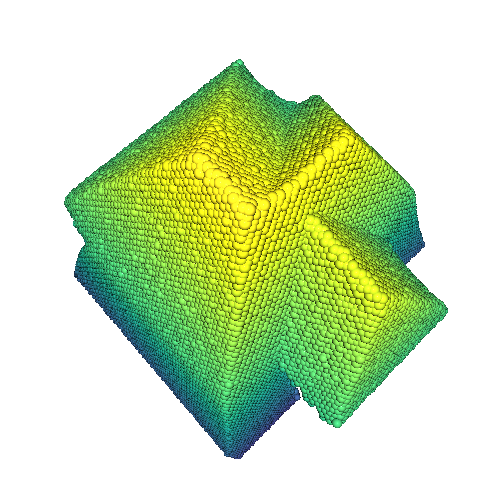}{}{fig:moreDales3d-a} & \smallSubImg{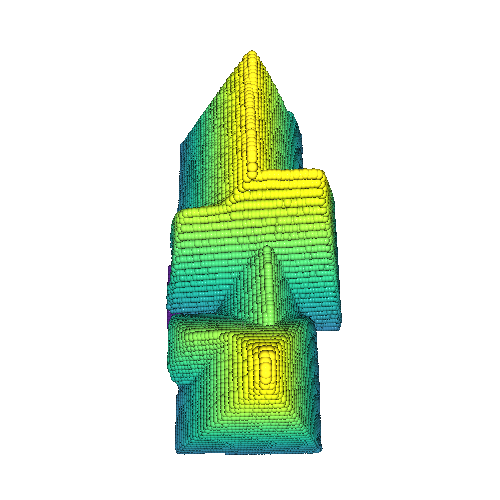}{}{fig:moreDales3d-b} & \smallSubImg{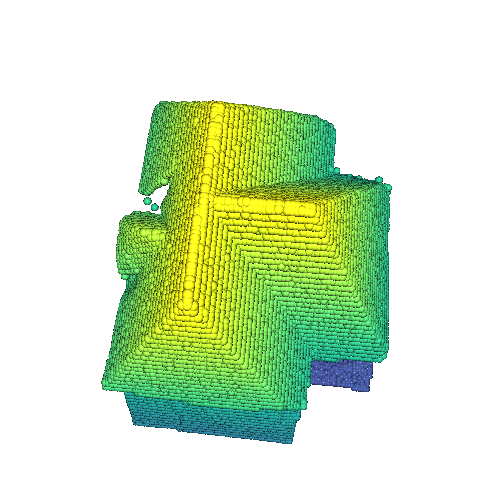}{}{fig:moreDales3d-c} & \smallSubImg{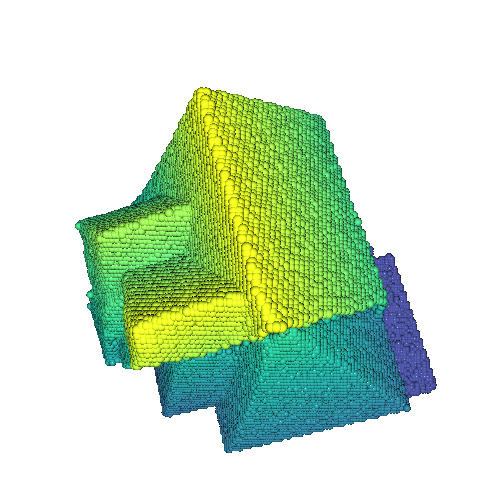}{}{fig:moreDales3d-d} &
        \smallSubImg{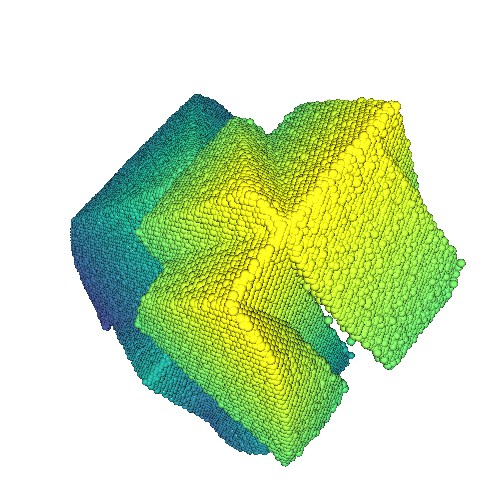}{}{fig:moreDales3d-e} & \smallSubImg{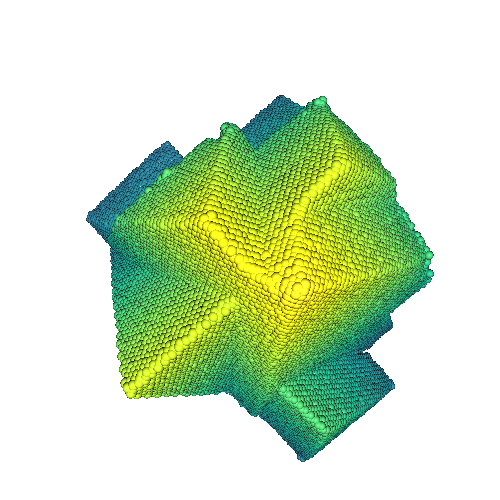}{}{fig:moreDales3d-f} & 
        \smallSubImg{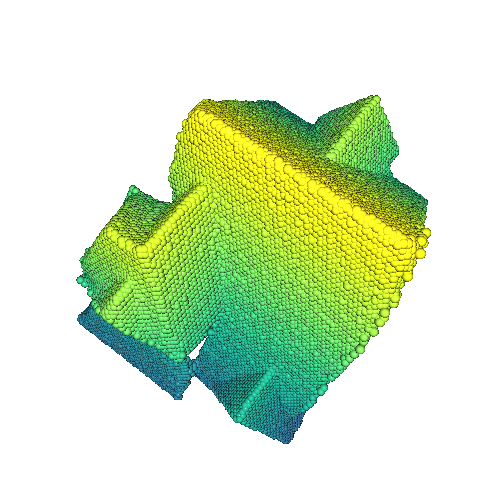}{}{fig:moreDales3d-g} & \smallSubImg{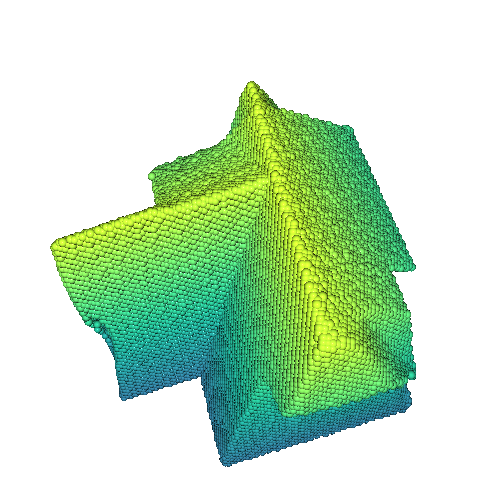}{}{fig:moreDales3d-h} & \smallSubImg{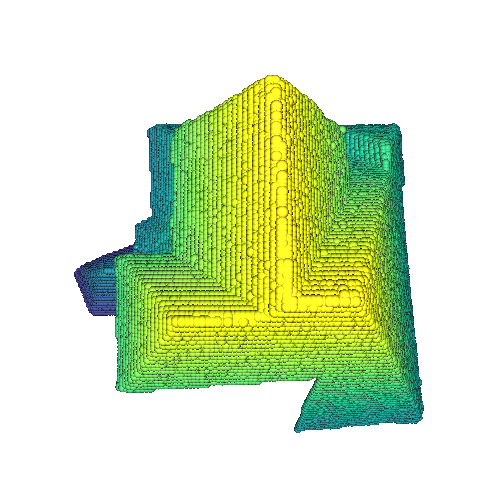}{}{fig:moreDales3d-i} \\ 
        
    \end{tabular}
    \vspace{-0.8\baselineskip} 
    \caption{Evaluation of the completion and denoising on the Dales3D dataset \cite{varney2020dales}.}
    \label{fig:moreDales3d}
    \vspace{-0.5\baselineskip}
\end{figure*}

\begin{figure*}[t]
    \newcommand{\repeatdots}[1]{%
      \foreach \n in {1,...,#1} {$\cdot$}%
    }
    \newcolumntype{G}{@{\hskip 1.9pt}}
    \newcolumntype{T}{m{0.015\textwidth}}
    \newcolumntype{I}{>{\centering\arraybackslash}m{\SmallImageSize}}
    \vspace{-\baselineskip}
    \centering

    \begin{tabular}{G T G I G I G I G I G I G I G I G I G I}

        \scriptsize\rotatebox{90}{Input (3D)} & \smallImg{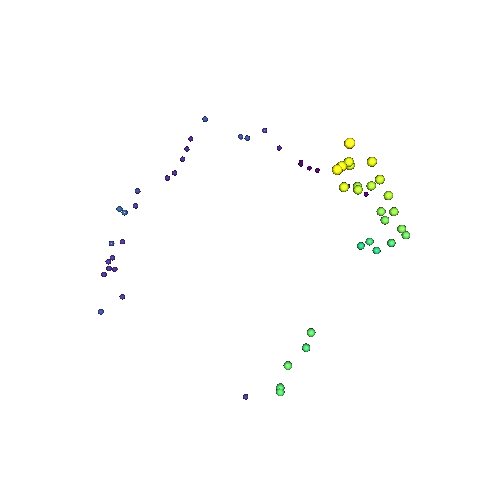} & \smallImg{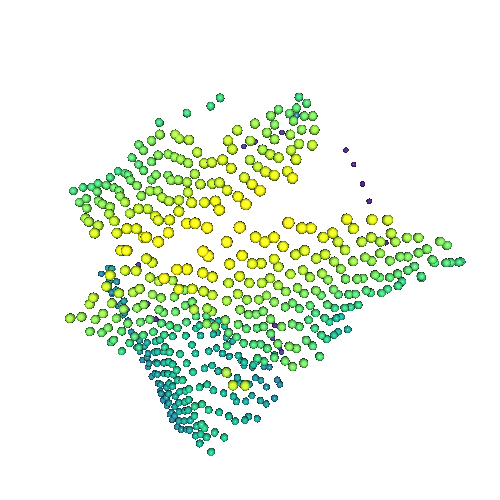} & \smallImg{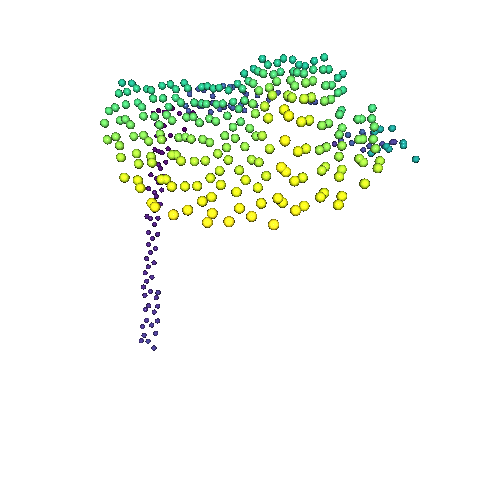} & \smallImg{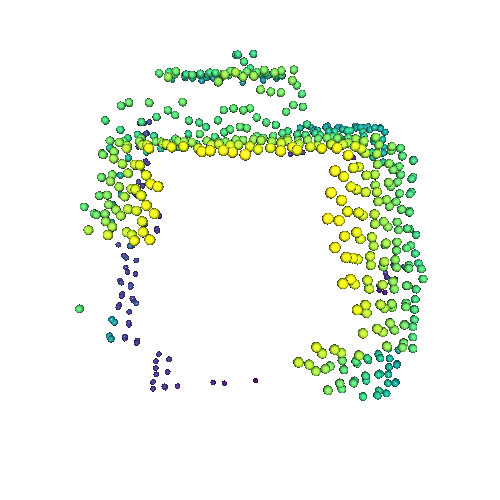} & \smallImg{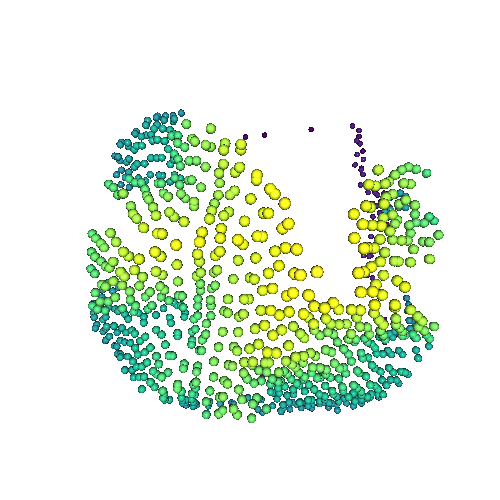} & \smallImg{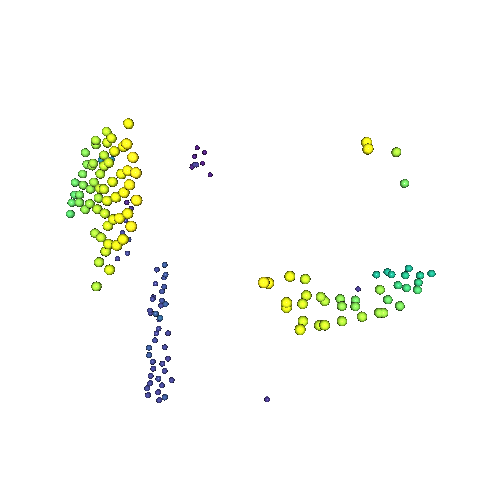} & 
        \smallImg{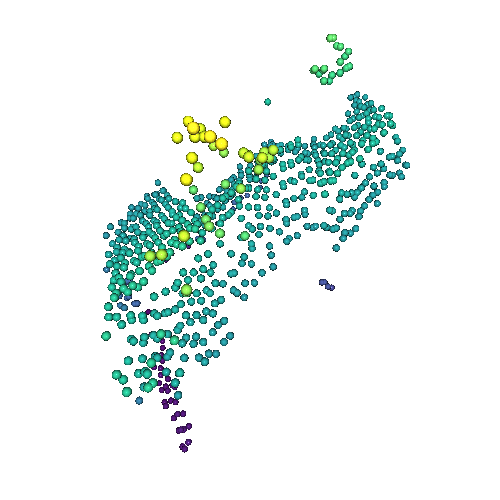} & \smallImg{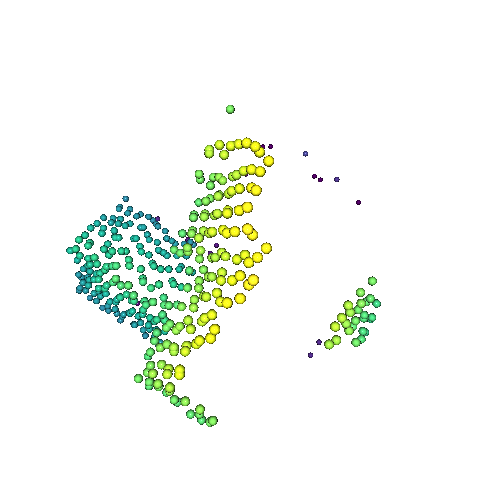} & \smallImg{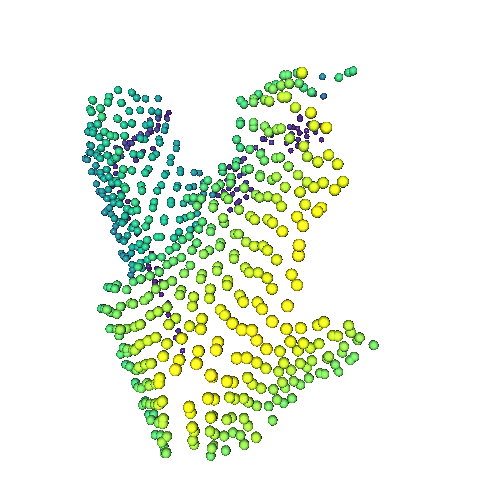} \\[-0.5ex]
    
        \scriptsize\rotatebox{90}{Input} & \smallImg{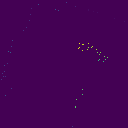} & \smallImg{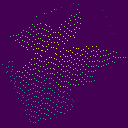} & \smallImg{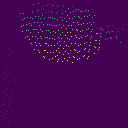} & \smallImg{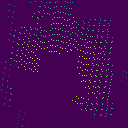} & \smallImg{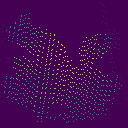} & \smallImg{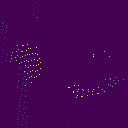} & 
        \smallImg{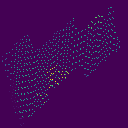} & \smallImg{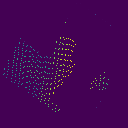} & \smallImg{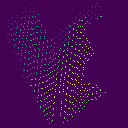} \\[-0.5ex]

        \scriptsize\rotatebox{90}{Linear} & \smallImg{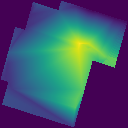} & \smallImg{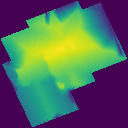} & \smallImg{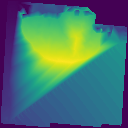} & \smallImg{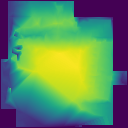} & \smallImg{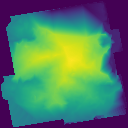} & \smallImg{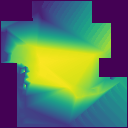} & 
        \smallImg{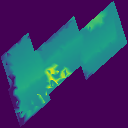} & \smallImg{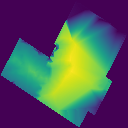} & \smallImg{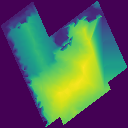} \\[-0.5ex]

        \scriptsize\rotatebox{90}{Nearest} & \smallImg{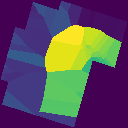} & \smallImg{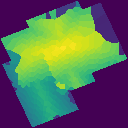} & \smallImg{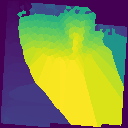} & \smallImg{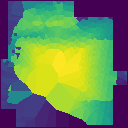} & \smallImg{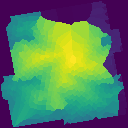} & \smallImg{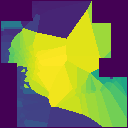} & 
        \smallImg{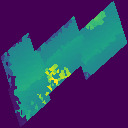} & \smallImg{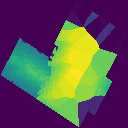} & \smallImg{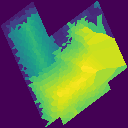} \\[-0.5ex] 

        \scriptsize\rotatebox{90}{Spline} & \smallImg{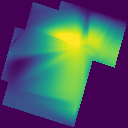} & \smallImg{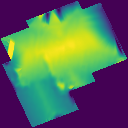} & \smallImg{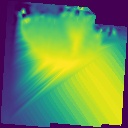} & \smallImg{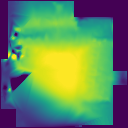} & \smallImg{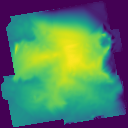} & \smallImg{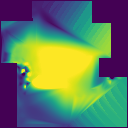} & 
        \smallImg{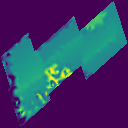} & \smallImg{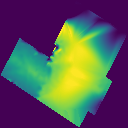} & \smallImg{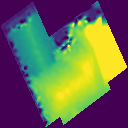} \\[-0.5ex] 
        
        \scriptsize\rotatebox{90}{IDW} & \smallImg{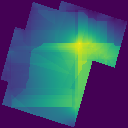} & \smallImg{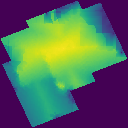} & \smallImg{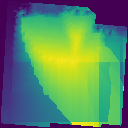} & \smallImg{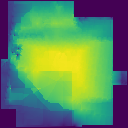} & \smallImg{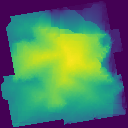} & \smallImg{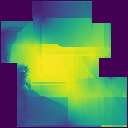} & 
        \smallImg{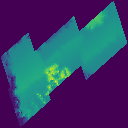} & \smallImg{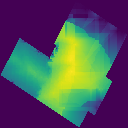} & \smallImg{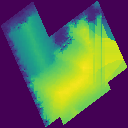} \\[-0.5ex]   

        \scriptsize\rotatebox{90}{P.M. Diff.} & \smallImg{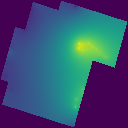} & \smallImg{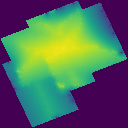} & \smallImg{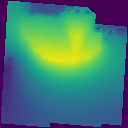} & \smallImg{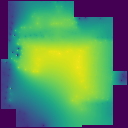} & \smallImg{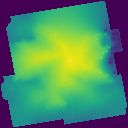} & \smallImg{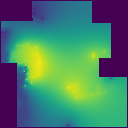} & 
        \smallImg{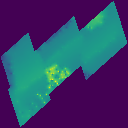} & \smallImg{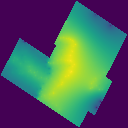} & \smallImg{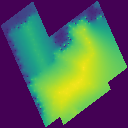} \\[-0.5ex] 

        \scriptsize\rotatebox{90}{Ours} & \smallImg{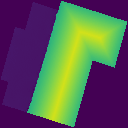} & \smallImg{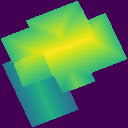} & \smallImg{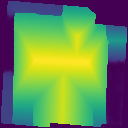} & \smallImg{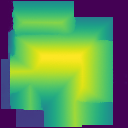} & 
        \smallImg{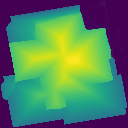} & \smallImg{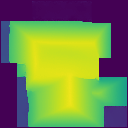} & \smallImg{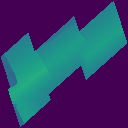} & \smallImg{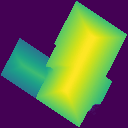} & \smallImg{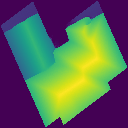} \\[-1ex]   

        \scriptsize\rotatebox{90}{Ours (3D)} & \smallSubImg{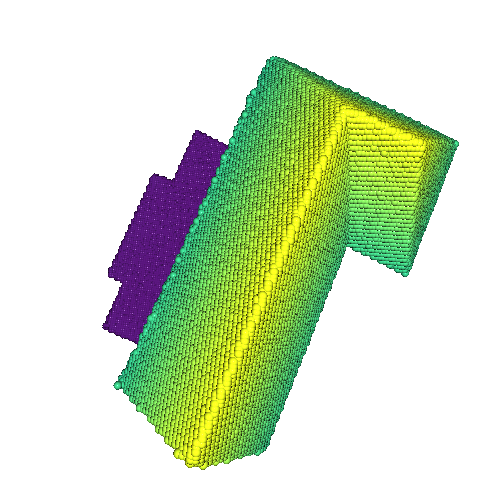}{}{} & \smallSubImg{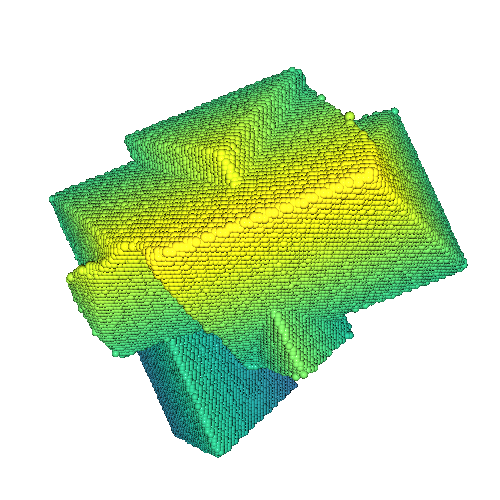}{}{} & \smallSubImg{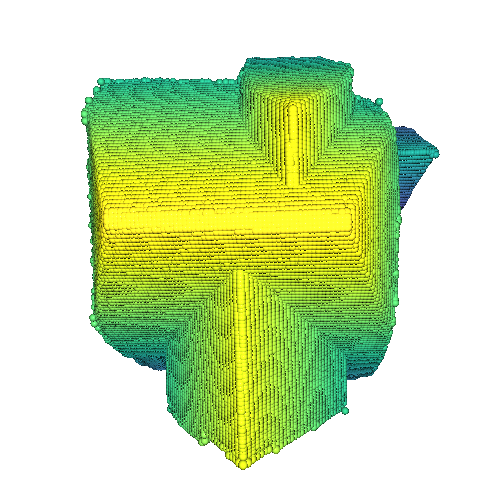}{}{} & \smallSubImg{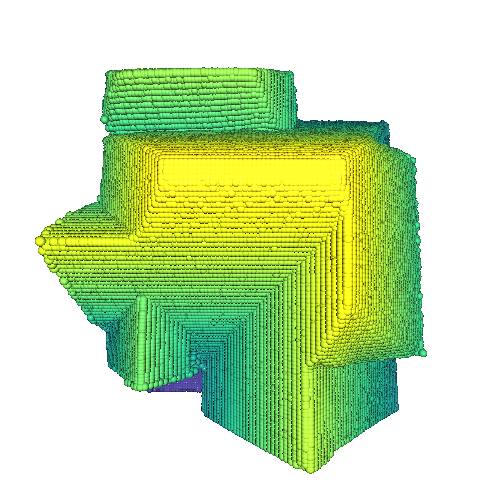}{}{} &
        \smallSubImg{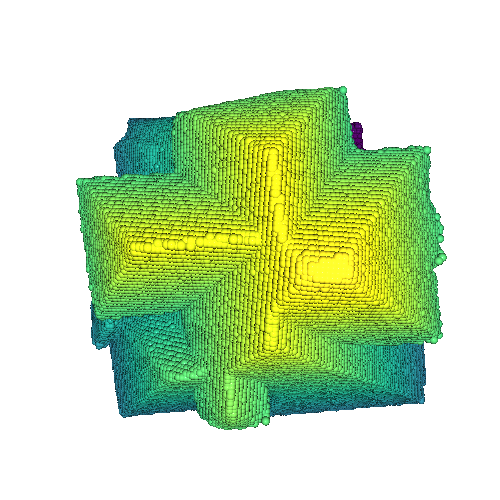}{}{} & \smallSubImg{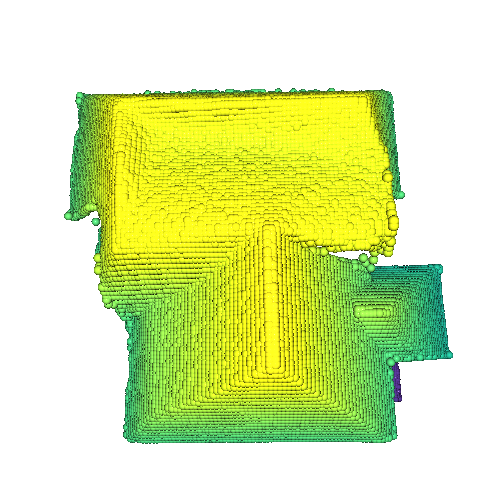}{}{} & 
        \smallSubImg{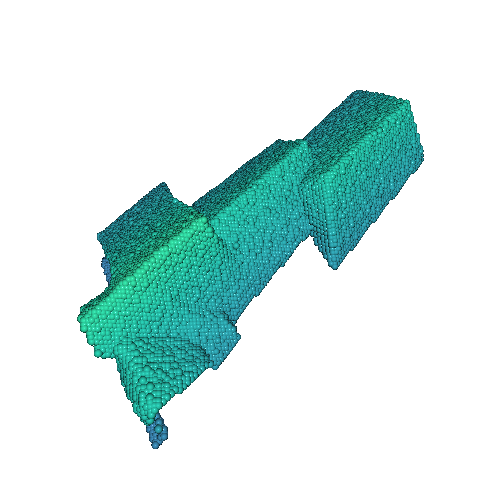}{}{} & \smallSubImg{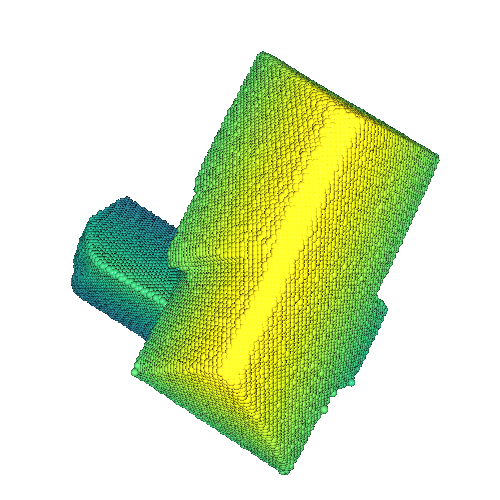}{}{} & \smallSubImg{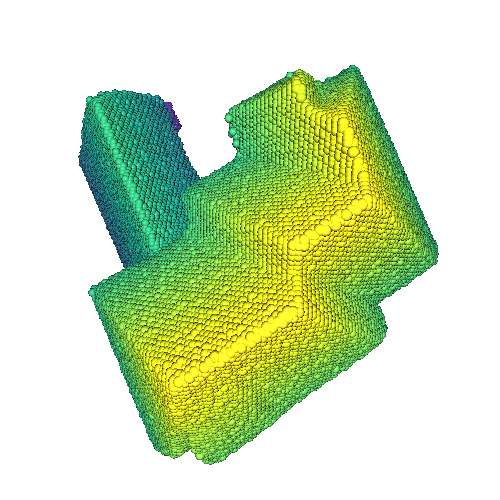}{}{} \\ 
        
    \end{tabular}
    \vspace{-0.8\baselineskip} 
    
    \caption{Evaluation of the completion and denoising on USGS 3DEP LiDAR data sampled over Wayne County, MI \cite{wayneco2017}.}
    \label{fig:morewayneco}
    \vspace{-0.5\baselineskip}
\end{figure*}

\begin{figure*}[t]
    \newcommand{\repeatdots}[1]{%
      \foreach \n in {1,...,#1} {$\cdot$}%
    }
    \newcolumntype{G}{@{\hskip 1.9pt}}
    \newcolumntype{T}{m{0.015\textwidth}}
    \newcolumntype{I}{>{\centering\arraybackslash}m{\SmallImageSize}}
    \vspace{-\baselineskip}
    \centering

    \begin{tabular}{G T G I G I G I G I G I G I G I G I G I}


        \scriptsize\rotatebox{90}{Input (3D)} & \smallImg{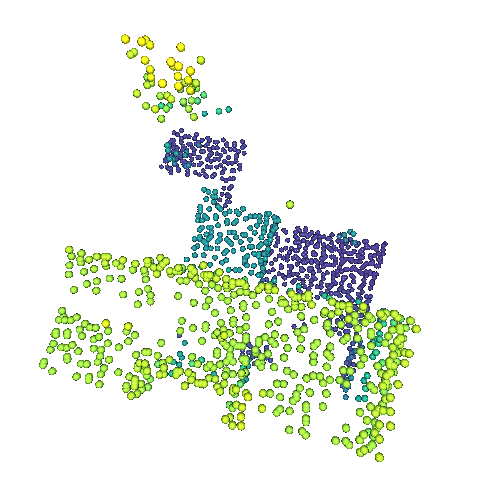} & \smallImg{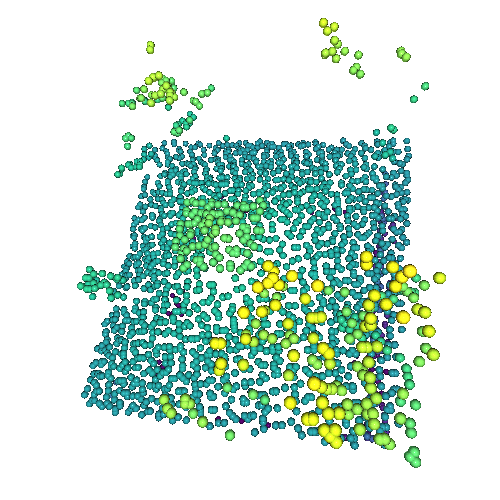} & \smallImg{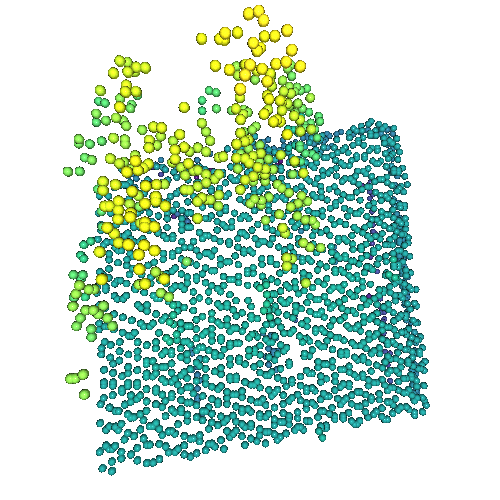} & \smallImg{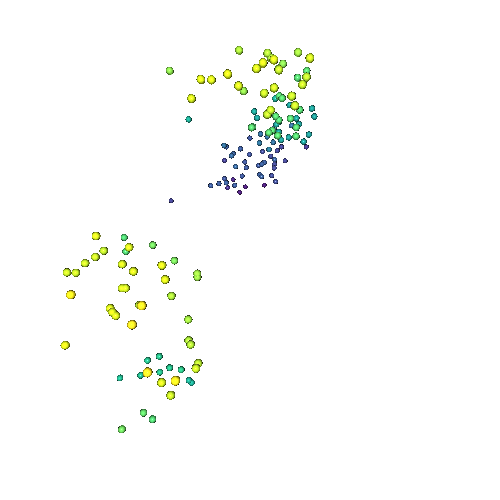} & \smallImg{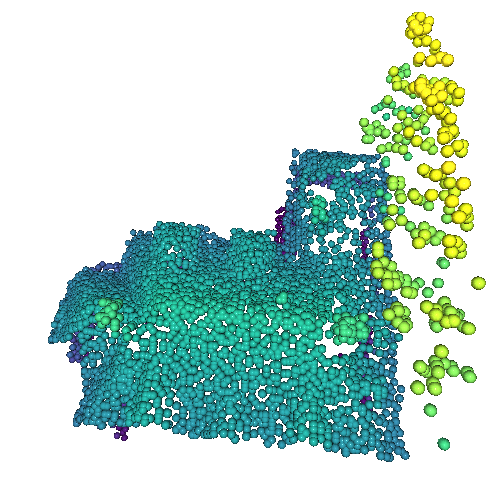} & \smallImg{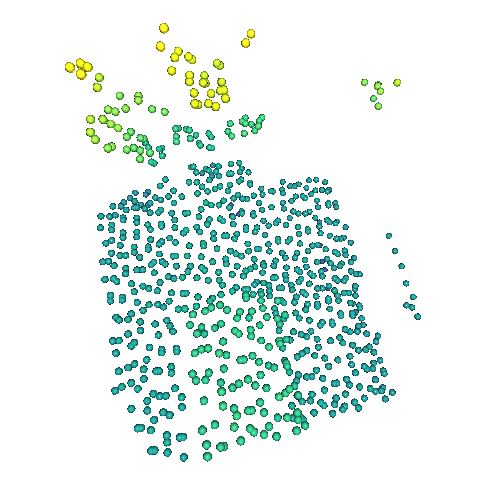} & 
        \smallImg{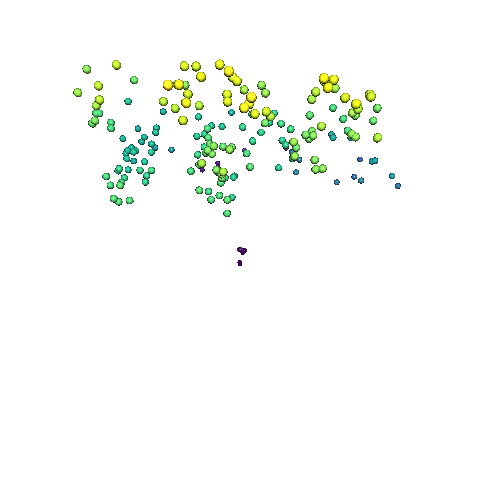} & \smallImg{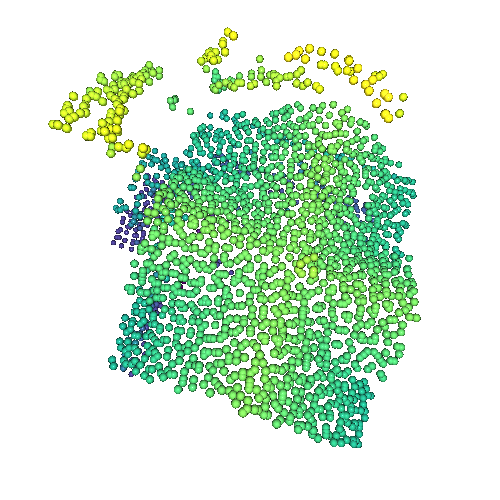} & \smallImg{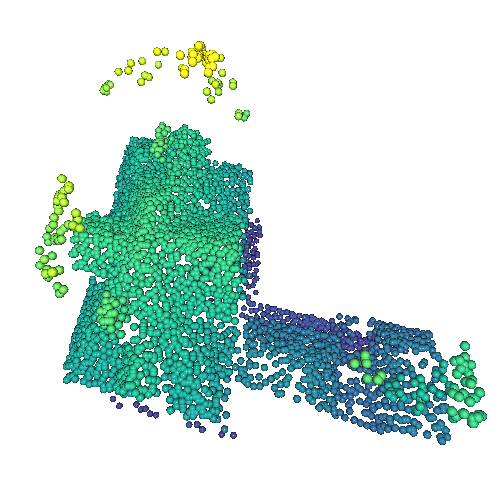} \\[-0.5ex]
    
        \scriptsize\rotatebox{90}{Input} & \smallImg{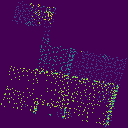} & \smallImg{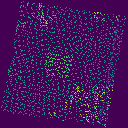} & \smallImg{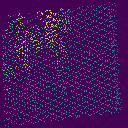} & \smallImg{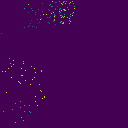} & \smallImg{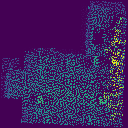} & \smallImg{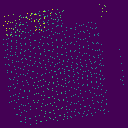} & 
        \smallImg{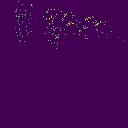} & \smallImg{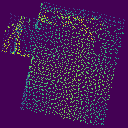} & \smallImg{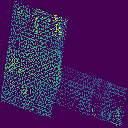} \\[-0.5ex]

        \scriptsize\rotatebox{90}{Linear} & \smallImg{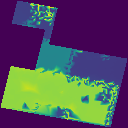} & \smallImg{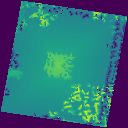} & \smallImg{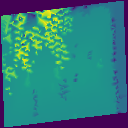} & \smallImg{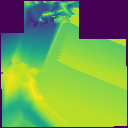} & \smallImg{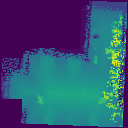} & \smallImg{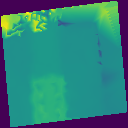} & 
        \smallImg{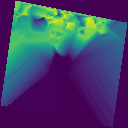} & \smallImg{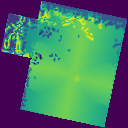} & \smallImg{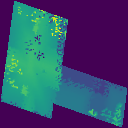} \\[-0.5ex]

        \scriptsize\rotatebox{90}{Nearest} & \smallImg{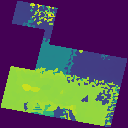} & \smallImg{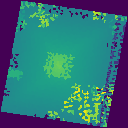} & \smallImg{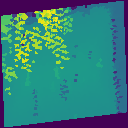} & \smallImg{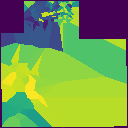} & \smallImg{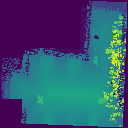} & \smallImg{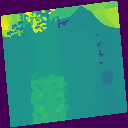} & 
        \smallImg{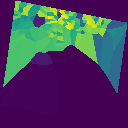} & \smallImg{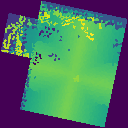} & \smallImg{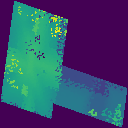} \\[-0.5ex] 

        \scriptsize\rotatebox{90}{Spline} & \smallImg{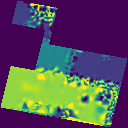} & \smallImg{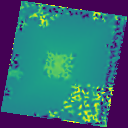} & \smallImg{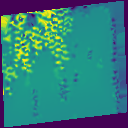} & \smallImg{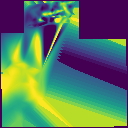} & \smallImg{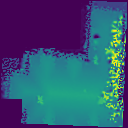} & \smallImg{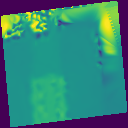} & 
        \smallImg{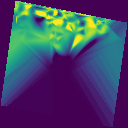} & \smallImg{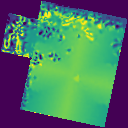} & \smallImg{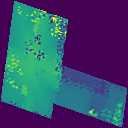} \\[-0.5ex] 
        
        \scriptsize\rotatebox{90}{IDW} & \smallImg{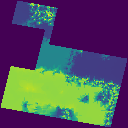} & \smallImg{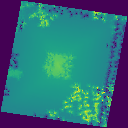} & \smallImg{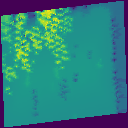} & \smallImg{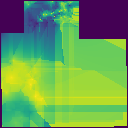} & \smallImg{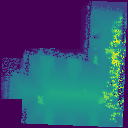} & \smallImg{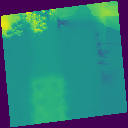} & 
        \smallImg{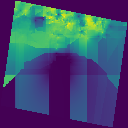} & \smallImg{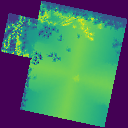} & \smallImg{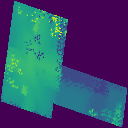} \\[-0.5ex]  

        \scriptsize\rotatebox{90}{P.M. Diff.} & \smallImg{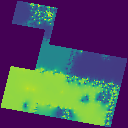} & \smallImg{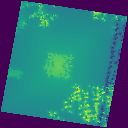} & \smallImg{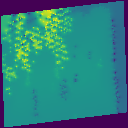} & \smallImg{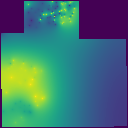} & \smallImg{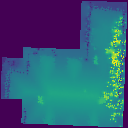} & \smallImg{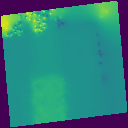} & 
        \smallImg{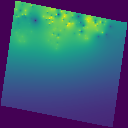} & \smallImg{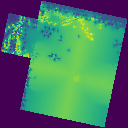} & \smallImg{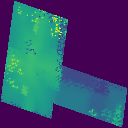} \\[-0.5ex]  

        \scriptsize\rotatebox{90}{Ours} & \smallImg{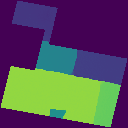} & \smallImg{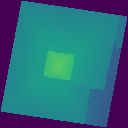} & \smallImg{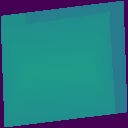} & \smallImg{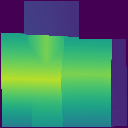} & 
        \smallImg{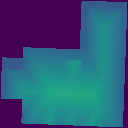} & \smallImg{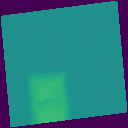} & \smallImg{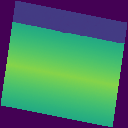} & \smallImg{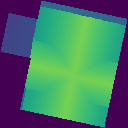} & \smallImg{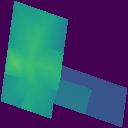} \\[-1ex]   

        \scriptsize\rotatebox{90}{Ours (3D)} & 
        \smallSubImg{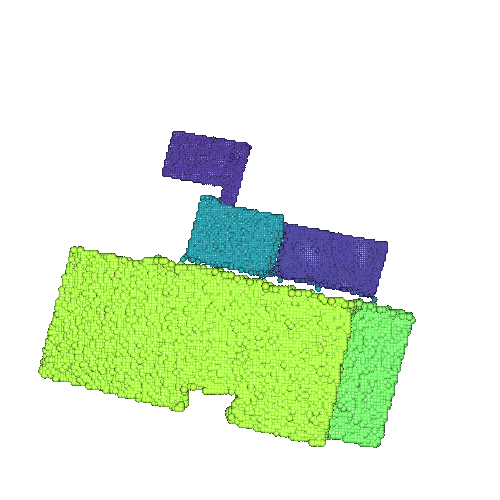}{}{fig:morecambridge-a} & 
        \smallSubImg{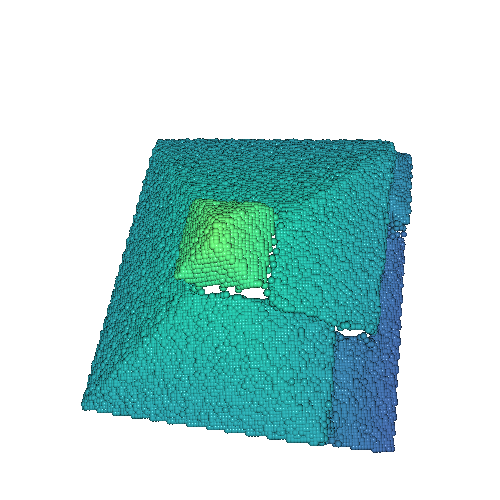}{}{fig:morecambridge-b} & 
        \smallSubImg{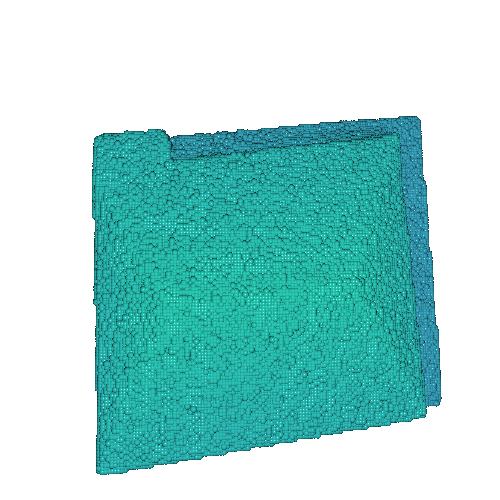}{}{fig:morecambridge-c} & 
        \smallSubImg{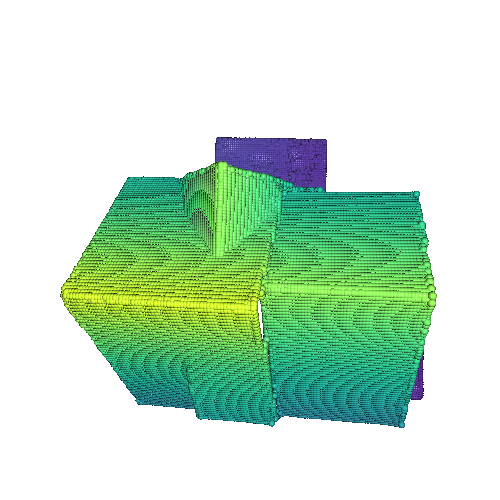}{}{fig:morecambridge-d} &
        \smallSubImg{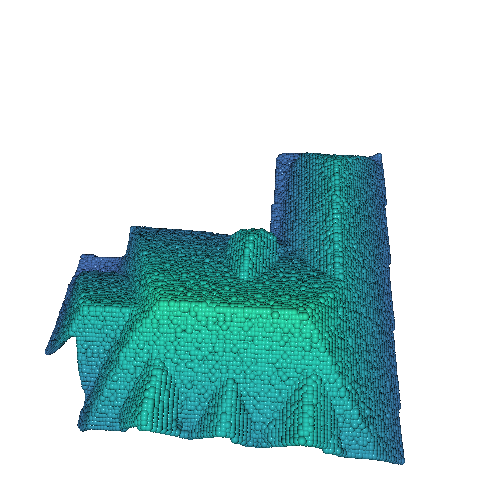}{}{fig:morecambridge-e} & 
        \smallSubImg{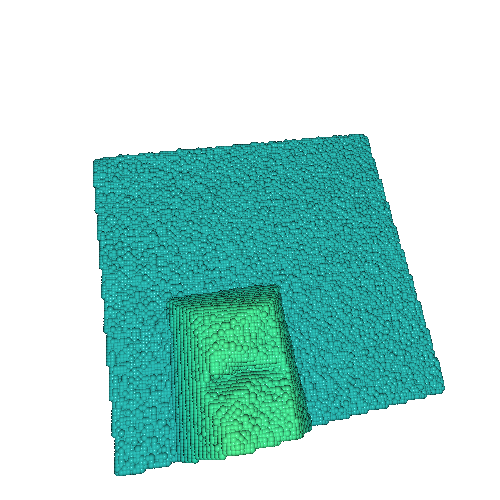}{}{fig:morecambridge-f} & 
        \smallSubImg{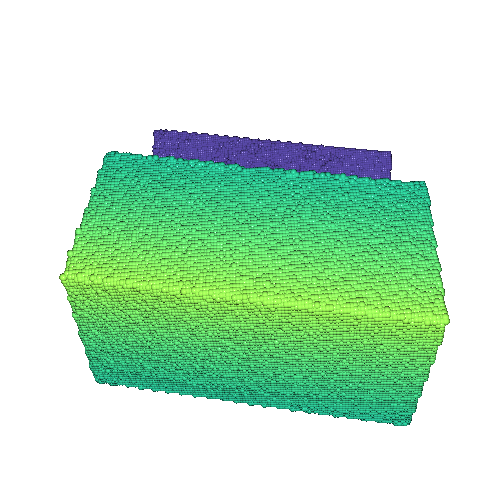}{}{fig:morecambridge-g} & 
        \smallSubImg{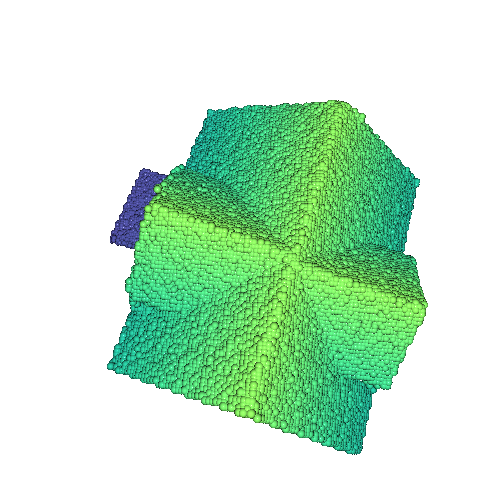}{}{fig:morecambridge-h} & 
        \smallSubImg{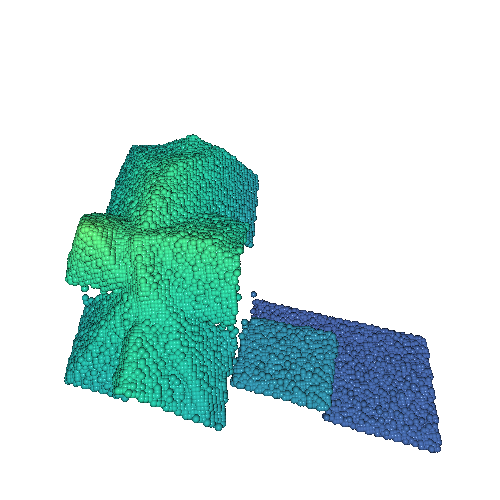}{}{fig:morecambridge-i} \\ 
        
    \end{tabular}
    \vspace{-0.8\baselineskip} 
    
    \caption{Evaluation of the completion and denoising on USGS 3DEP LiDAR data sampled over Cambridge, MA \cite{cambridge2021}.}
    \label{fig:morecambridge}
    \vspace{-0.5\baselineskip}
\end{figure*}

\noindent \textbf{\rd.}
\begin{itemize}
    \item \Cref{fig:morecity3d} showcases examples of applying various height map pre-processors for 3D reconstruction algorithm, City3D \cite{huang2022city3d}, on our \pdata.
    \item \Cref{fig:moreAHN3} illustrates the evaluation on AHN3 dataset \cite{huang2022city3d}.
    \item \Cref{fig:moreDales3d} shows the evaluation on Dales3D dataset \cite{varney2020dales}.
    \item \Cref{fig:morewayneco} provides the results on USGS 3DEP LiDAR sampled over Wayne County, MI \cite{wayneco2017}.
    \item \Cref{fig:morecambridge} provides the results on USGS 3DEP LiDAR sampled over Cambridge, MA \cite{cambridge2021}.
\end{itemize}

\noindent \textbf{No-FP \rd.}\footnote{Since the point clouds in AHN3 \cite{huang2022city3d} and Dales3D \cite{varney2020dales} are already cropped by footprints, we do not include these two datasets in our testing.}

\begin{figure*}[t]
    \newcommand{\repeatdots}[1]{%
      \foreach \n in {1,...,#1} {$\cdot$}%
    }
    \newcolumntype{G}{@{\hskip 1.9pt}}
    \newcolumntype{T}{m{0.015\textwidth}}
    \newcolumntype{I}{>{\centering\arraybackslash}m{\SmallImageSize}}
    \vspace{-\baselineskip}
    \centering

    \begin{tabular}{G T G I G I G I G I G I G I G I G I G I}


        \scriptsize\rotatebox{90}{Input (3D)} & 
        \smallImg{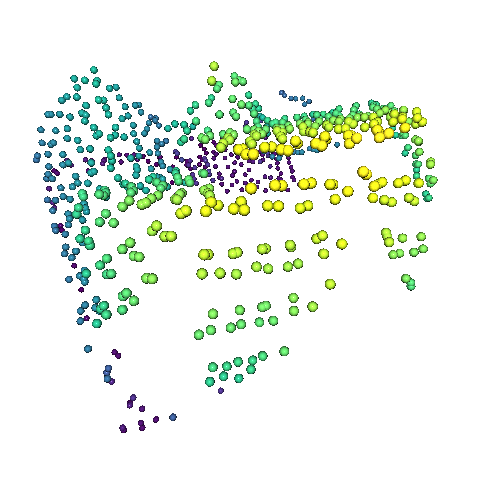} & 
        \smallImg{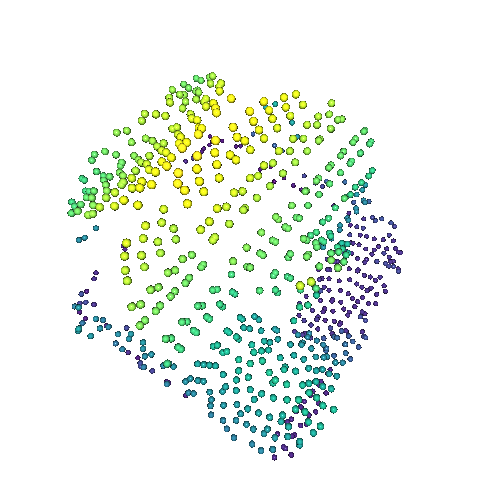} & 
        \smallImg{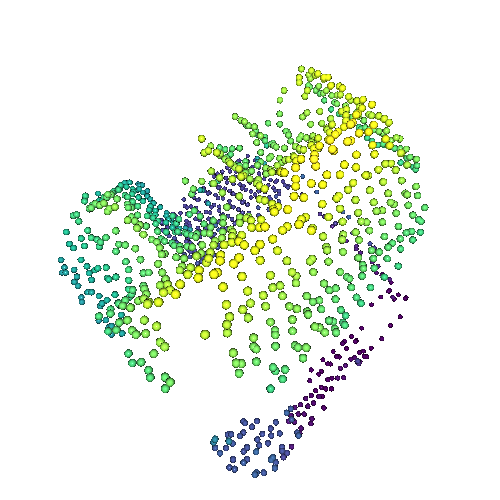} & 
        \smallImg{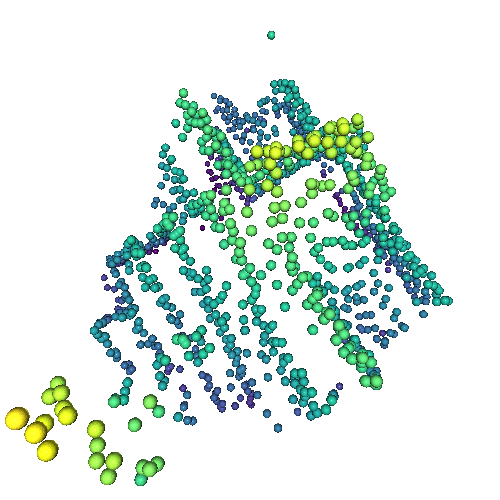} & 
        \smallImg{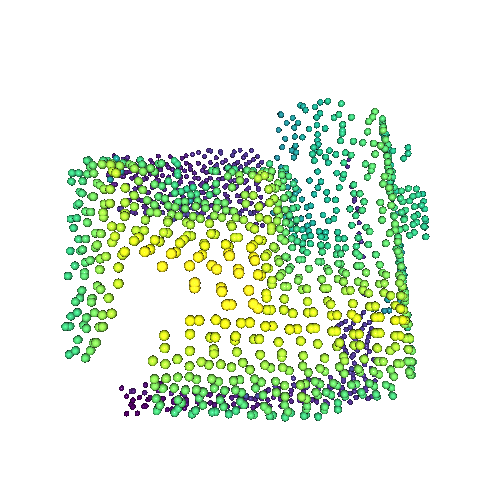} & 
        \smallImg{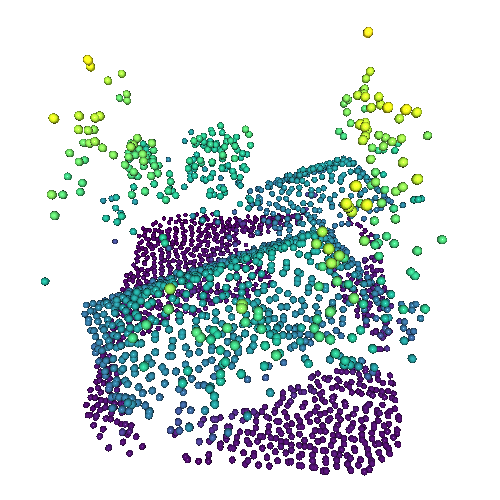} & 
        \smallImg{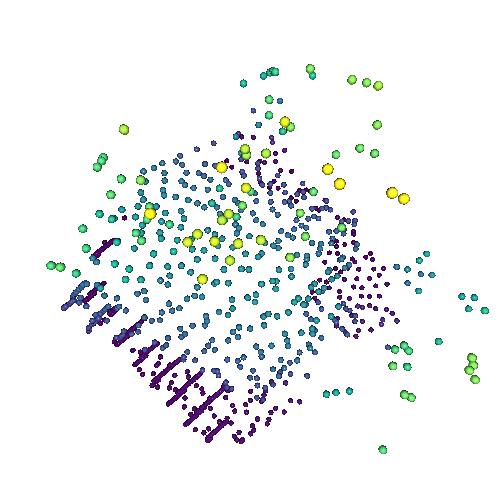} & 
        \smallImg{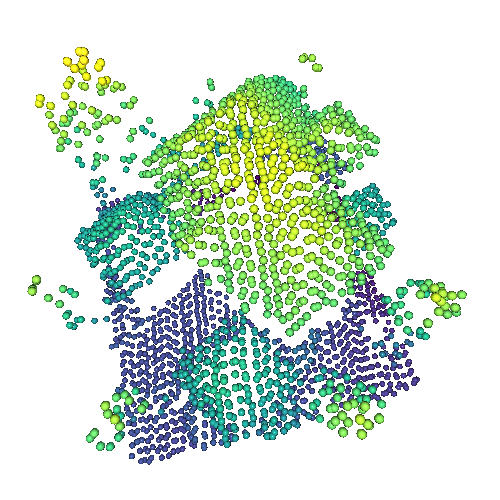} & 
        \smallImg{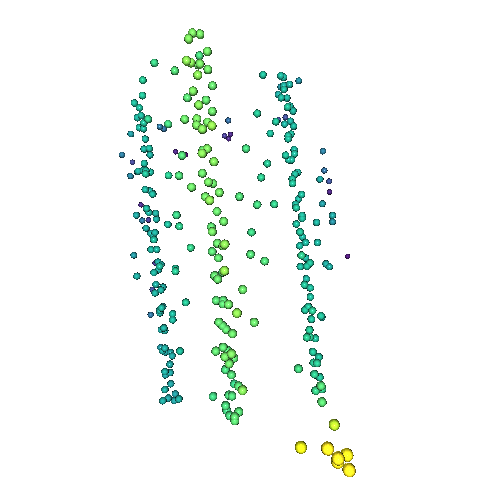} \\[-0.5ex]
    
        \scriptsize\rotatebox{90}{Input} & \smallImg{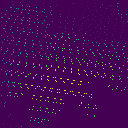} &
        \smallImg{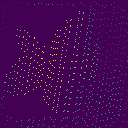} & 
        \smallImg{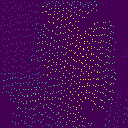} & 
        \smallImg{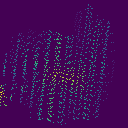} & 
        \smallImg{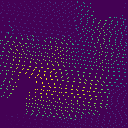} & 
        \smallImg{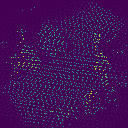} & 
        \smallImg{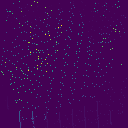} & 
        \smallImg{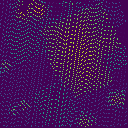} & 
        \smallImg{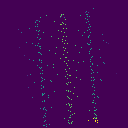} \\[-0.5ex]

        \scriptsize\rotatebox{90}{Linear} & \smallImg{fig_suppl/WayneCo_nf/Linear_building_103.png} & 
        \smallImg{fig_suppl/WayneCo_nf/Linear_building_102.png} &
        \smallImg{fig_suppl/WayneCo_nf/Linear_building_60.png} &
        \smallImg{fig_suppl/WayneCo_nf/Linear_building_106.png} & 
        \smallImg{fig_suppl/WayneCo_nf/Linear_building_11.png} & 
        \smallImg{fig_suppl/WayneCo_nf/Linear_building_25.png} & 
        \smallImg{fig_suppl/WayneCo_nf/Linear_building_81.png} & \smallImg{fig_suppl/WayneCo_nf/Linear_building_95.png} & \smallImg{fig_suppl/WayneCo_nf/Linear_building_120.png} \\[-0.5ex]
        
        \scriptsize\rotatebox{90}{pNCNN} & 
        \smallImg{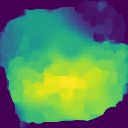} & 
        \smallImg{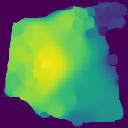} & 
        \smallImg{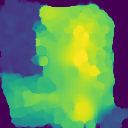} & 
        \smallImg{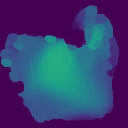} & 
        \smallImg{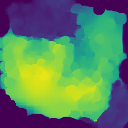} & 
        \smallImg{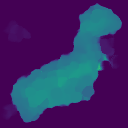} & 
        \smallImg{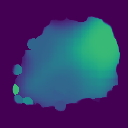} & 
        \smallImg{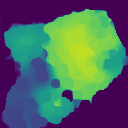} & 
        \smallImg{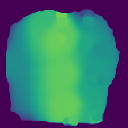} \\[-0.5ex]   

        \scriptsize\rotatebox{90}{CU-Net} & 
        \smallImg{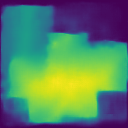} & 
        \smallImg{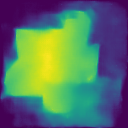} &
        \smallImg{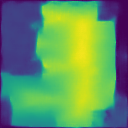} &
        \smallImg{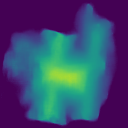} &
        \smallImg{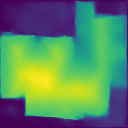} &
        \smallImg{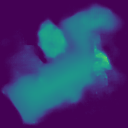} &
        \smallImg{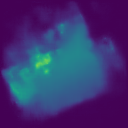} &
        \smallImg{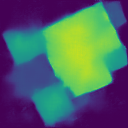} &
        \smallImg{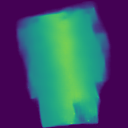} \\[-0.5ex] 

        \scriptsize\rotatebox{90}{Ours} & 
        \smallImg{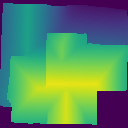} & 
        \smallImg{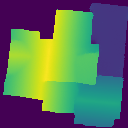} & 
        \smallImg{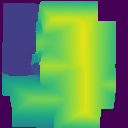} & 
        \smallImg{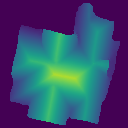} & 
        \smallImg{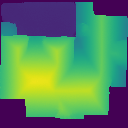} & 
        \smallImg{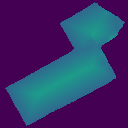} & 
        \smallImg{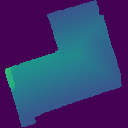} & 
        \smallImg{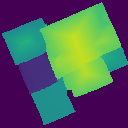} & 
        \smallImg{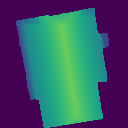} \\[-1ex]   

        \scriptsize\rotatebox{90}{Ours (3D)} & 
        \smallSubImg{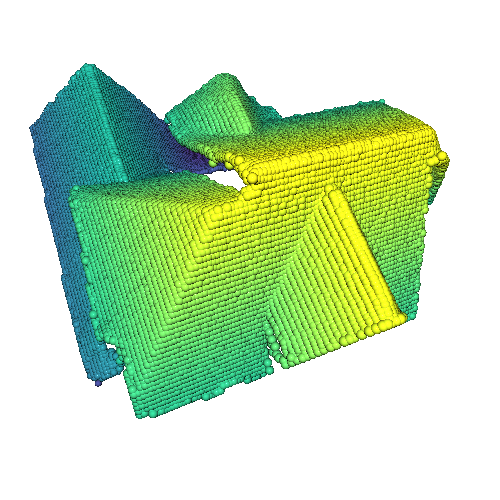}{}{} & 
        \smallSubImg{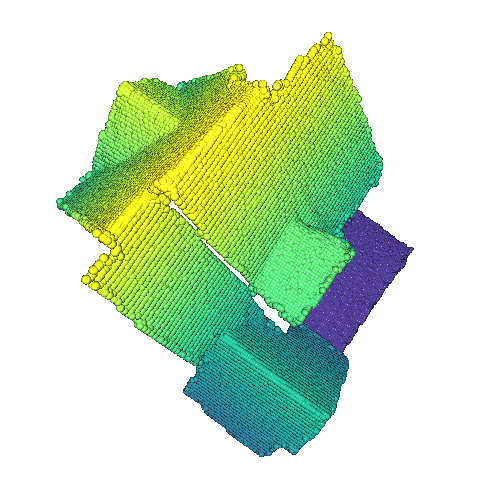}{}{} & 
        \smallSubImg{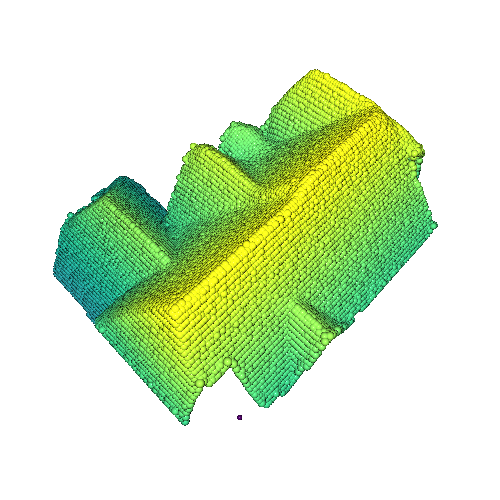}{}{} & 
        \smallSubImg{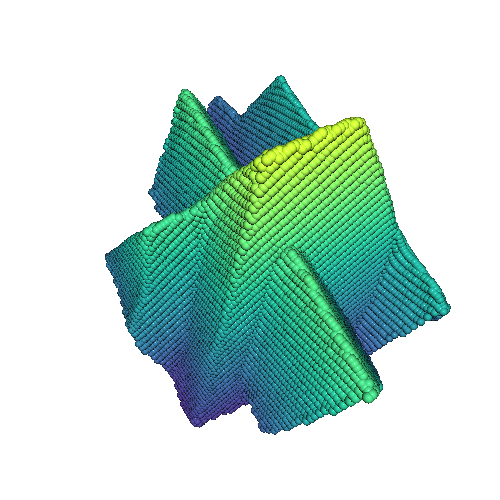}{}{} &
        \smallSubImg{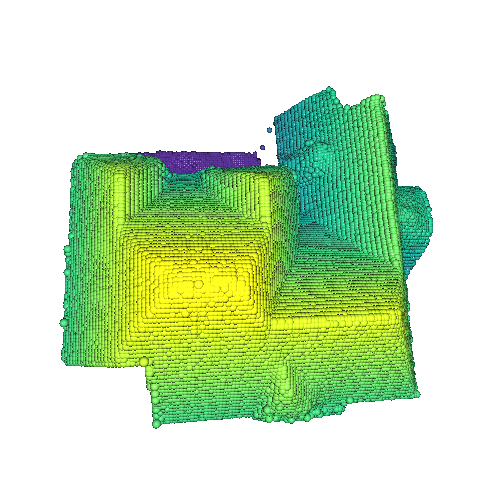}{}{} & 
        \smallSubImg{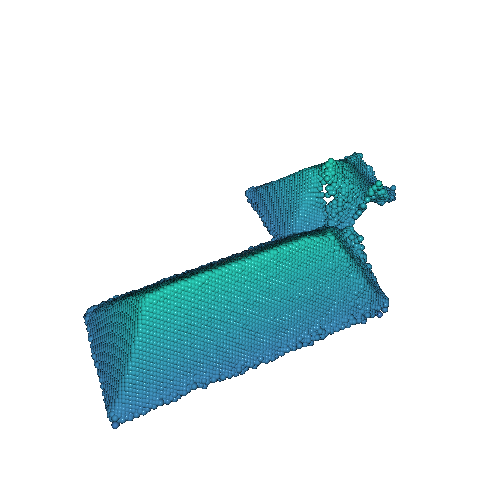}{}{} & 
        \smallSubImg{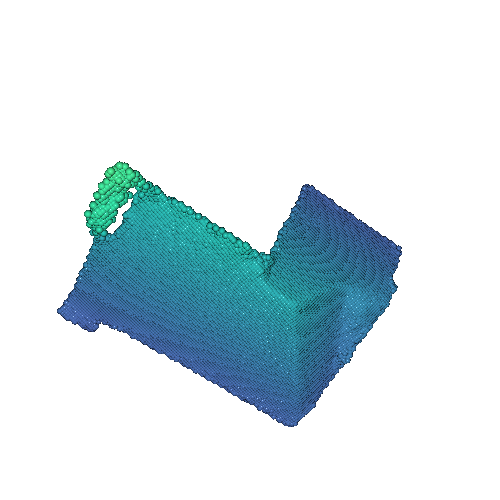}{}{} & 
        \smallSubImg{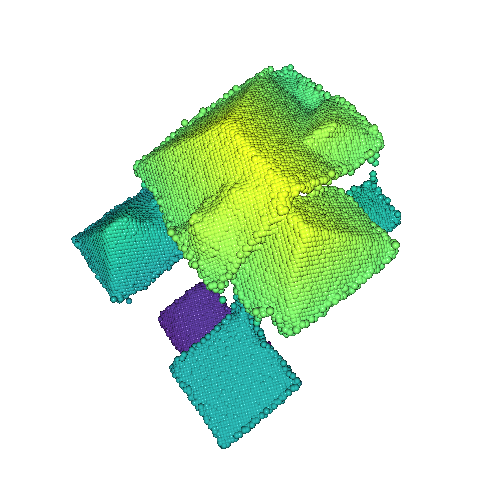}{}{} & 
        \smallSubImg{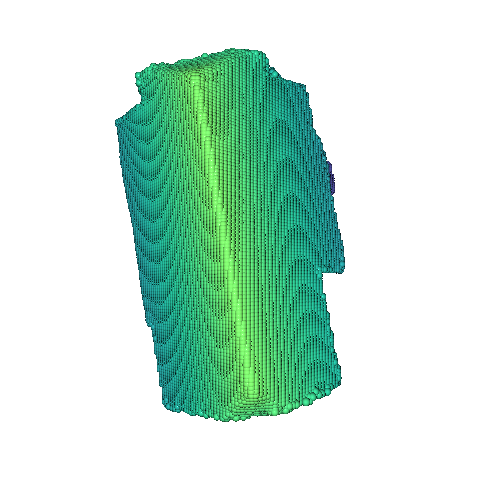}{}{} \\ 
        
    \end{tabular}
    \vspace{-0.8\baselineskip} 
    
    \caption{Evaluation of the completion and denoising for \rdN\ on USGS 3DEP LiDAR data sampled over Wayne County, MI \cite{wayneco2017}. Comparison with linear, pNCNN \cite{eldesokey2020uncertainty}, and CU-Net \cite{wang2022cu}.}
    \label{fig:morewayneco-nf}
    \vspace{-0.5\baselineskip}
\end{figure*}

\begin{figure*}[t]
    \newcommand{\repeatdots}[1]{%
      \foreach \n in {1,...,#1} {$\cdot$}%
    }
    \newcolumntype{G}{@{\hskip 1.9pt}}
    \newcolumntype{T}{m{0.015\textwidth}}
    \newcolumntype{I}{>{\centering\arraybackslash}m{\SmallImageSize}}
    \vspace{-\baselineskip}
    \centering

    \begin{tabular}{G T G I G I G I G I G I G I G I G I G I}


        \scriptsize\rotatebox{90}{Input (3D)} & 
        \smallImg{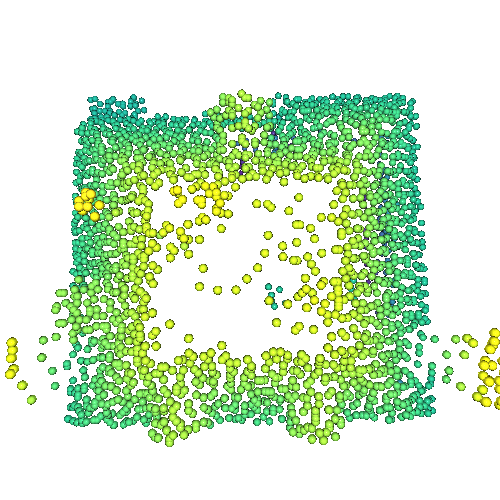} & 
        \smallImg{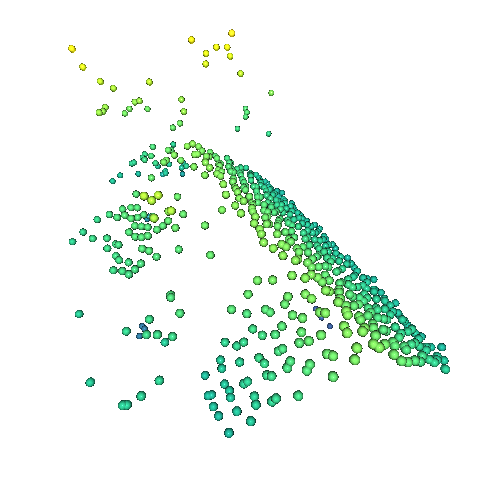} & 
        \smallImg{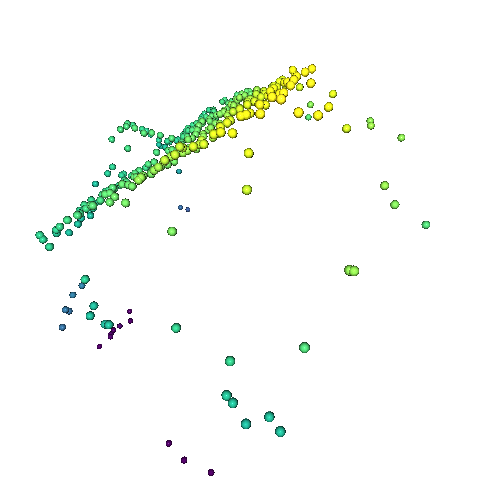} & 
        \smallImg{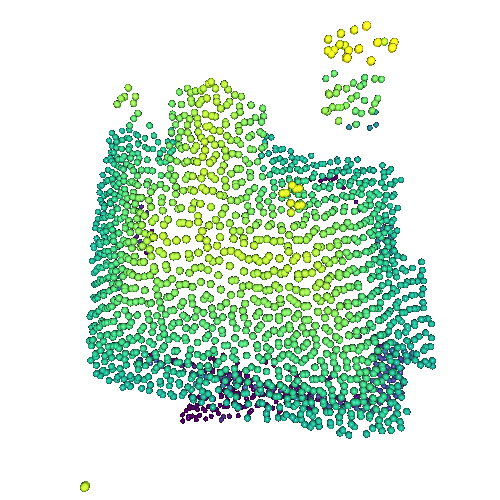} & 
        \smallImg{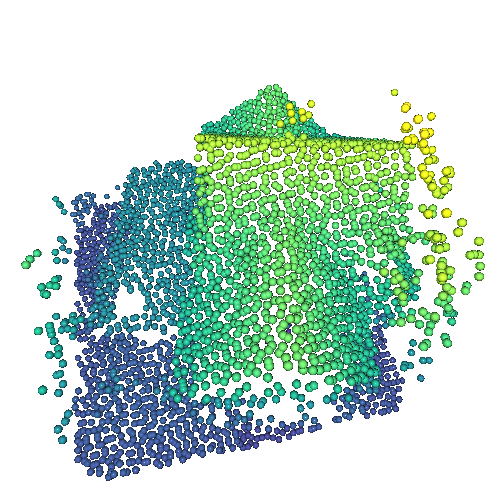} & 
        \smallImg{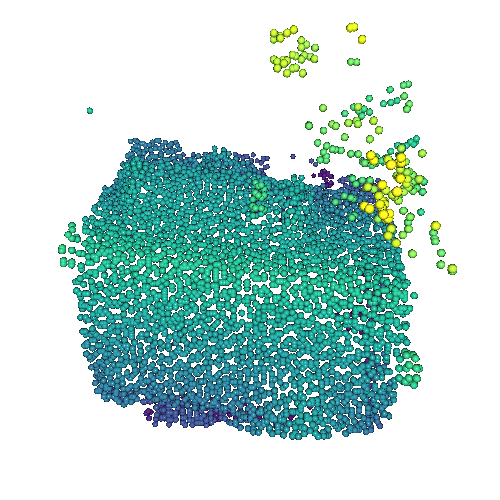} & 
        \smallImg{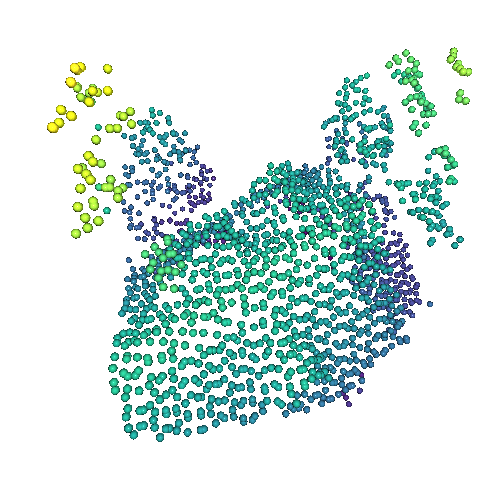} & 
        \smallImg{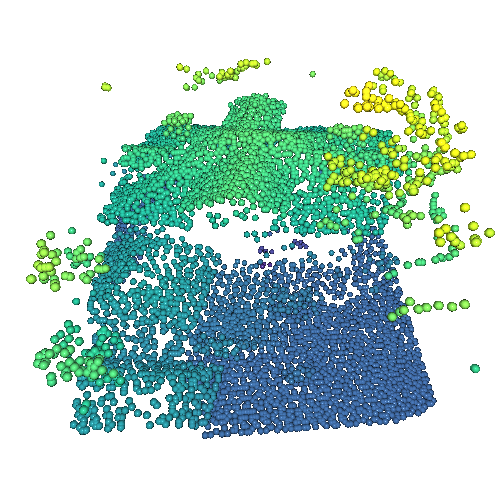} & 
        \smallImg{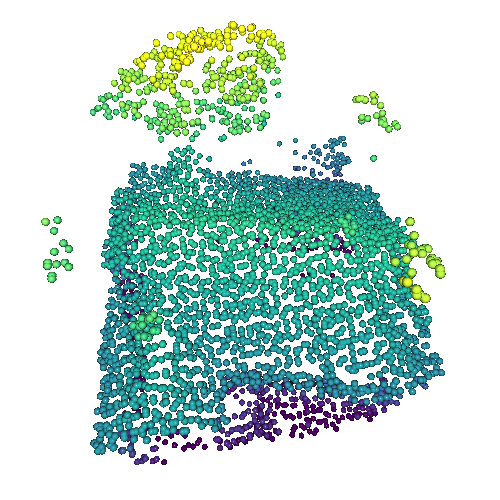} \\[-0.5ex]

        \scriptsize\rotatebox{90}{Input} & 
        \smallImg{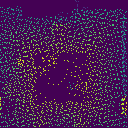} & 
        \smallImg{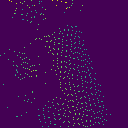} & 
        \smallImg{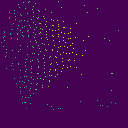} & 
        \smallImg{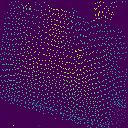} & 
        \smallImg{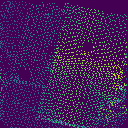} & 
        \smallImg{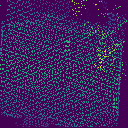} & 
        \smallImg{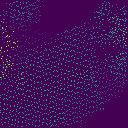} & 
        \smallImg{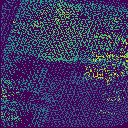} & 
        \smallImg{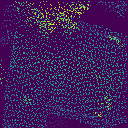} \\[-0.5ex]

        \scriptsize\rotatebox{90}{Linear} & 
        \smallImg{fig_suppl/Cambridge_nf/Linear_building_101.png} & 
        \smallImg{fig_suppl/Cambridge_nf/Linear_building_178.png} & 
        \smallImg{fig_suppl/Cambridge_nf/Linear_building_190.png} & 
        \smallImg{fig_suppl/Cambridge_nf/Linear_building_122.png} & 
        \smallImg{fig_suppl/Cambridge_nf/Linear_building_124.png} & 
        \smallImg{fig_suppl/Cambridge_nf/Linear_building_155.png} & 
        \smallImg{fig_suppl/Cambridge_nf/Linear_building_262.png} & 
        \smallImg{fig_suppl/Cambridge_nf/Linear_building_302.png} & 
        \smallImg{fig_suppl/Cambridge_nf/Linear_building_72.png} \\[-0.5ex]
        
        \scriptsize\rotatebox{90}{pNCNN} & 
        \smallImg{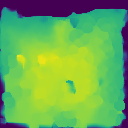} & 
        \smallImg{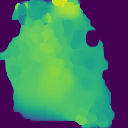} & 
        \smallImg{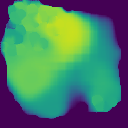} & 
        \smallImg{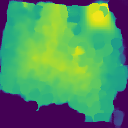} & 
        \smallImg{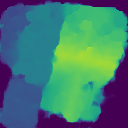} & 
        \smallImg{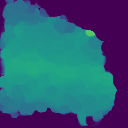} & 
        \smallImg{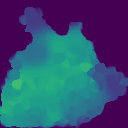} & 
        \smallImg{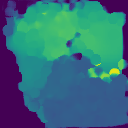} & 
        \smallImg{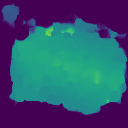} \\[-0.5ex]  

        \scriptsize\rotatebox{90}{CU-Net} & 
        \smallImg{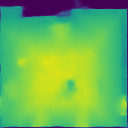} & 
        \smallImg{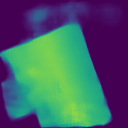} & 
        \smallImg{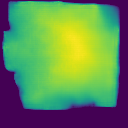} & 
        \smallImg{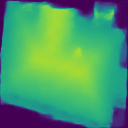} & 
        \smallImg{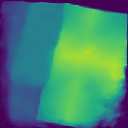} & 
        \smallImg{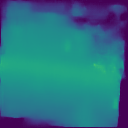} & 
        \smallImg{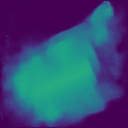} & 
        \smallImg{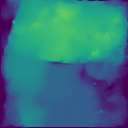} & 
        \smallImg{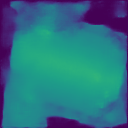} \\[-0.5ex]

        \scriptsize\rotatebox{90}{Ours} & 
        \smallImg{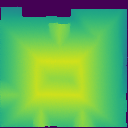} & 
        \smallImg{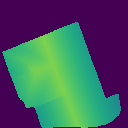} & 
        \smallImg{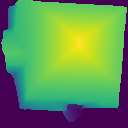} & 
        \smallImg{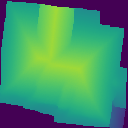} & 
        \smallImg{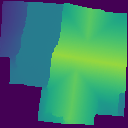} & 
        \smallImg{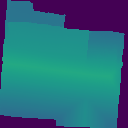} & 
        \smallImg{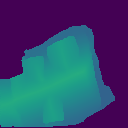} & 
        \smallImg{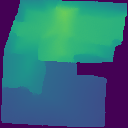} & 
        \smallImg{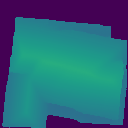} \\[-1ex]

        \scriptsize\rotatebox{90}{Ours (3D)} & 
        \smallSubImg{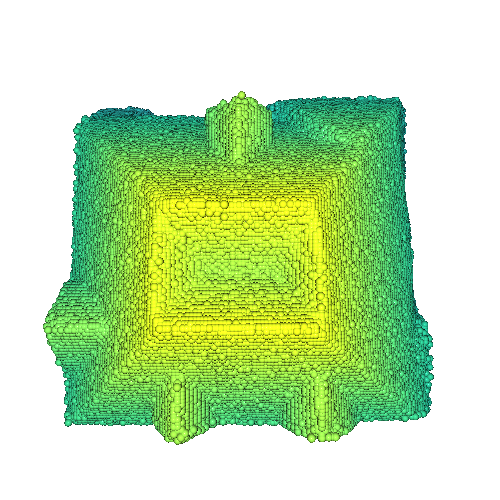}{}{} & 
        \smallSubImg{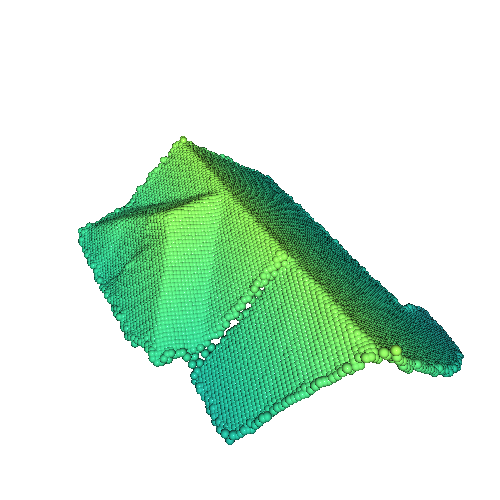}{}{} & 
        \smallSubImg{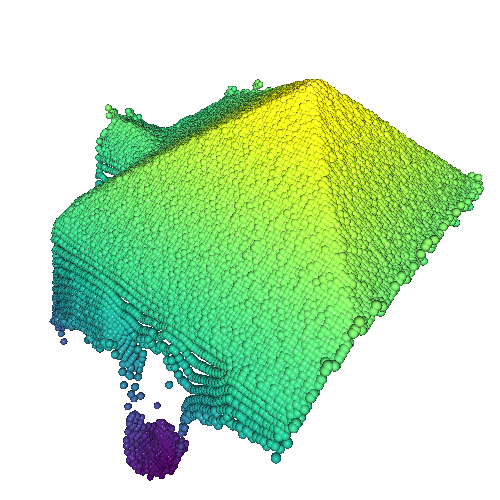}{}{} & 
        \smallSubImg{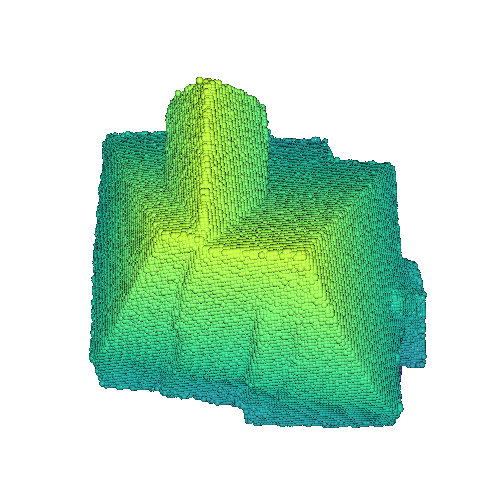}{}{} & 
        \smallSubImg{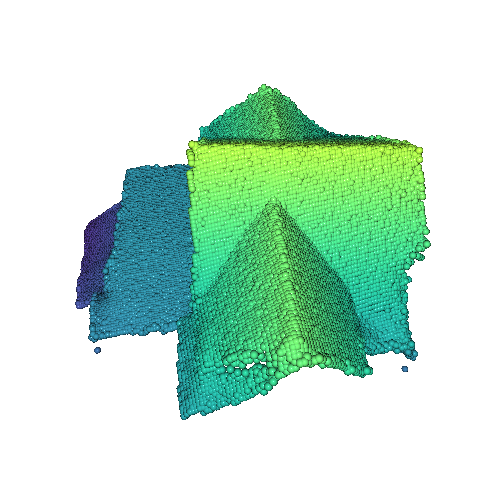}{}{} & 
        \smallSubImg{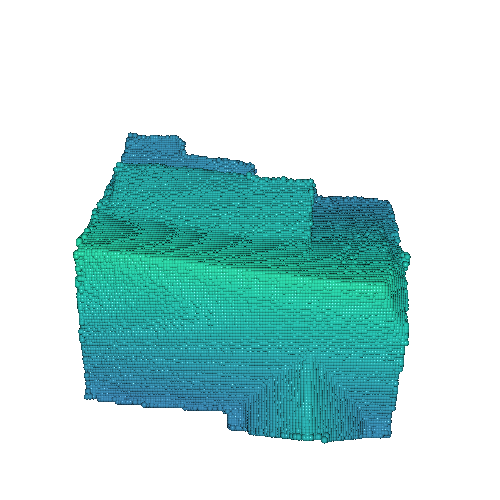}{}{} & 
        \smallSubImg{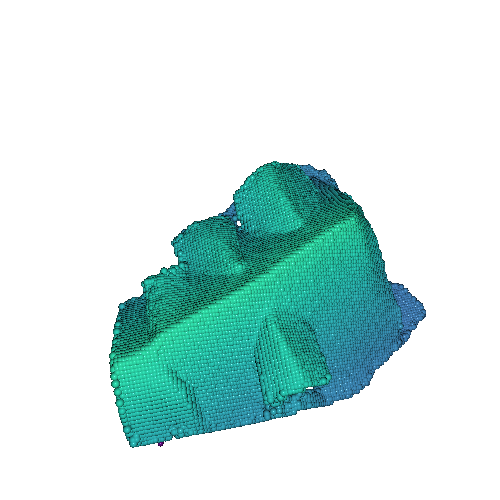}{}{} & 
        \smallSubImg{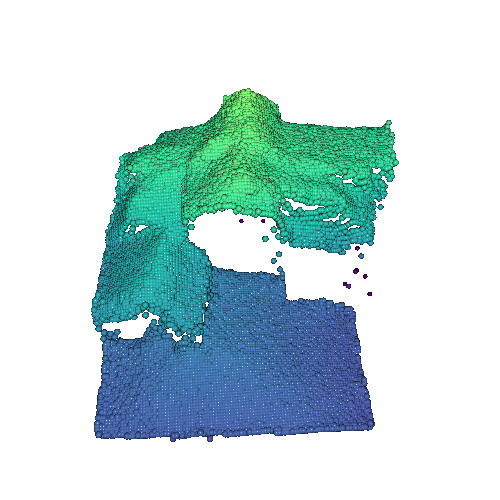}{}{} & 
        \smallSubImg{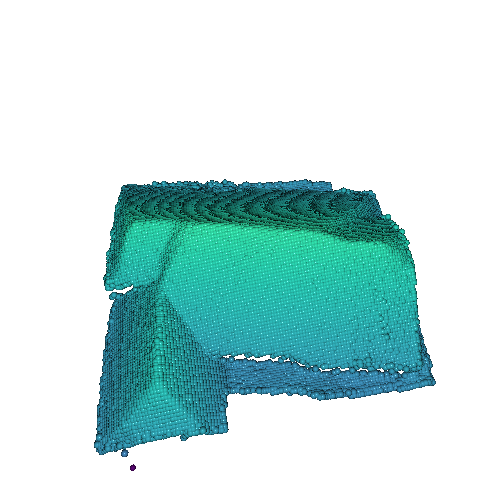}{}{} \\
        
    \end{tabular}
    \vspace{-0.8\baselineskip} 
    
    \caption{Evaluation of the  completion and denoising for \rdN on USGS 3DEP LiDAR data sampled over Cambridge, MA \cite{cambridge2021}.}
    \label{fig:morecambridge-nf}
    \vspace{-0.5\baselineskip}
\end{figure*}

\begin{itemize}
    \item \Cref{fig:morewayneco-nf} provides the results on USGS 3DEP LiDAR sampled over Wayne County, MI \cite{wayneco2017}.
    \item \Cref{fig:morecambridge-nf} provides the results on USGS 3DEP LiDAR sampled over Cambridge, MA \cite{cambridge2021}.
\end{itemize}

\clearpage 

\clearpage


\clearpage 

\clearpage


\clearpage

\clearpage

\clearpage 



\end{document}